\documentclass[a4paper, 11pt]{article}
\usepackage[affil-it]{authblk}
\newtheorem{definition}{Definition}
\newtheorem{theorem}{Theorem}

\usepackage{natbib}
\usepackage{graphicx}
\usepackage{amssymb}

\usepackage{pdflscape}
\usepackage{rotating}
\usepackage[paper=portrait,pagesize]{typearea}

\usepackage{amsmath}
\usepackage{algorithm2e}

\usepackage{tikz}
\usetikzlibrary{arrows, automata}

\usepackage{hyperref}
\usepackage{dsfont}

\begin{document}
\sloppy

\title{Distillation of Weighted Automata from Recurrent Neural Networks using a Spectral Approach\thanks{This paper is the authors' version of an article to be published at the Machine Learning Journal. This work is supported in part by the ANR TAUDoS project.}
}


\author{R\'emi Eyraud}
\affil{Univ Lyon, UJM-Saint-Etienne, CNRS, Laboratoire Hubert Curien UMR 5516, Saint-Etienne F-42023, France}

\author{St\'ephane Ayache}
\affil{Aix Marseille Univ, Universite de Toulon, CNRS, LIS, Marseille, France}



\maketitle

\begin{abstract}
This paper is an attempt to bridge the gap between deep learning and grammatical inference. Indeed, it provides an algorithm to extract a (stochastic) formal language from any recurrent neural network trained for language modelling. In detail, the algorithm uses the already trained network as an oracle -- and thus does not require the access to the inner representation of the black-box -- and applies a spectral approach to infer a weighted automaton. 

As weighted automata compute linear functions, they are computationally more efficient than neural networks and thus the nature of the approach is the one of knowledge distillation. 
We detail experiments on 62 data sets (both synthetic and from real-world applications) that allow an in-depth study of the abilities of the proposed algorithm.  The results show the WA we extract are good approximations of the RNN, validating the approach. Moreover, we show how the process provides interesting insights toward the behavior of RNN learned on data, enlarging the scope of this work to the one of explainability of deep learning models. 

\end{abstract}

\section{Introduction}
In his founding paper, \citet{klee56} studies the expressive power of the ancestors of our current neural networks, namely the nerve nets of \citet{mccu43}. To do so, he introduces two computational models that are now at the core of theoretical computer science: rational expressions and finite state automata. However, despite of this common origin, the fields of theoretical computer science and of deep learning look nowadays almost completely disjoint: the type of mathematics used, the pursued goals, and the followed research paths are rarely the same. 

The context of the work presented in this article lies exactly within the frontier of these two fields and is thus an attempt to bridge this gap: it proposes a new way to link finite state machines and deep neural networks for sequential data.

When this kind of data is under consideration, deep learning approaches often rely on the use of Recurrent Neural Network (RNN). Indeed,
these models are known to obtain state of the art learning results in various practical tasks, including speech processing~\citep{grav13}, automatic natural language translation~\citep{ChoMGBSB14}, or caption generation~\citep{xuc15}.

While a wide number of different RNN types exists~\citep{lipt16,sale18}, they are all relying on the same mechanism: elements of the sequence are taken care of one at a time; for each element an output is computed; this output depends on the current element and of the part of the sequence seen so far. The memory mechanism used to remember the previously seen part is often based on the values outputted by the recurrent layer(s) on the previous element: this is usually a vector in a latent space that can be understood as the state of the network while dealing with the current element. The current output is thus computed using the current input and the value of the state-vector, making RNN \textit{de facto} state machines. Their state space is infinite as it is a subset of $\mathbb{R}^n$, with $n$ being an integer fixed by the architecture of the network (it corresponds usually to the size of the recurrent layer).

Despite their impressive learning ability in practice, RNN suffer from two main drawbacks. The first one is usually called the black-box problem: the process allowing a RNN to take a decision or to deal with a given task is not understandable by human beings. The computational use of a large amount of parameters and the combination of non-linear functions forbid the interpretation of its behavior. In other words, though the learning phase is likely to provide a trustworthy model, it brings no human knowledge about the problem and its resolution. Worst, in case of failure, no feedback can be inferred to provide ways to tackle the issue. One has thus to accept to blindly follow an opaque model...

One usual path followed to tackle this issue is called \textit{local interpretability}~\citep{guid18}. In this line of work, one aims at explaining how a neural network takes a decision on a particular datum. This usually relies on finding clues linking the decision to a particular part of the input~\citep{Ribeiro0G16}. For RNN, this could take for instance the form of highlighting the importance of each elements of the sequence based on its importance to the model's prediction~\citep{LiMJ16a,wall18}.

But this approach is limited and it does not tackle the second drawback of RNN, that is, their computational cost. If it is nowadays admitted that a learning procedure could take hours or even days on state of the art computer grids, the use of the learned model should be computationally efficient. However, RNN may not have this property: their number of parameters grows quickly with their size and they are computing a composition of non-linear functions using these parameters. A consequence is for instance the wide use of cloud computing for cell phone applications requiring RNN~\citep{chung17}, which at the same time forbids users with no or poor data connection to use them and is an ecological non-sense, in contradiction with green computing goals~\citep{green09}. 

A way to overcome these two drawbacks, that are actually common to most deep artificial neural networks, is to achieve a \textit{(knowledge) distillation}~\citep{hint15,furl18}, a process closely related to variational approximation~\citep{metro49}. In this approach, a (deep) neural network is first learned, while in a second step a simpler and less computationally costly model is extracted from the network. A well-known example of such procedure is the distillation of decision trees from usual feed-forward neural networks~\citep{Haya90,crav95,schm99, fros17}.

\citet{hint15} have introduced the term distillation as the task of providing computationally efficient models by transferring knowledge from a complex architecture. This line of works usually focus on a teacher-student paradigm where a large "teacher" network guides the learning of a smaller "student" model. The knowledge transfer is typically achieved using a loss that aims at aligning 
outputs of the student with the ones of the teacher.






The models obtained by distillation usually enjoy one or both of two interesting features: they provide an easier \textit{global interpretation} of the decision process while requiring less computational power. This constitutes the main motivation for the present article that aims at distilling finite state machines, namely Weighted Automata (WA), from already trained RNN. These models have been extensively studied (see \citet{dros09} for an overview), require only linear computations, and benefit from a graphical representation. 

Our approach, whose preliminary details have been published in the proceedings of the International Conference on Grammatical Inference~\citep{ayac18}, relies on an adaptation of the spectral learning algorithm for WA~\citep{balle14}. While this approach requires a data set as input, our algorithm works from a trained RNN and uses it as an oracle, querying it well chosen sequences to get the value the network attributes to them. It then infers the automaton from a mathematical object called the Hankel matrix, storing the result of these queries.

The novelty of our distillation algorithm is mainly due to two factors: it does not require learning data and it does not need to access the inside of the RNN. This last point makes it adapted to work for any architecture, any type of cells, and even for black-boxes that are not RNN, providing that they are computing a function on sequential data. 

This article firstly includes a survey on methods to distill finite state machines from RNN (Section~\ref{sec:previous}). After carefully introducing notations and defining main notions in Section~\ref{sec:prelim}, we detail our approach in Section~\ref{sec:algo}. The experimental framework is given in Section~\ref{sec:framework} and the results of the experiments in Section~\ref{sec:results}. 
Section~\ref{sec:conclu} concludes this article.

\section{Scientific context}
\label{sec:previous}
In this section, we detail the main works that aim to bridge the gap between RNN and the models used in classical computer science, those of the formal language theory.

Though more details are given in Section~\ref{sec:prelim}, we need first to recall the formalization of the general behaviour of RNN: at time-step $t$, that is, on the $t^{th}$ element on the input sequence, the output of the recurrent layer(s) is given by:
$$h_t = \mathcal{R}(x_t,h_{t-1})$$
where $x_t$ is the  $t^{th}$ element of the sequence, and $\mathcal{R}$ the recurrent or update function. 
As already stated, the vector $h_t$ is the value of the state of the RNN after seeing the $t^{th}$ element.

\subsection{Relation of RNN classes to other computational models}
When the expressive power of RNN is under consideration, one usually refers to the work of \citet{sieg92,sieg94}: it shows that Simple Recurrent Neural Networks (SRNN)~\citep{elma90} are Turing complete, that is, that any function computable using a Turing machine can be computed using a SRNN. These networks deserve their name: the update function for the hidden state correspond to applying an activation function $f$ to the weighted sum of the previous state and the current input:
$$h_t = f(Wx_t+Uh_{t-1}+b)$$
where $W$ and $U$ are weights matrices, and $b$ is a scalar.

As impressive as this result might look, it is of little help in practice: the proof requires to have rational weights of infinite precision, an unbounded computation time, and relies on a mechanism allowing the network to "think" before computing an output. This is far from corresponding to the networks used nowadays.
In addition, not all recurrent neural networks can be learned using a gradient descent algorithm, the approach currently used for this task. However, few papers have tackled the more practical question of  the characterization of the classes of learnable finite precision and time neural nets.

One exception is the work of \citet{rabu19} where the authors show an equivalence between Weighted Automata (WA) (detailed in Section~\ref{subsec:wa}) and linear second order RNN. These models are defined as follow:
$$h_t = f(T\cdot x_t\cdot h_{t-1})$$
where $T$ is a tensor of order 3. 

\noindent 
The obtained result also asserts that the number of states in the minimal WA computing a given function is exactly the number of neurons in the hidden layer of the RNN for the same function, adding a structural equivalence to the expressivity one. In addition, the paper shows how a learning algorithm for WA can be adapted to this type of RNN, keeping its proven learning capacity. 

Another recent exception is the work of \citet{weis18b} that compares different recurrent network architectures. It empirically shows that finite precision SRNN and Gated Recurrent Units (GRU)~\citep{ChoMGBSB14} cannot compute non-regular recognizers, while Long Short Term Memory (LSTM)~\citep{hoch97} can be trained to recognize some context-free and even some context-sensitive dependencies. The paper also proposes a connection between LSTM and a simplified version of counter automata~\citep{hopc79}.

Following this work, \citet{merr19} focuses on the kinds of computation that real-time, bounded-precision neural networks can perform by relating them to different types of automata. In this theoretical work, different types of RNN (and CNN) are studied in a framework the author called \textit{asymptotic convergence}: their expressiveness is considered at point-wise convergence~\citep{rudi76}. In this context, the space state of RNN tends to be discrete and the following results can be proven: LSTM behave similarly to counter machines; the class of SRNN-acceptable and GRU-acceptable languages is exactly the regular languages.
Empirical studies partially confirm these results, though some experiments show a clear limit to the asymptotic framework by contradicting some of the obtained theoretical results. 
A recent direct continuation of this work~\citep{merr20} renames the framework into \textit{saturated RNN} study, adds to state expressiveness the notion of language expressiveness\footnote{While state expressiveness focuses on the variability that can be obtained within the latent space of a type of RNN, language expressivity deals with the languages potentially recognizable from this vector when using different functions that map it to a Boolean.}, and propose a hierarchy based on this elements covering a large amount of RNN classes.

Adopting a different research path, \citet{avcu17} investigate the learning ability of RNN on different sub-regular classes of languages: their experiments show that long term dependencies are not always learned by RNN, even by gated formalism like LSTM. \citet{maha19} recently continue this line of work and confirm its conclusions on large benchmarks.

Other approaches have been proposed for this goal of studying RNN in relation to other computational models. For instance, it is worth to cite the work of \citet{chen17}. Though it does not link RNN to another computational model \textit{per se}, it aims at studying single-layer, ReLU-activation, rational-weight RNN using usual formal language approaches. It concludes that classical properties of computational models, such as consistency,  equivalence,  and minimization, are not decidable for this type of RNN, while these problems are known to be tractable for models like WA.  

Finally, in a recent article, \citet{marz20} adopt also the classical approach of theoretical computer science to provide insights on the link between RNN and several classes of stochastic finite state machines. They show in particular that the equivalence problem of classic classes of finite state machines and weighted first-order RNN is not decidable in general. They also show how the problem becomes decidable when restricting the study to particular (and reasonable) RNN sub-classes, though its complexity remains intractable (NP-hard).

\subsection{Distillation of binary classifiers on sequential data}
First ideas for knowledge distillation from RNN concentrate on binary classifiers, sometime called RNN-acceptors. Indeed, early works train a RNN on data coming from a Deterministic Finite State Automaton (DFA) and then propose an algorithm to extract a DFA from the RNN. As DFA are language models, the output of the learned RNN consist of 2 possible values: $+1$ if the complete input sequence is in the language, $-1$ (or $0$) otherwise.

Though first attempts consist in feeding a RNN with training data coming from a DFA and then using a test set to determine if it learned correctly (see for instance the work of \citet{clee89} and \citet{watr92}), some of these works already focus on distilling a DFA from RNN. 

To our knowledge, the first attempt at this task is the work of \citet{gile92} where the authors use a quantization algorithm on the state space of second order RNN. This approach tries to partition the RNN state space by dividing each dimension into equal intervals. If this was tractable with the RNN used in the 90's, it clearly is not with today networks, given the explosion of the dimension of the internal state.

Another idea is to use a clustering algorithm on the state space of RNN. Though \citet{clee89} are already using hierarchical clustering to verify that the structure of the RNN state space corresponds to the one of the target DFA, distillation of DFA using that approach is
first introduced by \citet{gile92} and enjoys many refinements (see for instance the work of \citet{mano94}, \citet{omli96}, or \citet{cech03}).
As in the previous approach, data is fed to a RNN already trained on data coming from a DFA and the vector values obtained all along the different input sequences is stored. Then a clustering algorithm, like the k-means one~\citep{stei57}, is used on these vectors to get $k$ clusters, each one corresponding to a state. The transitions between states are then obtained by parsing selected sequences into the RNN to get the cluster the resulting vector belongs to. 

\noindent
These works are the starting point of numerous refinements, most of which are detailed in the survey of \citet{jaco05} and empirically compared on different RNN architectures in a recent paper by \citet{wang17}.

\citet{weis18} recently open a new approach for the distillation of DFA from RNN: they adapt the L* algorithm from \citet{angl87} so that the RNN plays the role of the oracle. The algorithm first completes a table whose rows correspond to prefixes and columns to suffixes by querying the RNN to know which sequence made of one given prefix and one given suffix is recognized by the RNN. Then a DFA is inferred from the table and the algorithm has to decide whether this DFA is equivalent to the RNN, a problem recently shown to be undecidable~\citep{marz20}. While in the initial theoretical framework this step is straightforward, it requires carefully defined heuristics when distilling from a RNN: the authors propose an iterative process that parse at the same time the distilled DFA and another one obtained by partitioning the RNN state space using SVM. 
The latter is refined recursively until one can conclude the RNN and the DFA are equivalent, or a sequence on which the distilled DFA and the RNN disagree is obtained. If this last event occurs, the process continues since the obtained counter example allows modifications of the table and thus the distillation of a better DFA. Their experiments show that their algorithm distills in an efficient way smaller and more accurate DFA than previous methods, allowing a better understanding of the behaviour of the RNN. 

To be complete on the subject, one should cite the work of \cite{wang2018verification} where the authors extract DFA from different types of RNN in order to find adversarial examples. Their experiments show that such examples can be exhibited for
almost all architectures, the type of examples challenging the robustness depending on the type of RNN considered. 

\subsection{Distillation of estimators}
While of great interest, distillation of DFA from neural networks suffers from one important drawback: it does not correspond to the practical use of RNN. Indeed, these models are rarely trained for binary classification. While other practical frameworks exist (see for instance the work of \citet{ChoMGBSB14}), RNN are mainly trained for \textit{language modeling}, a setting in which the models are trained to guess the probability of each potential symbol to be the next in the sequence (a particular symbol often marks the end of sequence). After each new input symbol $x_t$, these Language Modelling RNN (LM-RNN) thus output a vector of real values, each value corresponding to the estimated probability of a given symbol to be the following one in the sequence.

As far as we know, first attempts at such a distillation process take place in the context of speech recognition, where having computationally efficient model is critical. For instance, the work of \citet{deor11} proposes to sample a LM-RNN and then infers a $n$-gram from the sampled sequences. Experiments show better results than just inferring $n$-gram from training sample of usual NLP benchmarks.

A second approach, still in the context of early phases of speech recognition, is the work of \citet{leco12}. Their algorithm aims at distilling Weighted Automata from LM-RNN: they use the k-means algorithm to cluster vectors that consist of the concatenation of vectors from the state space of the network and the last symbol seen. Their approach refines iteratively this discretisation in a greedy manner and massively uses domain specific heuristics to avoid the explosion of the size of the distilled WA and to allow back-tracking. Experiments show globally better results than the previously cited approach using $n$-gram. 

In a recent article, \citet{okud19} extended the work on DFA of \citet{weis18} to WA, focusing on RNN assigning a single real value to sequences. Instead of having a binary table, they store the weights of the sequence made of the concatenation of the prefix defining the row and the suffix of the corresponding column. To complete the table, they replace the idea of distinct row values by the notion of co-linearity. Then, using \citet{ball15} algorithm, they distill a WA and, finally, need to check its (approximate) equivalence with the RNN. To achieve this last goal, they open the black-box and try to iteratively learn a regression function from the observed states of the RNN to WA configurations, that is, the real value vectors one can obtain along the parsing of a sequence into a WA (see Section~\ref{subsec:wa} for details). If they exhibit such function, then the WA and RNN are considered equivalent, up to a predefined threshold. If not, the process gives them a counter example, and the algorithm continues in the same way than in the DFA distillation case. 

Finally, \citet{weis19} have also recently extended their approach to distill deterministic weighted automata from LM-RNN. Their key ideas are to use conditional probabilities to fill the table and a tolerance hyper-parameter to distinguished the states from the table, as co-linearity is a too strong requirement, especially in the presence of noise. Reported empirical results are promising though the method requires some heuristics to be run in reasonable time and the obtained automata can consist of thousands of states.  


\section{Preliminaries}
\label{sec:prelim}
\subsection{Elements of language theory}
\label{subsec:elem}
In theoretical computer science, a finite set of symbols is called an \textit{alphabet}, usually denoted by the Greek letter $\Sigma$. Language theory mainly deals with finite sequences on an alphabet that are called \textit{strings} and are usually denoted by the letters $w$, $v$, or $u$. We denote the set of all possible strings over $\Sigma$ by $\Sigma^*$.
For instance, the alphabet can be the ASCII characters, the 4 nucleobases of DNA, Part-of-Speech tags or lemmas from Natural Language Processing, or even a set of symbols obtained by the discretization of a time series~\citep{dimi10}.  

Throughout the paper, we will use other notions from language theory: the \textit{length} of a string $w$ is the number of symbols of the sequence (denoted $|w|$); the string of length zero is denoted $\lambda$; 
given 2 strings $u$ and $v$ we note $uv$ their concatenation; if a string $w$ is the concatenation of strings 
$u$ and $v$, $w=uv$, we say that $u$ is a \textit{prefix} of $w$ and that $v$ is a \textit{suffix} of $w$. Given a set of strings $S$, we denote $\textrm{pref}(S)$ the set of all the prefixes of the elements of $S$.  

\subsection{Functions on sequences}
\label{subsec:func}
In this article we consider functions that assign real values to strings: $f: \Sigma^* \to \mathbb{R}$. These functions are known under the name of \textit{series}~\citep{saka2009}. In particular, probability distributions over strings are such functions.
Each of these functions is associated with a specific object that has been proven to be extremely useful: 

\begin{definition}[Hankel Matrix~\citep{bers88}] 
Let $f$ be a series over $\Sigma^*$. The Hankel matrix of $f$ is a bi-infinite
matrix $\mathcal{H} \in \mathbb{R}^{\Sigma^*\times\Sigma^*}$ whose entries are defined as $\mathcal{H}(u,v) = f(uv)$, $\forall u,v \in\Sigma^*$. Rows are thus indexed by prefixes and columns by suffixes.
\end{definition}

For obvious reasons, only finite sub-blocks of Hankel matrices are of interest. An 
easy way to define such sub-blocks is by using a \textit{basis} $\mathcal{B} = (\mathcal{P},\mathcal{S})$, where $\mathcal{P},\mathcal{S}\subseteq \Sigma^*$. 
If we note $p = |\mathcal{P}|$ and $s = |\mathcal{S}|$, 
the sub-block of $\mathcal{H}$ defined by $\mathcal{B}$ is the matrix $H_\mathcal{B}\in \mathbb{R}^{p\times s}$ with $H_\mathcal{B}(u, v) = \mathcal{H}(u, v)$ for any $u \in \mathcal{P}$ and $v \in \mathcal{S}$. 
We may write $H$ if the basis $\mathcal{B}$ is arbitrary or obvious from the context.

\subsection{Weighted Automata}
\label{subsec:wa}
The following definitions are adapted from \citet{mohri2009} and \citet{balle14}:
\begin{definition}[Weighted automaton]
A weighted automaton (WA) is a tuple $\langle \Sigma, Q, \mathcal{T}, \gamma, \rho \rangle$ such that:
\begin{itemize}
\item 
$\Sigma$ is a finite alphabet; 
\item
$Q$ is a finite set of states; 
\item 
$I\subseteq Q$ is the set of initial states; 
\item 
$\mathcal{T}: Q\times \Sigma \times Q \to \mathbb{R}$ is the transition function; 
\item 
$\gamma: Q \to \mathbb{R}$ is an initial weight function; 
\item 
$\rho: Q \to \mathbb{R}$ is a final weight function.
\end{itemize}
\end{definition}
A transition is usually denoted $(q_1,\sigma, p, q_2)$ instead of $\mathcal{T}(q_1,\sigma,q_2)=p$.
 We say that two transitions $t_1=(q_1,\sigma_1, p_1, q_2)$ and $t_2=(q_3,\sigma_2, p_2, q_4)$ are consecutive if $q_2=q_3$. A path $\pi$ is an element of $\mathcal{T}^*$ made of consecutive transitions. 
 We denote by $o[\pi]$ its origin and by $d[\pi]$ its destination. The weight of a path is defined by $\mu(\pi)=\gamma(o[\pi])\times\omega\times\rho(d[\pi])$ 
 where $\omega$ is the multiplication of the weights of the constitutive transitions of $\pi$.
 We say that a path $(q_0, \sigma_1, p_1, q_1)\ldots (q_{n-1}, \sigma_n, p_n, q_n)$ reads a string $w$ if $w=\sigma_1\ldots \sigma_n$.
 
The weight of a string $w$, that is, the real value assigned by the WA to $w$, is the sum of the weights of the paths that read $w$.
A weighted automaton assigns weights to strings, that is, it maps each string of $\Sigma^*$ to a real value. The set of functions computable by a WA is called the \textit{rational series}.

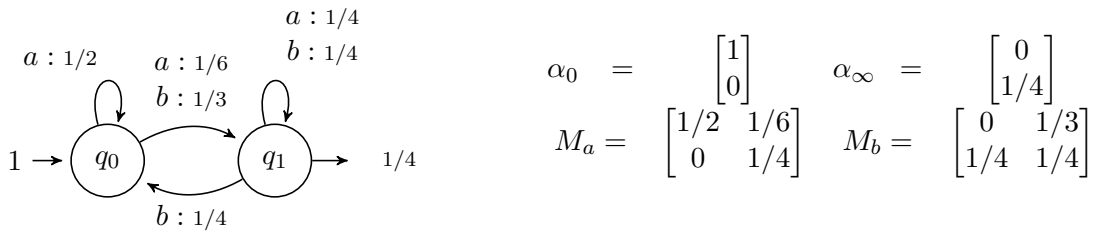
\begin{figure}[ht]
\begin{center}
\begin{tikzpicture}[->,>=stealth',shorten >=3pt,auto,node distance=2.2cm,semithick]
  \tikzstyle{every state}=[fill=none,draw=black,text= black]
  \tikzstyle{state with accepting}=[draw=none,fill=none]
  \node[initial left,state] (0)      [initial text={$1$}]              {$q_0$};
  \node[state]         (1) [right of=0] {$q_1$};

  \node[accepting] (T1) [right of=1]{};
  \path (0)edge                      [bend left]                           node {$\begin{matrix}a:{\scriptstyle 1/6}\\b:{\scriptstyle 1/3}\end{matrix}$} (1)
                 edge  [loop above,above left ] node {$a:{\scriptstyle 1/2}$} (0)
            (1) edge  [shorten >=31pt,right ] node {${\scriptstyle 1/4}$} (T1)
                 edge  [loop above ,above right] node {$\begin{matrix}a:{\scriptstyle 1/4} \\b:{\scriptstyle 1/4}\end{matrix}$} (1)
                  edge [bend left]  node {$b:{\scriptstyle 1/4}$} (0);
\end{tikzpicture}
   \hspace{1cm} 
$\begin{matrix}
\alpha_0\;\;\;=&\begin{bmatrix}1\\ 0 \end{bmatrix} & \alpha_\infty\;\;=&\begin{bmatrix}0\\1/4\end{bmatrix} \\
M_a=&\begin{bmatrix} 1/2&1/6 \\ 0&1/4 \end{bmatrix} &
M_b=&\begin{bmatrix} 0&1/3 \\ 1/4 &1/4 \end{bmatrix}\\
\ & \ & \ \\
\ & \ & \ \\
\ & \ & \ \\
\ & \ & \ \\
\ & \ & \ \\ 
\ & \ & \ \\ 
\ & \ & \ 
\end{matrix}$
\vspace{-1.8cm}
\end{center}
\caption{A WA in its graphical form and its equivalent linear representation.} 
\label{fig:PFA}
\end{figure}

\medskip
\noindent
WA admit an equivalent representation using linear algebra:
\begin{definition}[Linear representation~\citep{deni08,balle14}]
A~\textit{linear representation} of a WA $A$ is a triplet $\langle \alpha_0, (M_\sigma)_{\sigma\in\Sigma}, \alpha_\infty \rangle$ where
\begin{itemize}
    \item the vector $\alpha_0$ provides the initial weights (\textit{i.e.} the values of the function $\gamma$ for each state),
    \item the vector $\alpha_\infty$ is the terminal weights (\textit{i.e.} the values of function $\rho$ for each state),
    \item each matrix $M_\sigma$ corresponds to the $\sigma$-labeled transition weights ($M_\sigma(q_1,q_2)=p \Longleftrightarrow \mathcal{T}(q_1,\sigma,q_2)=p$).
\end{itemize}
The dimension of each element is the number of states of the automaton.
\end{definition}
Figure~\ref{fig:PFA} shows the same WA using the two representations. In what follows, we will consider that WA are defined in terms of linear representations, unless specified otherwise.

To compute the weight that a WA $A$ assigns to a string $w=\sigma_1 \sigma_2\ldots\sigma_n$ using a linear representation, it suffices to compute the product $A(w)=\alpha_0^\top M_w \alpha_\infty = \alpha_0^\top M_{\sigma_1} M_{\sigma_2}\ldots M_{\sigma_n}\alpha_\infty$. 

This can be interpreted as a projection into $\mathbb{R}^r$, where $r$ is the number of states of $A$, followed by an inner product~\citep{rabu19}. Indeed, $\alpha^0=\alpha_0^\top$ is a vector that corresponds to the initial projection; each of the following steps computes a new vector $\alpha^{i}$ in the same space, moving from one vector to the next one by computing the product with the corresponding symbol matrix ($\alpha^{i}=\alpha^{i-1} M_{\sigma_i})$; the final projection is 
$\alpha^{n}=\alpha_0^\top M_w$; the output of the WA on $w$ is given by the inner product $\langle
\alpha^{n}, \alpha_\infty\rangle$. 

$\alpha^i$ is sometime referred to the WA configuration after seeing the prefix $\sigma_1\ldots\sigma_i$. It is a vector of the latent state space of WA since the output of the models depends of this vector and the following symbols, in a standpoint analogue to the one of RNN. This is of crucial importance to understand WA: their state space is potentially infinite as it is a part of $\mathbb{R}^r$. It also distinguishes these models from classically used others, like DFA -- whose state space is discrete finite -- and in some sense Hidden Markov Models (HMM) -- which model distributions over a discrete finite set of states. 

If a WA computes a probability distribution over $\Sigma^*$, it is called a stochastic WA. In this case, it can easily be used to compute the probability of each symbol to be the next one of a given prefix~\citep{balle14}: the probability of $\sigma$ being the next symbol of the prefix $w$ is given by 
$$\alpha_0^\top M_w M_\sigma \tilde\alpha_\infty = \alpha^{|w|} M_\sigma \tilde\alpha_\infty$$
where $\tilde\alpha_\infty = (Id - (\sum_{\sigma\in\Sigma}M_\sigma))^{-1}\alpha_\infty$, with $Id$ the identity matrix.

\medskip
The following theorem is at the core of the spectral learning of WA (introduced independently by \citet{hsu2009} and \citet{bailly2009}) and of our approach (described in Section~\ref{sec:algo}):

\begin{theorem}[\citep{carl71,flie74}]
\label{thm:finiterank}
A function $f : \Sigma^* \to \mathbb{R}$ can be defined by a WA if and only if the rank of its Hankel matrix is finite. In that case this rank is equal to the minimal number of states of any WA computing $f$.
\end{theorem}

\subsection{Recurrent Neural Networks}
\label{subsec:RNN}
Recurrent Neural Networks (RNN) are artificial neural networks designed to handle sequential data. To do so, the RNN we consider incorporates an internal state that is used as memory to take into account the influence of previous elements of the sequence when computing the output for the current one.  

Two type of architecture units are mainly used: the widely studied Long Short Term Memory (LSTM)~\citep{hoch97} and the recent Gated Recurrent Unit (GRU)~\citep{ChoMGBSB14}. In both cases, these models realize a non-linear projection of the seen prefix into $\mathbb{R}^d$, where $d$ is the number of neurons on the penultimate layer: this vector is usually called latent representation of the part of the sequence seen so far. The last, usually dense, layer --- several layers can potentially be used --- specializes the RNN to its targeted task from this final latent layer.

A RNN is often trained to perform the next symbol prediction task\footnote{This task is often called \textit{language modelling} and the corresponding networks are denoted LM-RNN.}: given a prefix of a sequence, it outputs the probabilities for each symbol to be the next symbol of the sequence (a special symbol denoted $\rtimes$ [resp. $\ltimes$] is added to mark the start [resp. the end] of a sequence). Outputting the final symbol means that the input sequence is not a prefix of a longer sequence but instead a complete sequence.
Notice that it is easy to use such RNN to compute the probability given to a string $w=\sigma_1\sigma_2\ldots\sigma_n$: one only needs to compute
$$P(w) = P(\rtimes)P(\sigma_1|\rtimes)P(\sigma_2|\rtimes\sigma_1)\ldots P(\ltimes|\rtimes\sigma_1\sigma_2 \ldots\sigma_n)$$

\section{Algorithm}
\label{sec:algo}
In this section, we describe our approach: from an already trained LM-RNN, we fill a finite part of the Hankel matrix, then use a spectral approach to obtain a WA from the matrix. The following subsections detail the different steps of our proposition

\subsection{From Hankel to WA}
\label{subsec:H2WA}
The proof of Theorem~\ref{thm:finiterank} is constructive: it provides a way to generate a WA from its Hankel matrix $\mathcal{H}$. Moreover, the construction can be used on particular finite sub-blocks of this matrix: the ones defined by a complete and prefix-closed basis. Formally, a basis $\mathcal{B} = (\mathcal{P},\mathcal{S})$ is \textit{prefix-closed} iff for all $w\in \mathcal{P}$, all prefixes of $w$ are also elements of $\mathcal{P}$; $\mathcal{B}$ is \textit{complete} if the rank of the sub-block $H_\mathcal{B}$ is equal to the rank of $\mathcal{H}$.

Explicitly, from such sub-block $H_\mathcal{B}$ of $\mathcal{H}$ of rank $r$, one can compute a minimal WA 
using a rank factorization $PS=H_\mathcal{B}$, with $P\in\mathbb{R}^{p\times r}$, $S\in\mathbb{R}^{r\times s}$. Let us denote:
\begin{itemize}
    \item $H_\sigma$ the sub-block defined over $\mathcal{B}$ by $H_\sigma(u,v)=\mathcal{H}(u\sigma, v)$,
    \item $h_{\mathcal{P},\lambda}$ the $p$-dimensional vector with coordinates $h_{\mathcal{P},\lambda}(u)=\mathcal{H}(u,\lambda)$ ($h_{\mathcal{P},\lambda}$ is the first column of $H_\mathcal{B}$ if suffixes are given in lexicographic order),
    \item $h_{\lambda,\mathcal{S}}$ the $s$-dimensional vector with coordinates $h_{\lambda,\mathcal{S}}(v)=\mathcal{H}(\lambda,v)$ ($h_{\lambda,\mathcal{S}}$ is first line of $H_\mathcal{B}$ if prefixes are sorted in lexicographic order).
\end{itemize}
The WA $A=\langle \alpha_0, (M_\sigma)_{\sigma\in\Sigma}, \alpha_\infty\rangle$, with:
\begin{center}
$\alpha_0^\top=h_{\lambda,\mathcal{S}}^\top S^+$, $\alpha_\infty=P^+ h_{\mathcal{P},\lambda}$, and, for all $\sigma\in \Sigma$, $M_\sigma=P^+ H_\sigma S^+$,
\end{center}
is a minimal WA\footnote{As usual, $N^+$ denotes the Moore-Penrose pseudo-inverse~\citep{moore1920} 
of a matrix $N$.} whose Hankel matrix is exactly the initial one~\citep{balle14}.

This procedure is the core of the theoretically founded spectral learning algorithm~\citep{bailly2009,hsu2009,balle14}, where the content of the sub-blocks is estimated by counting the occurrences of strings in a learning sample. 
Contrary to that, the work presented here uses 
an already trained RNN to fill $H_\mathcal{B}$ and $H_\sigma$ on a carefully selected basis $\mathcal{B}$.

\subsection{Proposed Algorithm}
\label{subsec:prot}

Our algorithm is presented in synthetic form in Algorithm~1
. It can be broken down into three steps: from a LM-RNN model trained on some data, denoted $\mathcal{M}$, we first build a basis $\mathcal{B}$; 
second, we fill the required sub-blocks $H_\mathcal{B}$ and $(H_\sigma)_{\sigma\in\Sigma}$ with the values computed by the initial RNN $\mathcal{M}$; and third, we extract a WA from the Hankel matrix sub-blocks by low-rank factorisation of $H_\mathcal{B}$.

\begin{figure}[h]
    \centering
    \includegraphics[width=\textwidth]{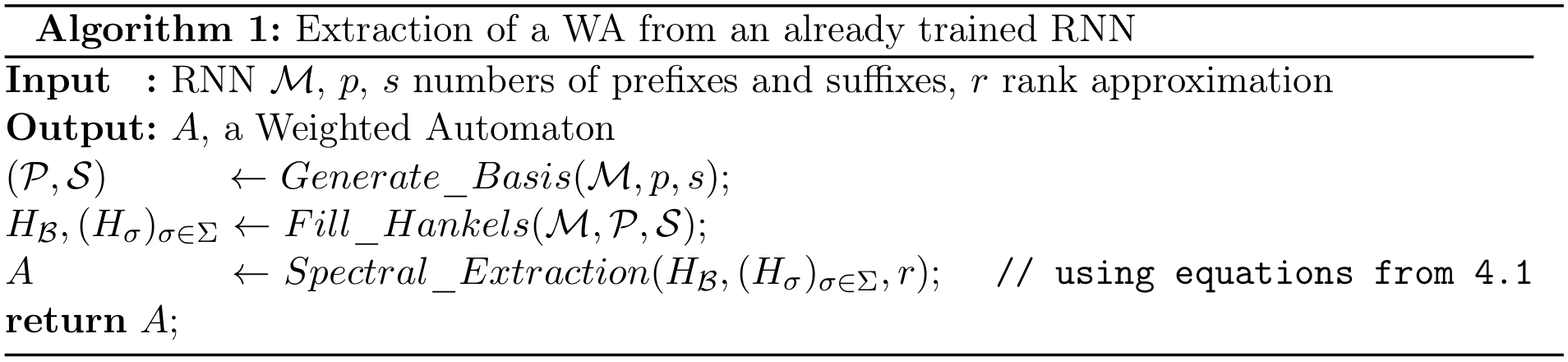}
    \label{algo1}
\end{figure}
      

\subsubsection{Generating basis}
\label{basis_strategies}
Choosing the right basis $\mathcal{B}=(\mathcal{P}, \mathcal{S})$ to define the right sub-block 
is an important task and different possibilities have been studied in the context of spectral learning~\citep{baillythesis,quat17}. 

A first idea is to implement $Generate\_Basis()$ by sampling from the uniform distribution on symbols with a maximum length parameter.
A second possibility, if the black box is a generative device like LM-RNN, is to use it to build a basis, for instance by recursively sampling a symbol from the next symbol distribution given by the network.

Though other ways of generating a basis can be designed, for instance by using training data if available, the experiments presented in this paper concentrate on these two approaches, in section~\ref{sec:basis}.

Once a string is obtained, we add all its prefixes to $\mathcal{P}$ (to be prefix-closed) and all its suffixes to $\mathcal{S}$. The process is reiterated until $|\mathcal{P}|\geq p$. If needed, the set of suffixes is then completed in the same way until $|\mathcal{S}|\geq s$.

\subsubsection{Filling Hankel matrix}
Once we have a basis $\mathcal{B}$, the procedure $Fill\_Hankels()$ uses the black box RNN to compute the content of the sub-blocks $H_\mathcal{B}$ and $(H_\sigma)_{\sigma\in\Sigma}$: it queries each string made of selected prefix and suffix to the black box and fill the corresponding cells in the sub-blocks of the Hankel matrix with its answer. This procedure is the core step for knowledge distillation in our algorithm, enabling the knowledge from RNN outputs to be directly transferred into the Hankel matrix from which we can extract WA by spectral extraction.

\subsubsection{Spectral extraction}

Finally, a rank factorization $H_\mathcal{B}=PS$ for a certain rank parameter $r$, has to be obtained: the function $Spectral\_Extraction()$ performs a Singular Value Decomposition on $H_\mathcal{B}$ and truncates the result to obtain the needed rank factorization (see \citep{balle14} for details). It then generates a WA using the formulas described in Section~\ref{subsec:H2WA}. Section~\ref{sec:rank_impact} elaborates on the top highest singular values and their relations to number of states of WA.

\section{Experimental framework}
\label{sec:framework}
\subsection{Data}
When one wants to systematically studies an algorithm working on sequences, there does not exist a wide number of benchmarks not specific to a particular field of application and thus that does not require large and impacting pre-treatment (for instance, NLP corpus require important and result influencing pre-processing steps, \textit{e.g.}, POS-tagging, lemmatisation, etc.). 
As a consequence, researchers tend to compare their work using well-known problems from the early times of machine learning, mostly the Reber's grammar~\citep{rebe67} and the Tomita problems~\citep{tomi82} (see for instance the work of \citet{weis18}, \citet{moze18}, or \citet{Wang19} for recent examples of uses of these data).

Though of great interest, these two sets of problems are too specific and neither large nor covering enough to convincingly support claims about an algorithm only tested on them. Fortunately, 2 benchmarks have been recently made available that allow an as exhaustive as possible study of the type of algorithms we are interested in. Each of them were made freely available in the context of on-line challenges : the PAutomaC~\citep{pautomac} and the SPiCe~\citep{spice} competitions. 

\subsubsection{PAutomaC data}
The first sets of data we experiment on are the ones of the competition PAutomaC whose goal was to learn distributions from multi-sets of sequences of symbols of varying length. The competition provides 48 artificial problems generated by randomly generated finite state machines : Deterministic Probabilistic Finite Automata (DPFA), non-deterministic Probabilistic Finite Automata (PFA), and Hidden Markov's Models (HMM). Large range of machine sizes, alphabet sizes, and sparsity indicators values are covered by the different instances and the competition results proved it is covering a wide range of difficulty levels.

Though all of these models are strictly less expressive than stochastic WA~\citep{deni08}, it is relevant to test our distillation procedure on this wide variety of problems.  Moreover, having access to the true target model gives an interesting point of comparison with the extracted automaton.

For each of these 48 problems, we have access to a training set (20\,000 or 100\,000 sequences), that we only use to train the RNN.


\subsubsection{SPiCe data}
The SPiCe benchmark is made of 15 problems covering a large variety of contexts, from various natural languages processing data to software engineering data, including bioinformatics and well-chosen synthetic ones. 

The aim of the competition was to learn a model from each of the available training samples and then to propose a ranking of the top 5 next potential symbols to given prefixes. 

These test sets made of prefixes are not what we need for our task, since they do not provide complete information on whole sequences. We thus split randomly the available learning samples into a training and a test sample, keeping 20\% for the test.

We also choose to not handle problem 11 of SPiCe, due to its large vocabulary (6\,722 symbols) that would require long computation times for each of the parameters we want to study. 

\subsection{RNN training}
\label{subsec:train}
We base the architecture on the work of \citet{shibataSpice}, who won the SPiCe competition, itself inspired by \citet{suts14}. The architecture is quite simple: it is composed of an initial embedding layer (with $3*|\Sigma|$ neurons), two GRU (or LSTM) layers, one or two dense layers, followed by a final dense layer with \texttt{softmax} activation composed of $|\Sigma|+1$ neurons. The networks are trained using the back-propagation through time algorithm~\citep{rume86,werb88} with the categorial cross-entropy loss function~\citep{rubi04} on the distribution over next symbols for all prefixes. 


Given this framework, we considered several hyper-parameters to tune. First, the number of neurons in the recurrent layers and the following dense layer: 
we tested a number of neurons between $30$ and $400$, the first following dense layer uses half of it, the second dense layer is set as the size of the input embedding layer, or ignored. We also compared \texttt{tanh} and \texttt{RELu} activations. We trained our networks during 150 epochs. We do witness expected over-fitting before this limit: after a phase in which the loss on a validation set decreases, it starts to rise. This confirms that this number is adequate as maximum iterations. 

We finally selected one single configuration for each problem that scores the best categorical cross-entropy on the validation set in average since we did not observe significant variations among tested configurations. The  results detailed below are obtained from the following configuration: an initial embedding layer, then two GRU (or LSTM) layers build with 400 neurons and \texttt{tanh} activations; followed by one hidden dense layer with 200 neurons and \texttt{RELu} activations, before the final output layer. We trained models using Adam optimizer for 150 epochs with batch size of 512, a learning rate of 0.001, and defaults beta values.

The evaluation of this protocol shows good learning results but it is also clear that better results could be obtained tuning the RNN on the chosen data independently for each problem. We decide to not push to its limits this learning part since it is not central in this work. Moreover, having RNN with various learning qualities is interesting from the standpoint of the evaluation of the WA extraction: we expect the WA to be as good --- or, thus, as bad --- as the RNN. We report in appendix the validation loss for each RNN model.

\subsection{Metrics}
\label{sec:metrics}

\subsubsection{Evaluation samples}
\label{subsec:sample}
In order to not bias the evaluation by picking a particular type of data, we use two evaluation sets to compute the different metrics chosen to evaluate the quality of the WA extraction. 
$S_{\text{Test}}$ is the test set from the competition: directly the one used for PAutomaC, and the one coming from a random partitioning of the learning sample for SPiCe.
$S_{\text{RNN}}$ consists in generated sequences sampled using the learned RNN. Both $S_{\text{Test}}$ and $S_{\text{RNN}}$ contain 1\,000 sequences. 

This way we test the quality of our distillation process using the distribution of initial learning task and the one of the RNN we are targeting. Notice that $S_{\text{RNN}}$ is more relevant for our task: it contains sequences enjoying high probability in the model we are distilling from. $S_{\text{Test}}$ has this property only if the RNN achieves good learning results.

We evaluate the quality of the approximation obtained by
the extraction using two 
metrics.

\subsubsection{Normalized Discounted Cumulative Gain}
The second 
metrics is 
the Normalized Discounted Cumulative Gain (NDCG, used in SPiCe). This metric is given by: for a prefix $w$ of a sequence in an evaluation set, 
$$\mbox{NDCG}_n(w, \widehat{\sigma}_1,...,\widehat{\sigma}_n) = \frac{ \sum_{k=0}^{n} \frac{P_{\text{RNN}}(\widehat{\sigma}_k|w)}{log(k+1)}}{ \sum_{k=0}^{n} \frac{P_{\text{RNN}}(\sigma_k|w)}{log(k+1)}}$$ 
where $\sigma_k$ is the k$^{\textrm{th}}$ most likely next symbol of $w$ following $P_{\text{RNN}}$ and
$\widehat{\sigma}_k$ is the k$^{\textrm{th}}$ most likely next symbol of $w$ following $P_{\text{WA}}$.

The NDCG$_n$ score of an extraction is the sum of NDCG$_n$ on each prefix of a sequence in the evaluation set, normalized by the number of these prefixes in the test set. In our experiment, we focus on NDCG$_5$ scores: this value for $n$ is the one used for the SPiCe competition and it is actually the largest computable on all datasets, since the smallest alphabet size is $4$ (to which a symbol marking the end of a sequence is added).

\subsubsection{Word Error Rate for Distillation}
Word Error Rate (WER) is a classical quality measurement in the field of Natural Language Processing. In its simplest and most used form, it is computed from a set of sequences by counting on all prefixes the number of times the most probable next symbol in the model is different to the actual next symbol in the sequence. This total number is then divided by the total number of prefixes in order to have a value between 0 and 1, the closer to 0 the better. 

Given that we have access to two evaluation samples (see Section~\ref{subsec:sample}), a simple idea would be to compute the classical WER on these sets. However, in the context of distillation we can design a different metric, inspired by WER, that will better suit the evaluation of our goal. 

The idea is to use the black-box as ground truth: instead of looking at the next symbol in the dataset, one can compare the most probable next symbol in the distilled model and the most probable one in the model from which the distillation occurred. We call this metric Word Error Rate for Distillation (WER-D). 
Formally:
$$\textrm{WER-D}_S(\textrm{RNN}, \textrm       {WA}) = \frac{1}{|\textrm{pref}(S)|}\sum_{w\in \textrm{pref}(S)} \mathds{1}_{\sigma_1(w)\neq\widehat{\sigma}_1(w)}  $$

\noindent
where $S$ is an evaluation set and, for any prefix $w$, $\sigma_1(w)$ is the most probable next symbol in the RNN and $\widehat{\sigma}_1(w)$ the one in the WA.

WER-D is thus the frequency observed on a test set of the disagreement between the two models on the next most probable symbol. 
Though this metric has not been described in any article yet, as far as we know, it is the one used by \citet{weis19}, as their available code shows.

\subsection{Experimental setting}
\label{sec:expe}

We describe here the protocol we followed to validate our method for RNN distillation. We also detail the experiments we conducted and  hyper-parameters we explored for WA extraction. 

\subsubsection{Global procedure}
We first trained RNN with the different architecture parameters detailed in previous section on the 62 benchmark problems of the PAutomaC and SPiCe competitions.
We then keep the best RNN in term of categorical cross-entropy for each problem and consider it as the black-box of the problem.\\

\noindent
We then extract WA from this RNN following the algorithm described in Section~\ref{sec:algo}. We test both already described approaches for basis generation (uniform distribution on symbols and RNN sampling) with a wide range of parameter values:
\begin{itemize}
    \item the size of the basis varies from 100 prefixes and suffixes up to 3\,000,
    \item the rank for the Hankel matrix factorization is also taken between 1 and the full rank of Hankel matrix (up to 3\,000).
\end{itemize}


\subsubsection{Hyper-parameters for extraction}
We compare different hyper-parameter values for the extraction algorithm: size $p$ and $s$ of the basis is varying between $(100,100)$ and $(3000,3000)$.
We tested different rank values from 1 to the number of prefixes. On Spice datasets, we considered all rank values below 15 and 25 others logarithmically picked in the other possible values. On PAutomaC datasets, we tested 20 ranks below 50 then 5 ranks logarithmically chosen until the number of prefixes.
We also compare two possible strategies for sampling the basis to build the Hankel matrix, as discussed in section~\ref{basis_strategies}.

Some values of the size of the basis exceed what is reasonably computable on our limited computation capacities for some datasets: few problems did not finished the extraction with $(3000, 3000)$ for basis size. However, as it is shown in Section~\ref{sec:results}, small values already allow the extracted WA to be a great approximation of the RNN. 

\subsubsection{Implementation details}
All experiments are conducted using the Scikit-SpLearn toolbox~\citep{splearn} to handle WA and their extraction, and the PyTorch framework~\citep{pasz17} for RNN learning and querying. A (simplified) version of our code can be find on \href{http://pageperso.lif.univ-mrs.fr/~remi.eyraud/RNN2WA-simple.tar.gz}{first author's website}.

\section{Results}
\label{sec:results}

\subsection{Overall quality extraction}

\begin{figure}[htbp!]
\begin{center}
\includegraphics[width=0.4\textwidth]{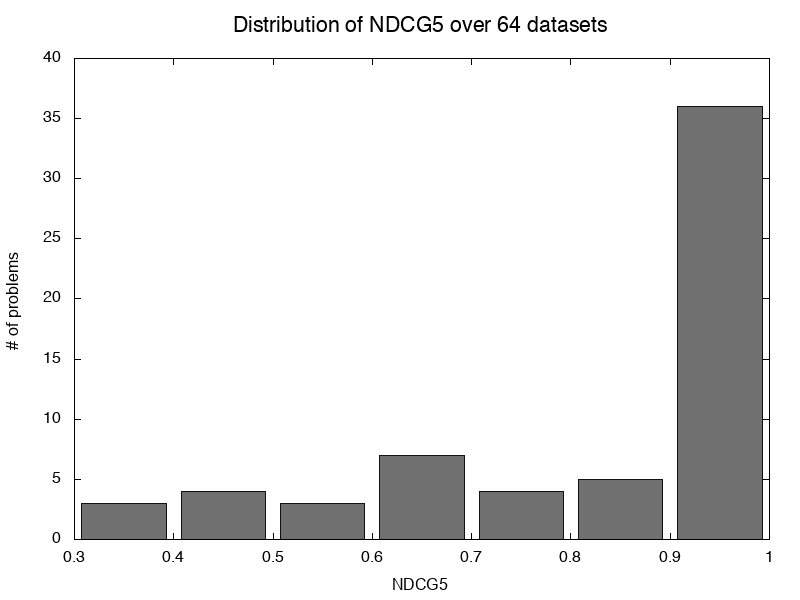} 
\includegraphics[width=0.4\textwidth]{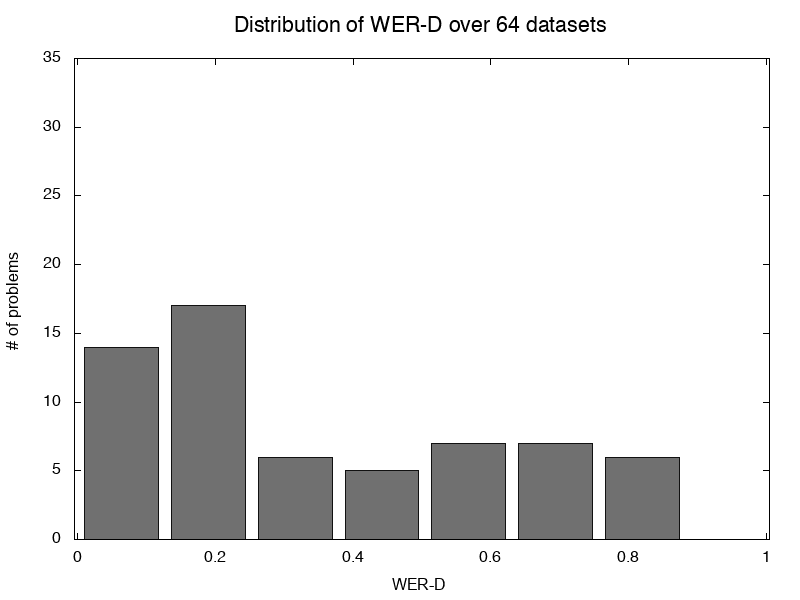}
\caption{Overall statistics on PAutomaC and SPiCe computed on $S_{\text{RNN}}$}. 
\label{fig:best_of_bests_rnnw}
\end{center}
\end{figure}

This section globally summarizes experiments to assess the quality of extracted WA on the 62 datasets from PAutomaC and SPiCe competitions. Figure~\ref{fig:best_of_bests_rnnw} shows the best obtained quality on $S_{\text{RNN}}$ through the various hyper-parameters and sampling strategies we tested. It highlights that a large part of NDCG scores are higher than 0.9 while a majority of WER-D scores are below 0.2. This shows clearly the great ability of our algorithm to reproduce the behaviour of a LM-RNN. 

\noindent
In the following, we provide a finer analysis related to PAutomaC and SPiCe problems.

\subsection{Basis generation}
\label{sec:basis}
The impact of the basis for the quality of the task of WA extraction is of crucial interest. We thus carefully study the difference between the two ways to generate the basis: by sampling uniformly at random using a parameter for maximum length, and using the RNN to be distilled.


\begin{figure}[h]
    \hspace{-1cm}
    \includegraphics[width=1.2\textwidth]{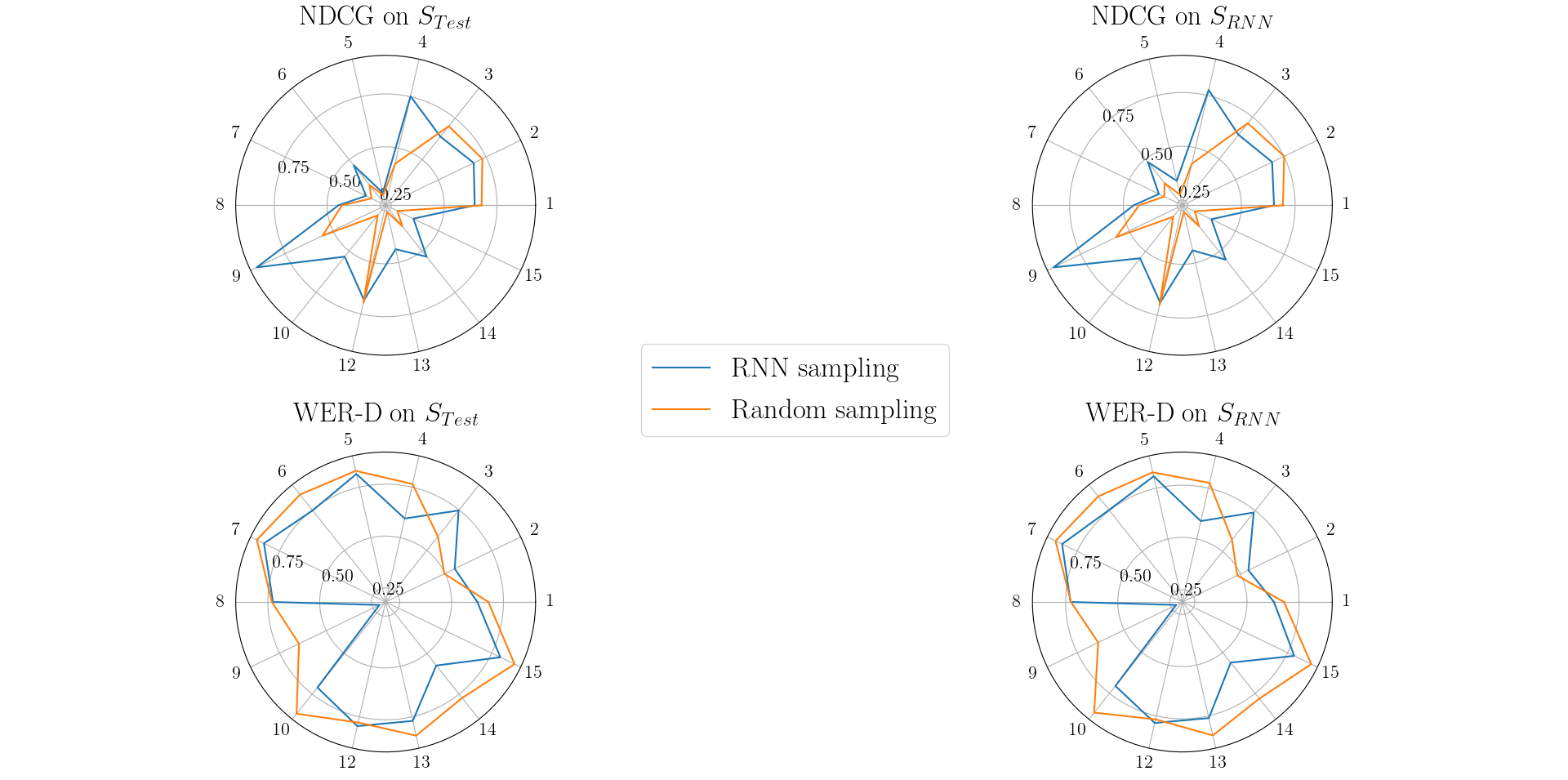}
    \caption{Influence of the basis sampling method on the quality of the extraction for the SPiCe problems.}
    \label{fig:basis_sampling_spice}
\end{figure}

Figure~\ref{fig:basis_sampling_spice} 
shows the best values obtained for each chosen metric on the evaluation datasets for the two methods. 
In this figure, each radius of a polar plot corresponds to one problem whose corresponding number is given outside of the circle. 
It details only the results on the difficult problems of the SPiCe competition: the average results on the PAutomaC problems are summed up in Table~\ref{tab:basis_pautomac}. The difference on these datasets is even more significant.


\begin{table}[h]
    \centering
    \begin{tabular}{c|c|c|c|c|}
    \cline{2-5}     
         & \multicolumn{2}{c|}{WER-D $\downarrow$} & \multicolumn{2}{c|}{NDCG $\uparrow$}\\
    \cline{2-5}
         &  on $S_{\text{Test}}$ & on $S_{\text{RNN}}$ & on $S_{\text{Test}}$ & on $S_{\text{RNN}}$ \\
    \hline
    \multicolumn{1}{|c|}{random basis} &  0.334 & 0.336  & 0.851  & 0.848 \\
    \hline
    \multicolumn{1}{|c|}{RNN sampled basis} &   \textbf{0.254}  &  \textbf{0.259}   &    \textbf{0.921} &   \textbf{0.915} \\
    \hline
    \end{tabular}
    \caption{Comparison of the average metric values on the PAutomaC problems.}
    \label{tab:basis_pautomac}
\end{table}
Recalling that best possible NDCG is 1 while best possible WER-D is 0, these elements indicate a clear trend: the generation of the prefixes and suffixes of the basis based on sampling from RNN allows better WA to be extracted.

\noindent
Therefore, unless it is specified differently, all the following results concern WA extracted using a basis obtained by sampling the RNN.

\subsection{Impact of the size of the basis}

\begin{figure}[h]
    \centering
    \includegraphics[width=\textwidth]{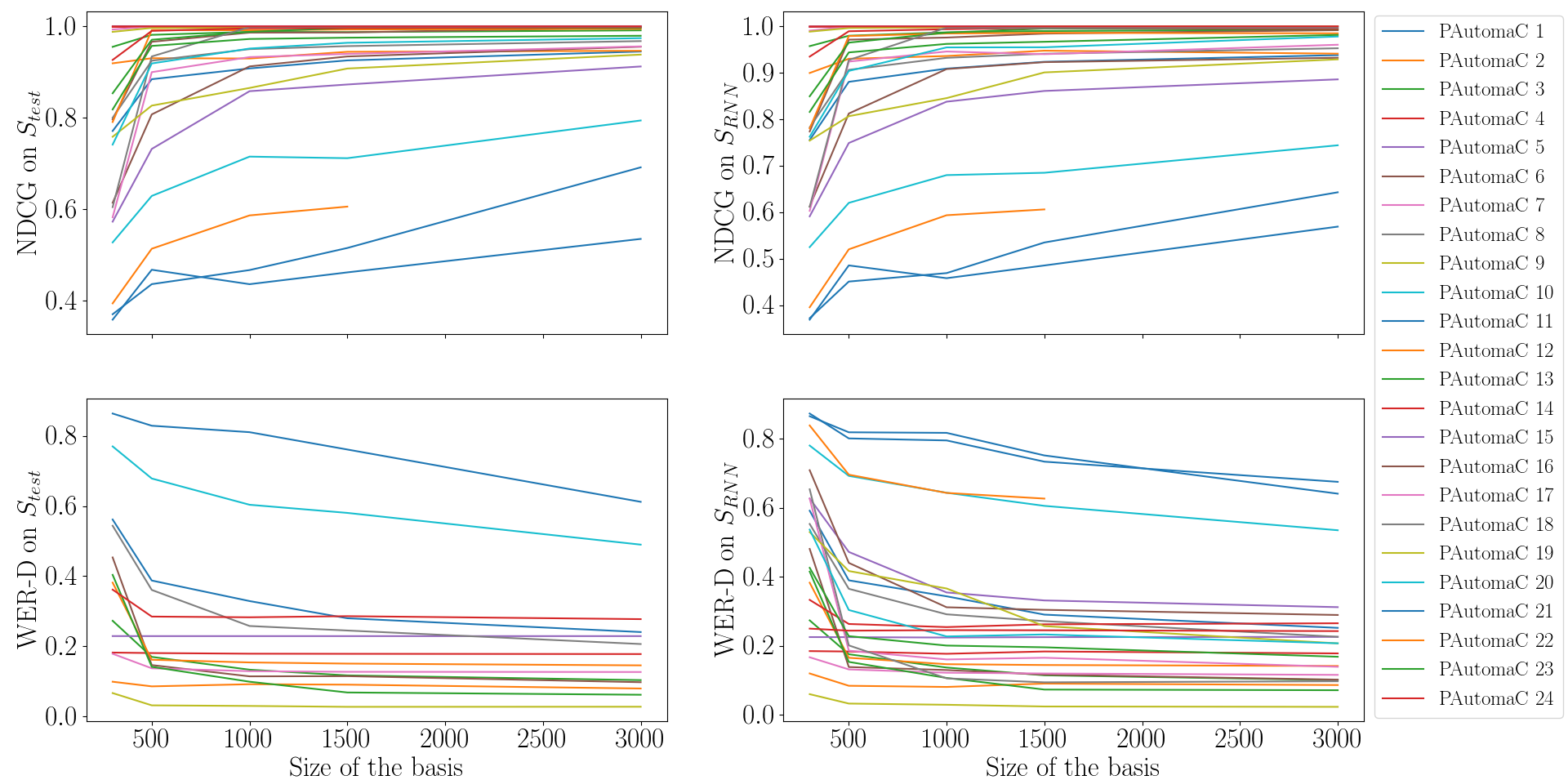}
    \caption{Best obtained metric values on the 24$^{\textrm{th}}$ first problems of PAutomaC by the extraction using a basis sampled from the RNN. Each plot corresponds to one metric on one evaluation sample. }
    \label{fig:basis1}
\end{figure}

\begin{figure}[h]
    \centering
    \includegraphics[width=\textwidth]{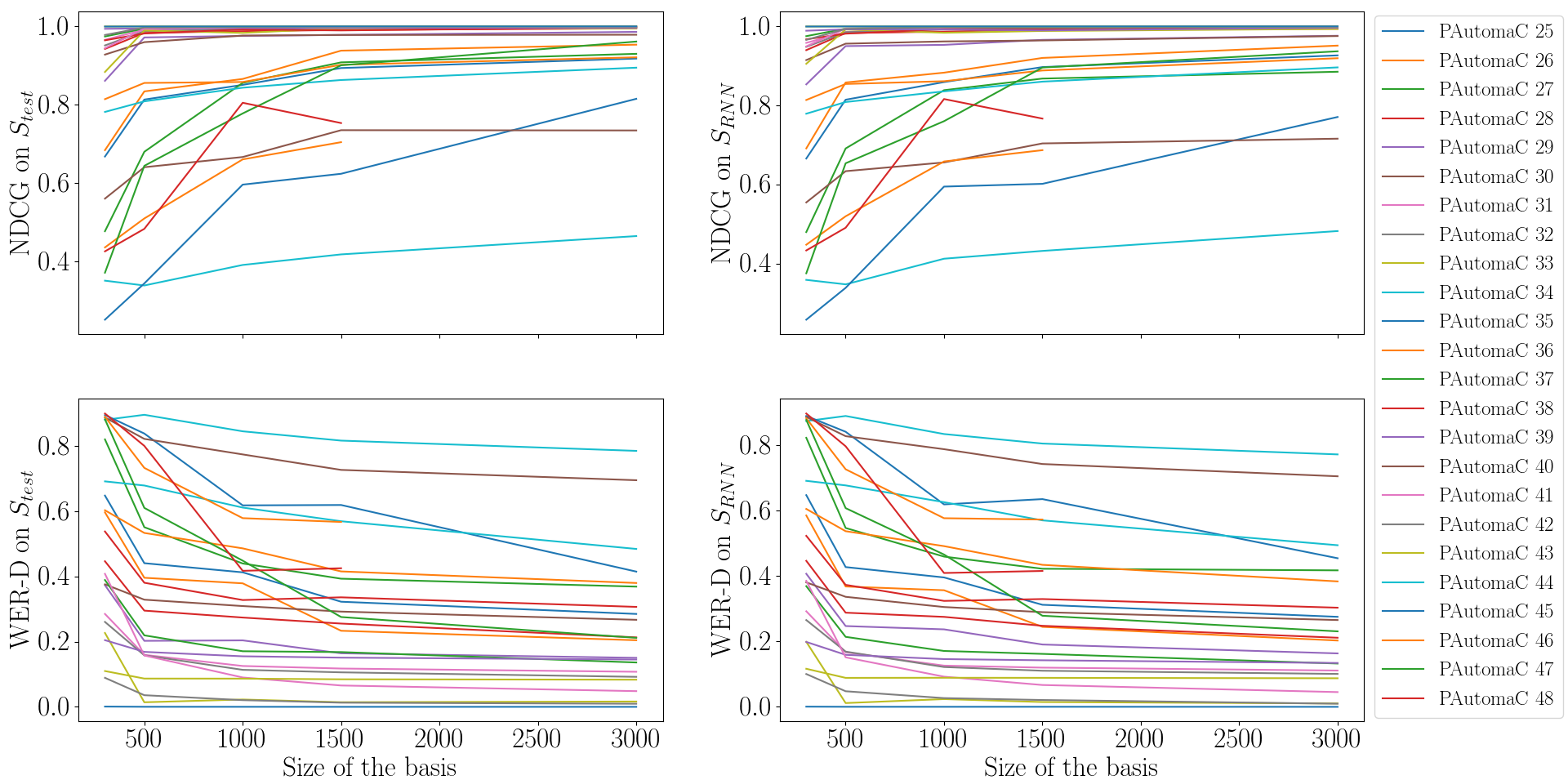}
    \caption{Best obtained metric values on the 24$^{\textrm{th}}$ last problems of PAutomaC }
    \label{fig:basis2}
\end{figure}

Figures~\ref{fig:basis1} and \ref{fig:basis2} show the evolution of the NDCG and WER-D on the two evaluation sets when the size of the basis varies, on the 48 PAutomaC benchmarks. A global trend tends to occur: the larger the size of the basis is, the better the extracted WA seems to be.

This result is confirmed on SPiCe: all but one problems obtained the best quality measures with the largest basis size our experimental setup allowed us to test. 

\noindent
Though it is worth to be verified, this observation is not surprising: having larger basis means having access to more information about the RNN to distill, which is likely to be useful for the process.

It is worth noticing that, if for most PAutomaC problems we managed to run experiments up to 3000 times 3000 bases, it is not the case for SPiCe: the hardness of these tasks, in particular the large size of the vocabulary, narrowed what was possible on our (limited) computing server. Concretely, only 3 of the 14$^{\textrm{th}}$ experiments finished with 3000$\times$3000 basis (Problems~2, 3, and 9), while two of them did not even allowed a 1500$\times$1500 to run till the end (Problems~6 and 13).  


\subsection{Comparison with n-grams}
We compared the quality of our distilled WA with a classical baseline: n-gram~\citep{shan48}. This knowingly powerful model requires a sample of data to be learned: for each problem we randomly generate a multi-set of sequences using the corresponding trained RNN until the sum of the lengths of these strings exceeds 10 millions. This dataset is then used for training, with the window size of the model varying between 2 and 20.


\begin{figure}[h]
    \hspace{-1cm}
    \includegraphics[width=1.2\textwidth]{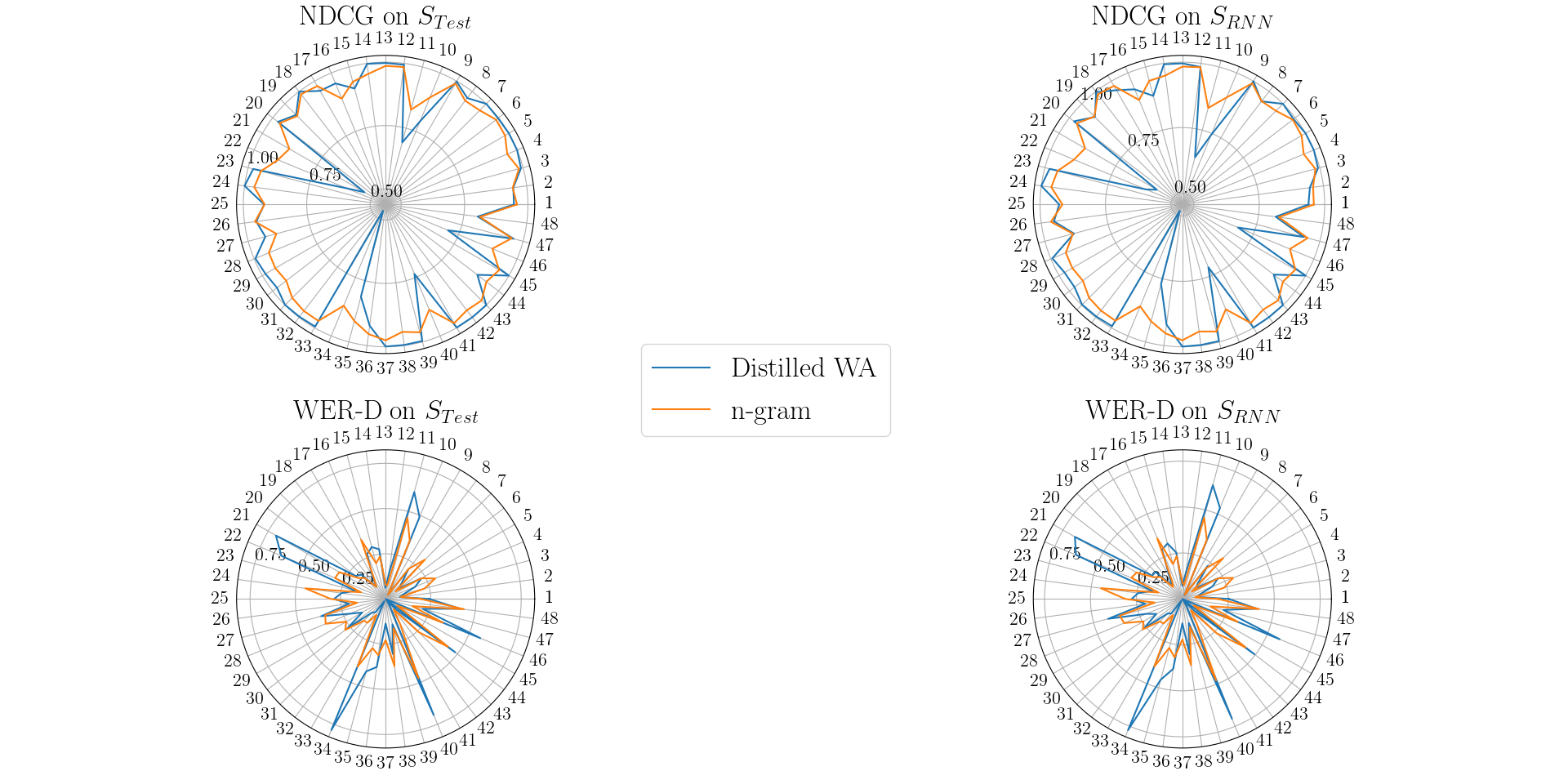}
    \caption{Quality comparison between the best obtained results of n-grams and distilled WA for the PAutomaC problems.}
    \label{fig:n-gram_pautomac}
\end{figure}

Figure~\ref{fig:n-gram_pautomac} show the comparison on the 2 evaluation sets, for the two metrics on the PAutomaC problems. Globally, the two models seems to offer comparable quality: the best distilled WA is doing slightly better than the n-gram on a majority of datasets, while facing some difficulties on few of them (6 over 48). 

We investigated these troublesome problems to see if the characteristics of the models used to generate them can explain these lack of performance. We did not find any obvious reasons: the 3 types of machines (DPFA, PFA, HMM) are present and the values of the two sparsity parameters cover the whole range available. A common feature, though shared by problems on which the distilled WA are doing great, is that the models that generated these datasets have both a large alphabet and a big number of states. This may indicate that the Hankel matrix is not large enough, that is, it is not complete: increasing the number of prefixes and suffixes may allow better results.

\begin{figure}[htbp!]
    \hspace{-1cm}
    \includegraphics[width=1.2\textwidth]{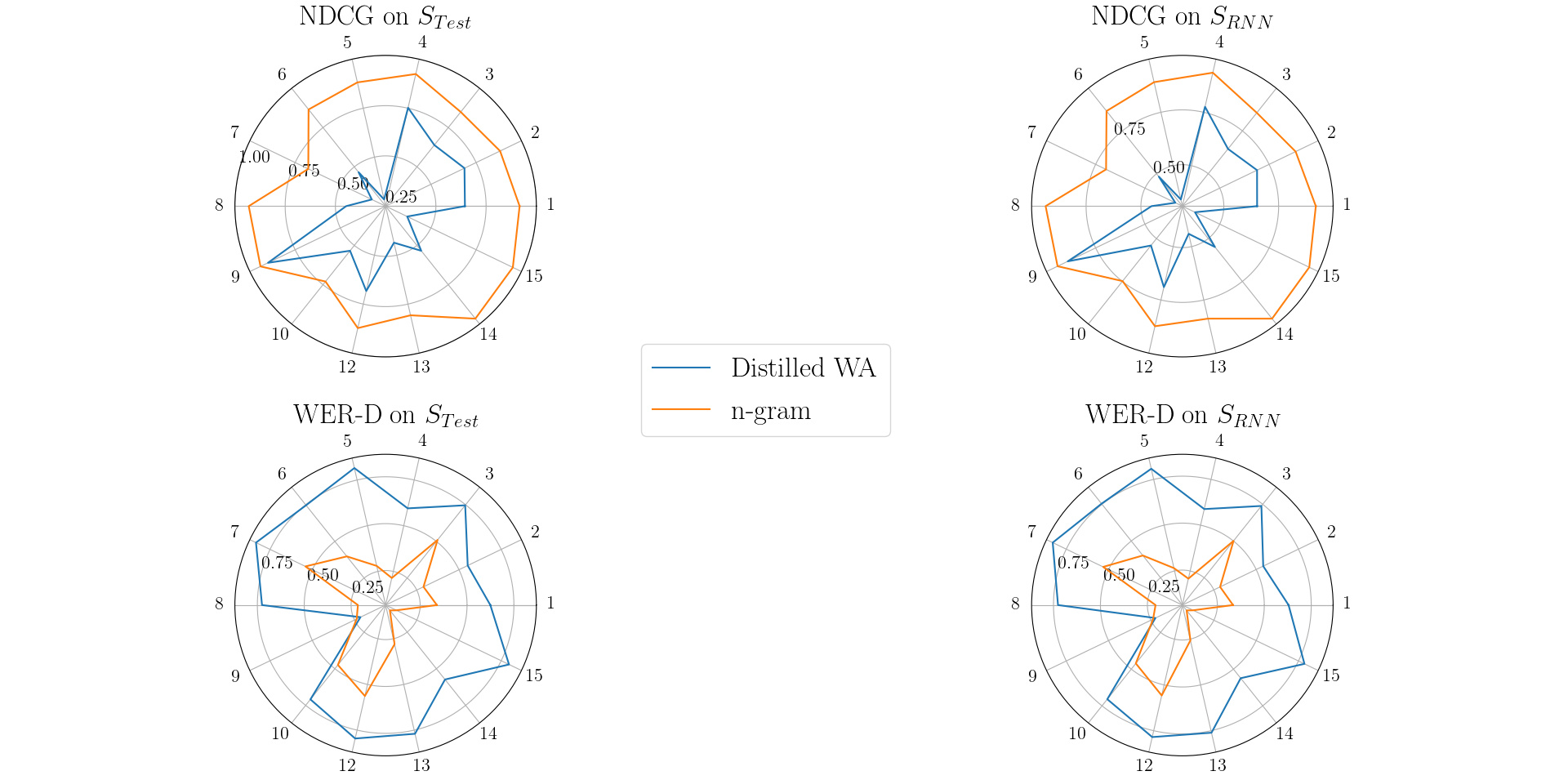}
    \caption{Quality comparison between the best obtained results of n-grams and distilled WA for the SPiCe problems.}
    \label{fig:n-gram_spice}
\end{figure}

Figure~\ref{fig:n-gram_spice} shows the same curves for the more challenging problems of the SPiCe competition. As already noticed by \citet{weis19}, the n-gram is a (very) strong baseline on these datasets. On some problems our algorithm obtains close results but the overall difference is notable, in particular when the size of the vocabulary is important. 

\subsection{Impact of the rank}
\label{sec:rank_impact}
\begin{figure}[htbp!]
 \begin{center}
\includegraphics[width=0.23\textwidth]{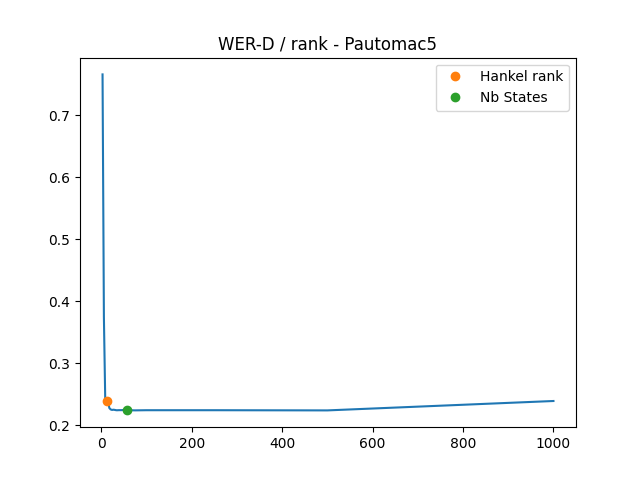}
\includegraphics[width=0.23\textwidth]{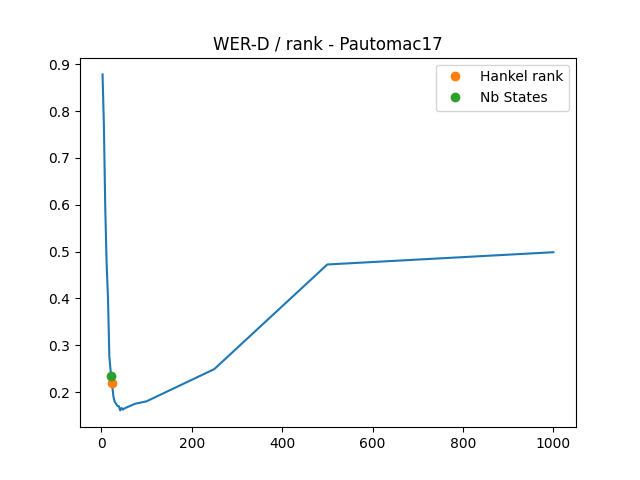}
\includegraphics[width=0.23\textwidth]{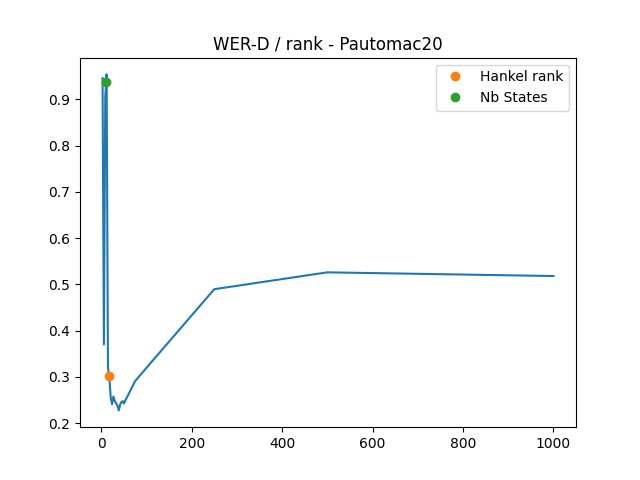}
\includegraphics[width=0.23\textwidth]{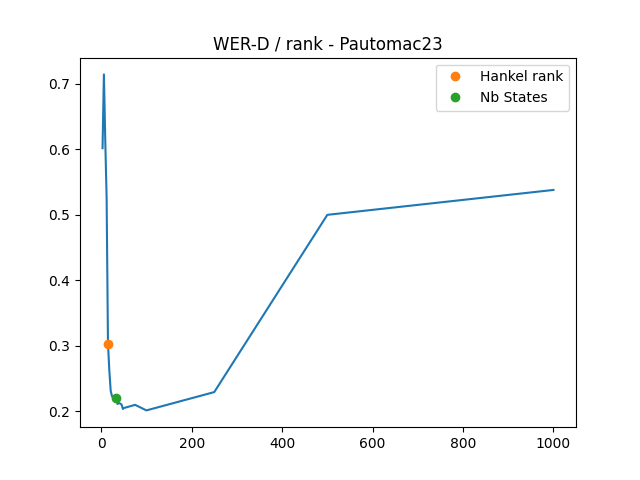}
\includegraphics[width=0.23\textwidth]{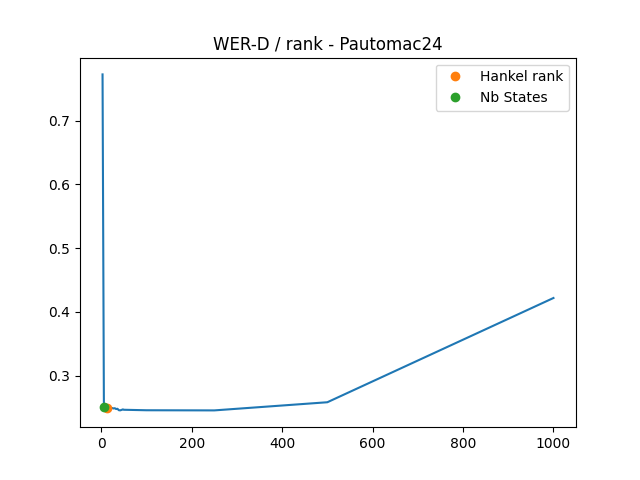}
\includegraphics[width=0.23\textwidth]{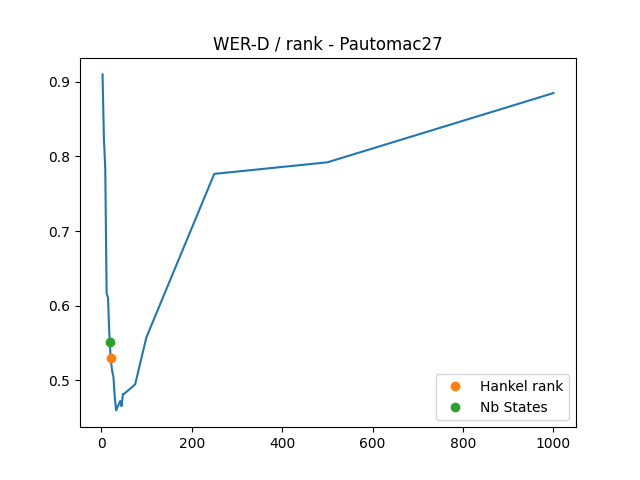}
\includegraphics[width=0.23\textwidth]{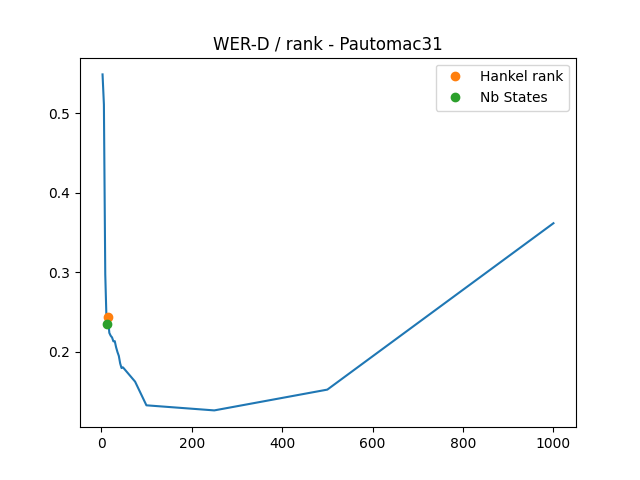}
\includegraphics[width=0.23\textwidth]{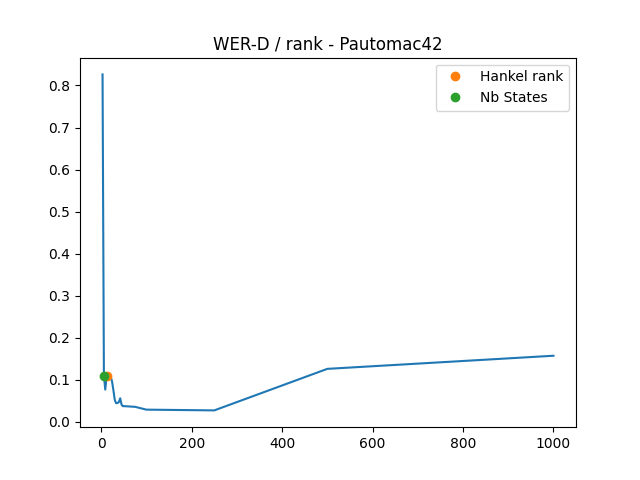}
\includegraphics[width=0.23\textwidth]{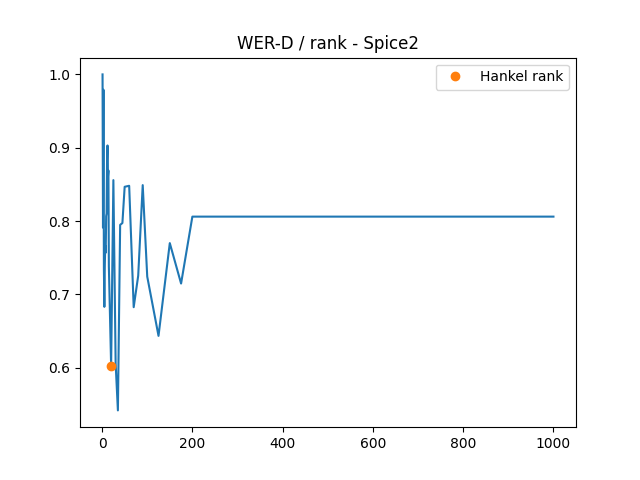}
\includegraphics[width=0.23\textwidth]{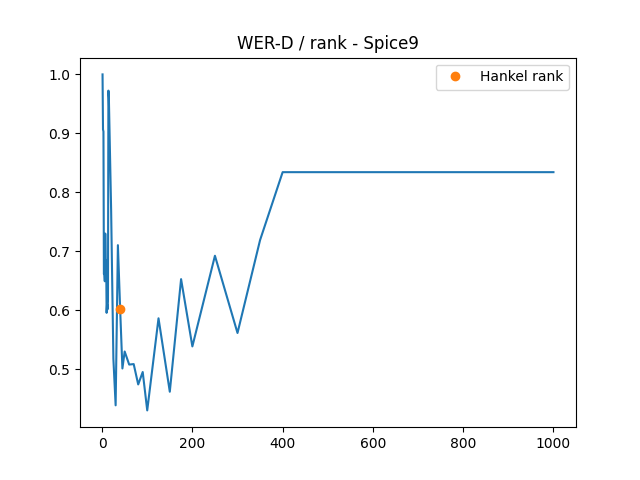}
\includegraphics[width=0.23\textwidth]{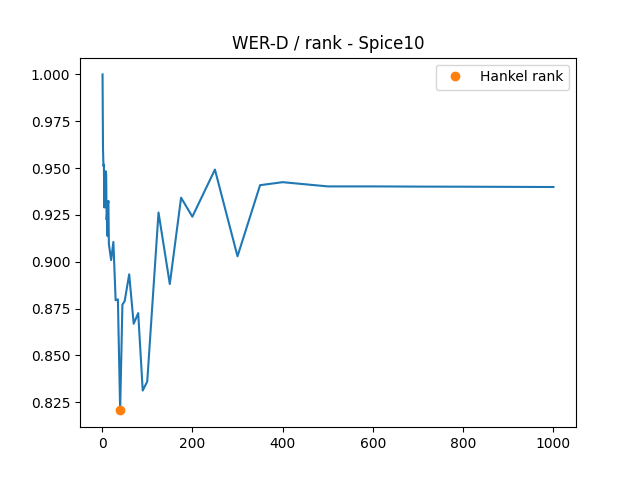}
\includegraphics[width=0.23\textwidth]{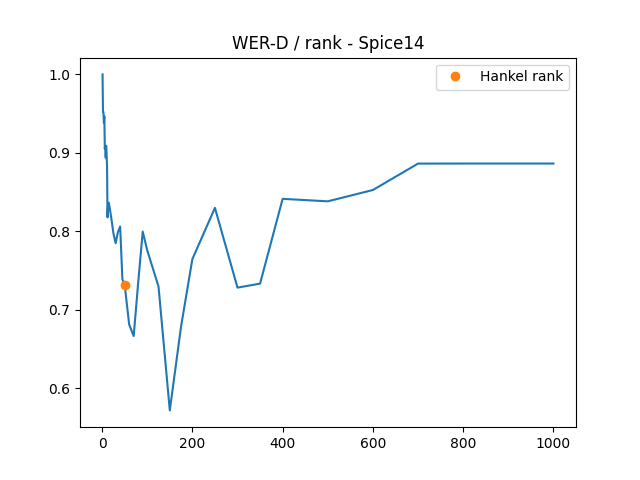}

\caption{Evolution of the quality in term of WER-D on $S_{RNN}$ with the value of the rank hyper-parameter on some PAutomaC and SPiCe problems (basis size is 1000 prefixes and suffixes).}
\label{fig:ranks}
\end{center}
\end{figure}

Globally, in our experiments, the quality of the extracted WA quickly increases at first when the rank grows, then, when a relatively small rank is reached, 
the quality decreases and often stabilizes far from the optimum after a while. This behaviour is exemplifies on some datasets in Figure~\ref{fig:ranks} that shows the evolution of WER-D when the rank hyper-parameter increases (same plots for all the problems are given in Figures~\ref{fig:ranks-pautomac-1-40} and \ref{fig:ranks-pautomac-spice} in the Appendix).

A first element of discussion is that the rank allowing the best result reinforces the interest of our distillation procedure. Indeed, the number of parameters of a WA is $|Q|^2|\Sigma|+2|Q|$ where $Q$ is the number of states, that is, the rank in our experiments, and $\Sigma$ the alphabet: with the best overall rank average being around 175, the average number of parameters in the extracted WA is $30\,625|\Sigma|+ 350$ while the used RNN have already $2$ million parameters just for the recurrent part (the number of parameters of the embedding layer is quadratic in the size of the alphabet and it is linear on $|\Sigma|$ for the final dense layer).

\medskip
Investigating the reasons behind the relatively low-rank observation, we witness an interesting behavior of the singular values of the Hankel matrix built from the RNN. Their fast decrease, followed by a long plateau of very small values, indicates that the matrix is almost enjoying a finite number of non null singular values (the singular values for all problems are given in Figures~\ref{fig:svds} and \ref{fig:svds2} in the Appendix). Indeed, first values are several order of magnitude larger than the following ones. 

By computing the gradient of the singular values curves, and applying the same chosen threshold on them, we automatically identify for each problem a number of significant singular values\footnote{Though we witness small variations when the size of the bases increases, it is remarkably stable.}. By a slight abuse of language, we called this number the \textit{Hankel rank}. It is shown in orange in the previously introduced figures and its numerical value.

An important observation is that the evaluation metrics at the Hankel ranks are close to the optimum. The last two columns of Tables~\ref{tab:pautomac} and \ref{tab:spice} of the Appendix provide the difference of the metrics at these ranks and at the ranks allowing the best results. This shows that even when the best rank is larger than the Hankel rank -- we do witness an overall increase of these best ranks when the size of the basis grows -- the improvement is not significant.




 This is of great interest for the understanding of the RNN behavior. Indeed, Theorem~\ref{thm:finiterank} states that functions whose Hankel matrix is of finite rank, that is, with a finite number of non null singular values, are exactly the ones computable by WA. Our discovery thus tends to show that our trained RNN are computing distributions almost corresponding to finite state machines, even on data that do not correspond to that kind of models, like the ones of SPiCe. 


On data sets knowingly generated by finite state machines, as in PAutomaC, our study can be refined. Indeed, the green dots in the PAutomaC plot of Figures~\ref{fig:ranks}, \ref{fig:ranks-pautomac-1-40}, and \ref{fig:ranks-pautomac-spice}, indicates the number of states of the original finite state machine. 
We obviously do not aim to retrieve the exact amount of states since models used for generating the PAutomaC data may require more states than WA to represent the same distribution\footnote{HMM and their expressively equivalent representations of PFA may required more states than WA for the same distribution~\citep{deni08}. DPFA are well-known to need more states than PFA to represent the same stochastic language.}.
We however notice that our framework is remarkably able to get close to the correct number of states.

\medskip
This provides important understandings both on the quality of the learned RNN -- as the number of significant singular values is close to the number of states of the target, validating the quality of the learning phase -- and of our distillation process -- as the quality of the extracted WA with Hankel rank states is almost always comparable to the optimum.

Finally, as stated in different theoretical works summarized in Section~\ref{sec:previous}, finite precision and time GRU RNN are expected to behave like finite state machines (at least asymptotically), and our empirical conclusions tend to validate this point. This is why we re-run our experiments using LSTM cells instead of GRU ones: we observed the exact same behaviour despite the fact that the theoretical studies indicates a greater expressivity power for this type of RNN. We think this finding is an important outcome of our study.

\section{Conclusion}
\label{sec:conclu} 
This article presents a spectral approach to extract weighted automata from recurrent neural networks already trained on a language modelling task. The proposed algorithm uses the RNN as an oracle, querying it to know values it assigned to carefully selected data, and then produces the automaton using a singular value decomposition of the Hankel matrix, a mathematical object it uses to store the RNN answers.
Importantly, the algorithm does not require access to the inner representation of the networks: this allows it to work on any type of black box that computes a real value from an input sequence -- like models used for sentiment analysis from text~\citep{chau19} -- or that can be straightforwardly used for this task -- like language modelling RNN. 
This point is an important advantage of the proposed method compared to existing ones.

Our experiments on 62 different datasets show the overall quality of the extraction in the sense that the obtained WA represent good approximations of the distributions of the RNN. This validates our distillation process: WA computing linear functions, our algorithm provides linear, and thus computably efficient, models from non linear ones. This is emphasized by the fact that even the small extracted WA are already acceptable approximations, while requiring far less parameters.

Moreover, our process allows a deeper understanding of the underlying decision process of the trained RNN: the quick decrease of the singular values of the Hankel matrices indicates that these RNN are computing (approximations of) rational series, that is, their behaviour is the one of a linear finite state machine. This peek into the internal and opaque functioning of RNN allows to raise an fundamental question : to which extend RNN trained on data are biased towards functions of the lowest level of the Chomsky hierarchy?

Though interpretability of automata is well-established, partly but not entirely due to the existence of a graphical representation (see \citet{hamm16} for a detailed discussion), one can argue that the non-deterministic nature of our WA, and the potential presence of negative weights, diminish the interest of the proposed work toward explainability. If this critic is understandable, the observation of the quick decrease of the singular values reinforces the merit of our approach for that task. Moreover, WA enjoy characteristics that RNN do not, as for instance the existence of a computable distance between WA~\citep{dros09}, the opportunity to efficiently decide their equivalence~\citep{tzen92}, or the possibility to modify any WA such that it consists of a deterministic tree-structured part as deep as one wants followed by a non-deterministic part~\citep{baillythesis}. These elements indicate clear paths for the continuation of the study presented here. 

This paper rises other questions that constitute potential interesting future works. The first one is to take into account the observation on the size of the singular values directly in a distillation algorithm. For instance, one can imaging that another, smaller RNN can be designed from this information, following the framework of born again neural networks~\citep{furl18}. 

Another interesting line of research is to extend this approach to multi-valued weighted automata~\citep{rabu17}: these finite state machines can directly be trained on a language modelling task or can be used for multi-task learning. Testing whether this allows better distillation is an important question that should be answered. 

Extending our study to other types of RNN is also something that is worth to be mined. In particular, it would be interesting to know if second order RNN learned on data exhibit the same behavior than the one observed here: as their linear version is equivalent to WA~\citep{rabu19}, the process would correspond to a linearization. 

\noindent
Verifying whether bi-directional RNN~\citep{schu97} also share the finite state flavour that we detect in this paper provides also an interesting future line of research. Because of the works on learning context-free grammars~\citep{clar07,clar10} and beyond context-free~\citep{clar14}, we conjecture that it would not be the case. Indeed, bi-directional RNN use information from both the prefix and the suffix of a given element, which positions them within the spectrum of distributional learning~\citep{clar16}. The work on spectral learning of non finite state models~\citep{bail13} would then be then a good starting point for a distillation process for these networks. 

\section*{Acknowledgments}
We want to thanks our former student No\'e Goudian for developing the first version of the code used for the experiments. 
We owe likewise a great debt of gratitude to Gail Weiss for the long discussions, the code sharing, and the hours passed confronting our results.
We are also very grateful to Guillaume Rabusseau, Fran{\c c}ois Denis, and Matthias Gal\'e for fruitful discussions and for pointing to us important pieces of literature.
This work was performed using HPC resources from Centre de Calcul Intensif d’Aix-Marseille and GENCI-IDRIS (Grant 2020-AD011011920).





\bibliographystyle{plainnat}
\bibliography{RNN2WA}
\section*{Appendix}

Tables~\ref{tab:pautomac} and \ref{tab:spice} gives numerical detailed results of our experiments on PAutomaC and SpiCe, respectively. For each metric, for each evaluation set, the tables give the best value obtained by the spectral extraction with a RNN sampled basis and the n-gram baseline. For the spectral extraction, corresponding best parameters are given as a pair: the first value is the number of prefixes and suffixes of the basis, the second one is the rank. The value between parenthesis for n-gram is the size of the window. 
To give an idea of the quality of the trained RNN on each problem, the tables also provide their loss on a validation set.
The Table also gives the Hankel rank for each RNN for bases made of 1\,000 prefixes and suffixes (column \textit{HR}). 
The difference between the best obtained NDCG [resp. WER-D] by a WA and the NDCG [resp. WER-D] at the Hankel rank for the corresponding basis size is given in column $\Delta_{\textrm{NDCG}}$ [resp. $\Delta_{\textrm{WER-D}}$].

\newpage

\begin{landscape}


\begin{table}
  \hspace{-5.5cm}
  \begin{tabular}{|c|c|c|c|c|c|c|c|c|c|c|c|c|}
    \cline{2-9}
    \multicolumn{1}{c|}{} & \multicolumn{4}{|c|}{NDCG $\uparrow$} & \multicolumn{4}{|c|}{WER-D $\downarrow$} & \multicolumn{4}{|c}{} \\
    \cline{2-11}
    \multicolumn{1}{c|}{} & \multicolumn{2}{|c|}{$S_{Test}$} & \multicolumn{2}{|c|}{$S_{RNN}$} & \multicolumn{2}{|c|}{$S_{Test}$} & \multicolumn{2}{|c|}{$S_{RNN}$} & \multicolumn{2}{|c}{RNN} & \multicolumn{2}{|c}{} \\
    \hline
    Pb & WA & n-gram & WA & n-gram & WA & n-gram & WA & n-gram & loss & HR & $\Delta_{\textrm{NDCG}}$ & $\Delta_{\textrm{WER-D}}$ \\ \hline

1 & 0.945 (3000, 100) & 0.9569 (4) & 0.938 (3000, 100) & 0.9587 (4) & 0.24 (3000, 250) & 0.1957 (4) & 0.252 (3000, 100) & 0.1952 (4) & 0.008 & 15 & 0.063 & 0.181 \\ \hline 
2 & 0.947 (3000, 1000) & 0.9444 (3) & 0.948 (1500, 250) & 0.9602 (3) & 0.079 (3000, 1000) & 0.0837 (3) & 0.081 (1000, 24) & 0.0695 (3) & 0.007 & 13 & 0.021 & 0.017 \\ \hline 
3 & 0.991 (3000, 250) & 0.9832 (6) & 0.993 (3000, 250) & 0.9824 (6) & 0.103 (3000, 500) & 0.2236 (6) & 0.101 (3000, 250) & 0.2343 (6) & 0.006 & 18 & 0.018 & 0.092 \\ \hline 
4 & 1.0 (500, 100) & 0.9579 (5) & 1.0 (500, 100) & 0.9584 (5) & 0.178 (3000, 1000) & 0.297 (5) & 0.177 (1000, 100) & 0.2987 (5) & 0.005 & 13 & 0.001 & 0.003 \\ \hline 
5 & 1.0 (100, 39) & 0.9827 (4) & 1.0 (100, 42) & 0.9824 (4) & 0.229 (3000, 33) & 0.2281 (9) & 0.224 (1000, 18) & 0.2255 (8) & 0.003 & 10 & 0.001 & 0.015 \\ \hline 
6 & 0.998 (3000, 100) & 0.9879 (6) & 0.992 (3000, 250) & 0.9882 (6) & 0.097 (3000, 250) & 0.0698 (6) & 0.102 (3000, 250) & 0.066 (6) & 0.004 & 16 & 0.014 & 0.021 \\ \hline 
7 & 1.0 (500, 75) & 0.9637 (3) & 1.0 (1000, 250) & 0.9622 (3) & 0.126 (3000, 1000) & 0.3078 (4) & 0.116 (3000, 1000) & 0.314 (4) & 0.006 & 16 & 0.004 & 0.052 \\ \hline 
8 & 0.968 (3000, 250) & 0.9543 (4) & 0.953 (3000, 250) & 0.9537 (4) & 0.206 (3000, 250) & 0.2106 (4) & 0.226 (3000, 250) & 0.2087 (4) & 0.007 & 28 & 0.047 & 0.112 \\ \hline 
9 & 0.998 (1500, 250) & 0.9899 (6) & 1.0 (3000, 500) & 0.9908 (6) & 0.027 (1500, 250) & 0.0261 (6) & 0.024 (3000, 500) & 0.0262 (6) & 0.002 & 16 & 0.012 & 0.036 \\ \hline 
10 & 0.794 (3000, 100) & 0.8955 (4) & 0.744 (3000, 75) & 0.899 (4) & 0.49 (3000, 100) & 0.3483 (4) & 0.535 (3000, 75) & 0.3479 (4) & 0.007 & 20 & 0.114 & 0.15 \\ \hline 
11 & 0.692 (3000, 75) & 0.8254 (3) & 0.643 (3000, 75) & 0.8381 (3) & 0.612 (3000, 75) & 0.4654 (3) & 0.641 (3000, 100) & 0.4586 (3) & 0.008 & 21 & 0.178 & 0.221 \\ \hline 
12 & 0.995 (3000, 15) & 0.9855 (4) & 0.986 (1500, 48) & 0.9852 (4) & 0.145 (3000, 100) & 0.0756 (3) & 0.142 (3000, 75) & 0.0757 (3) & 0.005 & 15 & 0.011 & 0.025 \\ \hline 
13 & 0.997 (1500, 250) & 0.9853 (8) & 0.995 (1500, 250) & 0.983 (7) & 0.062 (3000, 500) & 0.0791 (7) & 0.072 (3000, 250) & 0.0882 (7) & 0.004 & 30 & 0.045 & 0.144 \\ \hline 
14 & 0.998 (3000, 75) & 0.9583 (3) & 0.997 (1500, 250) & 0.9533 (3) & 0.277 (3000, 500) & 0.2371 (4) & 0.255 (1000, 100) & 0.2359 (4) & 0.008 & 12 & 0.065 & 0.062 \\ \hline 
15 & 0.912 (3000, 45) & 0.9393 (3) & 0.886 (3000, 42) & 0.9453 (3) & 0.297 (3000, 42) & 0.2044 (3) & 0.312 (3000, 42) & 0.1958 (3) & 0.007 & 20 & 0.035 & 0.057 \\ \hline 
16 & 0.955 (3000, 75) & 0.8905 (4) & 0.932 (3000, 100) & 0.8881 (4) & 0.268 (3000, 250) & 0.3546 (4) & 0.29 (3000, 100) & 0.3568 (4) & 0.007 & 19 & 0.225 & 0.288 \\ \hline 
17 & 0.956 (3000, 100) & 0.978 (3) & 0.96 (3000, 75) & 0.9794 (3) & 0.146 (3000, 100) & 0.1251 (3) & 0.14 (3000, 75) & 0.1108 (3) & 0.006 & 22 & 0.021 & 0.049 \\ \hline 
18 & 0.999 (3000, 250) & 0.9866 (4) & 0.997 (1000, 75) & 0.9868 (4) & 0.084 (3000, 250) & 0.0841 (3) & 0.095 (1500, 42) & 0.0823 (3) & 0.005 & 23 & 0.107 & 0.0 \\ \hline 
19 & 0.938 (3000, 250) & 0.9318 (5) & 0.929 (3000, 250) & 0.9314 (5) & 0.19 (3000, 250) & 0.1639 (5) & 0.207 (3000, 250) & 0.1697 (5) & 0.005 & 12 & 0.201 & 0.35 \\ \hline 
20 & 0.974 (3000, 42) & 0.9656 (3) & 0.978 (3000, 39) & 0.965 (3) & 0.208 (3000, 42) & 0.1902 (3) & 0.208 (3000, 39) & 0.1936 (3) & 0.008 & 15 & 0.022 & 0.019 \\ \hline 
21 & 0.535 (3000, 100) & 0.8772 (3) & 0.569 (3000, 100) & 0.8853 (3) & 0.701 (3000, 100) & 0.2965 (3) & 0.675 (3000, 100) & 0.2875 (3) & 0.008 & 22 & 0.267 & 0.267 \\ \hline 
22 & 0.606 (1500, 75) & 0.8985 (3) & 0.606 (1500, 75) & 0.9043 (3) & 0.625 (1500, 39) & 0.3029 (3) & 0.627 (1500, 75) & 0.3003 (3) & 0.007 & 27 & 0.028 & 0.018 \\ \hline 
23 & 0.979 (3000, 250) & 0.9475 (4) & 0.982 (3000, 250) & 0.9501 (4) & 0.174 (3000, 250) & 0.1409 (4) & 0.169 (3000, 250) & 0.1341 (4) & 0.006 & 12 & 0.067 & 0.123 \\ \hline 
24 & 1.0 (100, 18) & 0.9613 (5) & 1.0 (100, 18) & 0.9615 (5) & 0.246 (500, 100) & 0.4483 (3) & 0.243 (3000, 39) & 0.4468 (5) & 0.005 & 10 & 0.001 & 0.004 \\ \hline 

 \end{tabular}
  \caption{Detailed results on the PAutomaC benchmark.}
  \label{tab:pautomac}
\end{table}

\end{landscape}

\newpage
\begin{landscape}
\begin{table}
  \hspace{-5.5cm}
  \begin{tabular}{|c|c|c|c|c|c|c|c|c|c|c|c|c|}
    \cline{2-9}
    \multicolumn{1}{c|}{} & \multicolumn{4}{|c|}{NDCG $\uparrow$} & \multicolumn{4}{|c|}{WER-D $\downarrow$} & \multicolumn{4}{|c}{} \\
    \cline{2-11}
    \multicolumn{1}{c|}{} & \multicolumn{2}{|c|}{$S_{Test}$} & \multicolumn{2}{|c|}{$S_{RNN}$} & \multicolumn{2}{|c|}{$S_{Test}$} & \multicolumn{2}{|c|}{$S_{RNN}$} & \multicolumn{2}{|c}{RNN} & \multicolumn{2}{|c}{} \\
    \hline
    Pb & WA & n-gram & WA & n-gram & WA & n-gram & WA & n-gram & loss & HR & $\Delta_{\textrm{NDCG}}$ & $\Delta_{\textrm{WER-D}}$ \\ \hline

25 & 0.918 (3000, 250) & 0.9163 (4) & 0.927 (3000, 250) & 0.9146 (4) & 0.285 (3000, 250) & 0.3035 (4) & 0.276 (3000, 250) & 0.3113 (4) & 0.008 & 9 & 0.177 & 0.252 \\ \hline

26 & 0.953 (3000, 100) & 0.9587 (5) & 0.951 (3000, 100) & 0.9629 (5) & 0.204 (3000, 250) & 0.1592 (6) & 0.203 (3000, 250) & 0.1509 (6) & 0.005 & 34 & 0.073 & 0.2 \\ \hline 
27 & 0.93 (3000, 45) & 0.885 (3) & 0.885 (3000, 45) & 0.8891 (3) & 0.369 (3000, 100) & 0.3462 (3) & 0.418 (3000, 45) & 0.3527 (3) & 0.008 & 17 & 0.036 & 0.083 \\ \hline 
28 & 0.995 (3000, 500) & 0.9372 (4) & 0.994 (3000, 500) & 0.9383 (4) & 0.213 (3000, 500) & 0.3591 (4) & 0.212 (3000, 500) & 0.3431 (4) & 0.008 & 10 & 0.054 & 0.351 \\ \hline 
29 & 0.986 (3000, 100) & 0.9413 (5) & 0.976 (3000, 250) & 0.9418 (5) & 0.15 (3000, 500) & 0.2516 (5) & 0.164 (3000, 1000) & 0.244 (5) & 0.005 & 20 & 0.042 & 0.144 \\ \hline 
30 & 0.979 (3000, 33) & 0.9329 (3) & 0.976 (3000, 250) & 0.9353 (3) & 0.267 (3000, 500) & 0.2819 (3) & 0.266 (3000, 250) & 0.2713 (3) & 0.008 & 11 & 0.018 & 0.07 \\ \hline 
31 & 0.999 (3000, 500) & 0.9599 (4) & 0.998 (3000, 250) & 0.9604 (4) & 0.108 (3000, 500) & 0.1698 (5) & 0.111 (3000, 250) & 0.1726 (5) & 0.006 & 12 & 0.013 & 0.123 \\ \hline 
32 & 0.997 (3000, 1000) & 0.9684 (7) & 0.996 (3000, 500) & 0.9673 (7) & 0.092 (3000, 1000) & 0.1659 (8) & 0.101 (3000, 500) & 0.1685 (7) & 0.005 & 24 & 0.016 & 0.097 \\ \hline 
33 & 0.995 (1500, 39) & 0.969 (3) & 0.994 (3000, 42) & 0.969 (3) & 0.014 (500, 6) & 0.1008 (3) & 0.01 (3000, 50) & 0.0999 (3) & 0.007 & 11 & 0.02 & 0.04 \\ \hline 
34 & 0.465 (3000, 50) & 0.8707 (3) & 0.482 (3000, 50) & 0.8775 (3) & 0.785 (3000, 50) & 0.4048 (3) & 0.773 (3000, 50) & 0.398 (3) & 0.008 & 15 & 0.07 & 0.076 \\ \hline 
35 & 0.815 (3000, 100) & 0.9148 (3) & 0.771 (3000, 75) & 0.9186 (3) & 0.415 (3000, 100) & 0.2816 (3) & 0.455 (3000, 100) & 0.2757 (3) & 0.006 & 26 & 0.23 & 0.284 \\ \hline 
36 & 0.921 (3000, 50) & 0.9542 (3) & 0.919 (3000, 50) & 0.9527 (3) & 0.38 (3000, 100) & 0.3128 (3) & 0.384 (3000, 48) & 0.3212 (3) & 0.008 & 13 & 0.089 & 0.234 \\ \hline 
37 & 0.999 (3000, 250) & 0.9747 (3) & 0.999 (3000, 250) & 0.9752 (3) & 0.136 (3000, 250) & 0.2277 (3) & 0.133 (3000, 50) & 0.2206 (3) & 0.009 & 11 & 0.008 & 0.145 \\ \hline 
38 & 0.998 (3000, 250) & 0.9455 (3) & 0.998 (3000, 250) & 0.9459 (3) & 0.307 (3000, 250) & 0.3744 (3) & 0.303 (3000, 250) & 0.3643 (3) & 0.009 & 6 & 0.017 & 0.123 \\ \hline 
39 & 0.996 (3000, 30) & 0.9608 (4) & 0.997 (3000, 100) & 0.9588 (4) & 0.145 (3000, 100) & 0.1687 (4) & 0.134 (3000, 6) & 0.1583 (4) & 0.006 & 9 & 0.006 & 0.135 \\ \hline 
40 & 0.735 (1500, 48) & 0.888 (3) & 0.716 (3000, 50) & 0.8898 (3) & 0.695 (3000, 100) & 0.4751 (3) & 0.706 (3000, 100) & 0.4795 (3) & 0.009 & 10 & 0.129 & 0.211 \\ \hline 
41 & 1.0 (3000, 250) & 0.9793 (3) & 1.0 (3000, 250) & 0.9792 (3) & 0.048 (3000, 250) & 0.1602 (3) & 0.045 (3000, 250) & 0.1579 (3) & 0.008 & 7 & 0.045 & 0.432 \\ \hline 
42 & 1.0 (1500, 250) & 0.9648 (4) & 0.999 (3000, 500) & 0.9632 (4) & 0.01 (3000, 250) & 0.0723 (4) & 0.01 (3000, 100) & 0.0761 (4) & 0.006 & 9 & 0.01 & 0.089 \\ \hline 
43 & 1.0 (100, 45) & 0.9743 (4) & 1.0 (100, 45) & 0.9743 (4) & 0.083 (3000, 500) & 0.2612 (2) & 0.088 (3000, 12) & 0.2649 (2) & 0.008 & 8 & 0.001 & 0.041 \\ \hline 
44 & 0.895 (3000, 50) & 0.9396 (3) & 0.896 (3000, 50) & 0.9408 (2) & 0.485 (3000, 250) & 0.435 (3) & 0.495 (3000, 50) & 0.434 (3) & 0.009 & 8 & 0.085 & 0.195 \\ \hline 
45 & 1.0 (500, 45) & 0.956 (3) & 1.0 (100, 30) & 0.9548 (3) & 0.0 (500, 27) & 0.0732 (2) & 0.0 (500, 75) & 0.0708 (2) & 0.009 & 4 & 0.001 & 0.001 \\ \hline 
46 & 0.705 (1500, 45) & 0.8945 (3) & 0.687 (1500, 39) & 0.9011 (3) & 0.567 (1500, 45) & 0.3421 (3) & 0.573 (1500, 42) & 0.3258 (3) & 0.009 & 10 & 0.154 & 0.231 \\ \hline 
47 & 0.961 (3000, 100) & 0.9536 (4) & 0.937 (3000, 100) & 0.952 (4) & 0.212 (3000, 100) & 0.1531 (4) & 0.231 (3000, 100) & 0.1571 (4) & 0.005 & 9 & 0.52 & 0.593 \\ \hline 
48 & 0.805 (1000, 24) & 0.8139 (3) & 0.816 (1000, 24) & 0.8282 (3) & 0.417 (1000, 39) & 0.4369 (3) & 0.41 (1000, 24) & 0.4209 (3) & 0.008 & 13 & 0.108 & 0.231 \\ \hline

\end{tabular}
  \caption{Detailed results on the PAutomaC benchmark (end).}
  \label{tab:pautomac2}
\end{table}

\end{landscape}

\newpage

\begin{landscape}


\begin{table}
  \hspace{-5.5cm}
  \begin{tabular}{|c|c|c|c|c|c|c|c|c|c|c|c|c|}
    \cline{2-9}
    \multicolumn{1}{c|}{} & \multicolumn{4}{|c|}{NDCG $\uparrow$} & \multicolumn{4}{|c|}{WER-D $\downarrow$} & \multicolumn{4}{|c}{} \\
    \cline{2-11}
    \multicolumn{1}{c|}{} & \multicolumn{2}{|c|}{$S_{Test}$} & \multicolumn{2}{|c|}{$S_{RNN}$} & \multicolumn{2}{|c|}{$S_{Test}$} & \multicolumn{2}{|c|}{$S_{RNN}$} & \multicolumn{2}{|c}{RNN} & \multicolumn{2}{|c}{} \\
    \hline
    Pb & WA & n-gram & WA & n-gram & WA & n-gram & WA & n-gram & loss & HR & $\Delta_{\textrm{NDCG}}$ & $\Delta_{\textrm{WER-D}}$ \\ \hline

1 & 0.645 (1000, 175) & 0.9569 (4) & 0.652 (100, 10) & 0.9587 (4) & 0.624 (3000, 40) & 0.1957 (4) & 0.629 (1500, 60) & 0.1952 (4) & 0.003 & 14 & 0.191 & 0.068 \\ \hline 
2 & 0.686 (1500, 30) & 0.9444 (3) & 0.689 (1500, 30) & 0.9602 (3) & 0.552 (1000, 35) & 0.0837 (3) & 0.542 (1000, 35) & 0.0695 (3) & 0.002 & 17 & 0.112 & 0.061 \\ \hline 
3 & 0.639 (3000, 90) & 0.9832 (6) & 0.645 (3000, 90) & 0.9824 (6) & 0.746 (3000, 90) & 0.2236 (6) & 0.739 (3000, 500) & 0.2343 (6) & 0.002 & 19 & 0.078 & 0.149 \\ \hline 
4 & 0.752 (1500, 250) & 0.9579 (5) & 0.776 (1500, 250) & 0.9584 (5) & 0.595 (1000, 250) & 0.297 (5) & 0.588 (1500, 250) & 0.2987 (5) & 0.005 & 165 & 0.021 & 0.014 \\ \hline 
5 & 0.284 (500, 20) & 0.9827 (4) & 0.342 (1000, 2) & 0.9824 (4) & 0.815 (500, 2) & 0.2281 (9) & 0.808 (500, 2) & 0.2255 (8) & 0.001 & 15 & 0.05 & 0.039 \\ \hline 
6 & 0.466 (1000, 35) & 0.9879 (6) & 0.483 (1000, 50) & 0.9882 (6) & 0.747 (1000, 60) & 0.0698 (6) & 0.754 (1000, 50) & 0.066 (6) & 0.002 & 31 & 0.016 & 0.036 \\ \hline 
7 & 0.326 (3000, 150) & 0.9637 (3) & 0.346 (3000, 150) & 0.9622 (3) & 0.833 (500, 3) & 0.3078 (4) & 0.83 (500, 3) & 0.314 (4) & 0.002 & 4 & 0.115 & 0.009 \\ \hline 
8 & 0.446 (1500, 25) & 0.9543 (4) & 0.452 (1500, 25) & 0.9537 (4) & 0.725 (100, 4) & 0.2106 (4) & 0.726 (100, 4) & 0.2087 (4) & 0.002 & 28 & 0.069 & 0.164 \\ \hline 
9 & 0.9 (3000, 50) & 0.9899 (6) & 0.891 (3000, 150) & 0.9908 (6) & 0.215 (3000, 200) & 0.0261 (6) & 0.223 (3000, 150) & 0.0262 (6) & 0.001 & 36 & 0.133 & 0.26 \\ \hline 
10 & 0.534 (1500, 40) & 0.8955 (4) & 0.541 (1500, 40) & 0.899 (4) & 0.709 (1500, 40) & 0.3483 (4) & 0.705 (1500, 40) & 0.3479 (4) & 0.003 & 33 & 0.0 & 0.0 \\ \hline 
12 & 0.683 (1500, 70) & 0.8254 (3) & 0.69 (1500, 70) & 0.8381 (3) & 0.796 (1500, 70) & 0.4654 (3) & 0.785 (1500, 70) & 0.4586 (3) & 0.002 & 33 & 0.161 & 0.141 \\ \hline 
13 & 0.436 (500, 90) & 0.9855 (4) & 0.44 (500, 90) & 0.9852 (4) & 0.77 (500, 90) & 0.0756 (3) & 0.761 (500, 90) & 0.0757 (3) & 0.004 & 22 & 0.235 & 0.171 \\ \hline 
14 & 0.534 (1000, 150) & 0.9853 (8) & 0.549 (1500, 150) & 0.983 (7) & 0.574 (1000, 150) & 0.0791 (7) & 0.562 (1500, 150) & 0.0882 (7) & 0.001 & 46 & 0.229 & 0.197 \\ \hline 
15 & 0.369 (1000, 60) & 0.9583 (3) & 0.375 (1000, 60) & 0.9533 (3) & 0.796 (1000, 60) & 0.2371 (4) & 0.785 (1000, 60) & 0.2359 (4) & 0.002 & 45 & 0.116 & 0.127 \\ \hline  
  \end{tabular}
  \caption{Detailed results on the SPiCe benchmark.}
  \label{tab:spice}
\end{table}


\end{landscape}

\thispagestyle{empty}
\begin{figure}[htbp!]
\vspace{-1.5cm}
 \begin{center}
\includegraphics[width=0.23\textwidth]{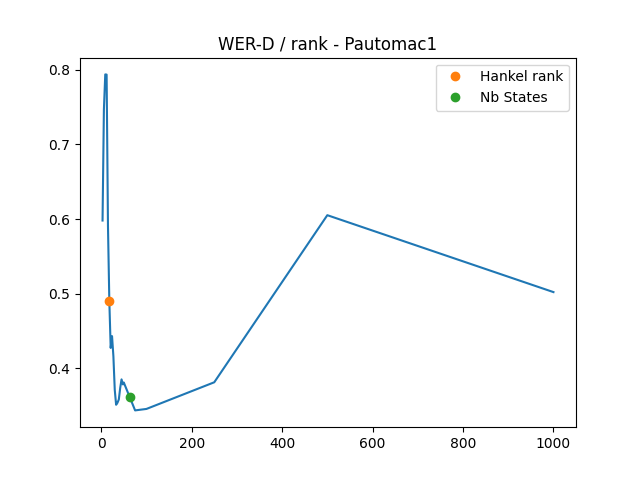}
\includegraphics[width=0.23\textwidth]{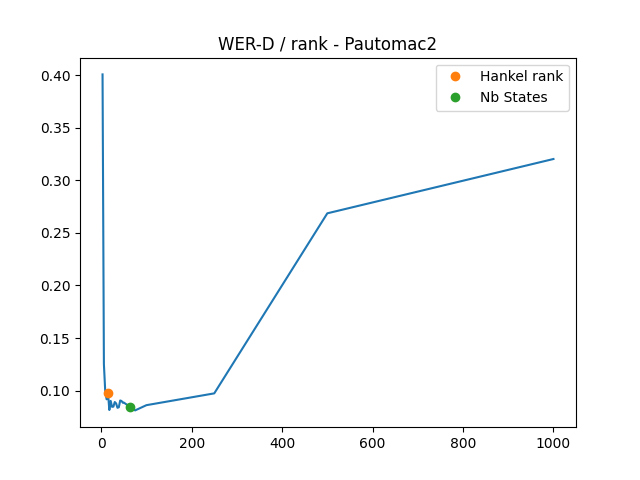}
\includegraphics[width=0.23\textwidth]{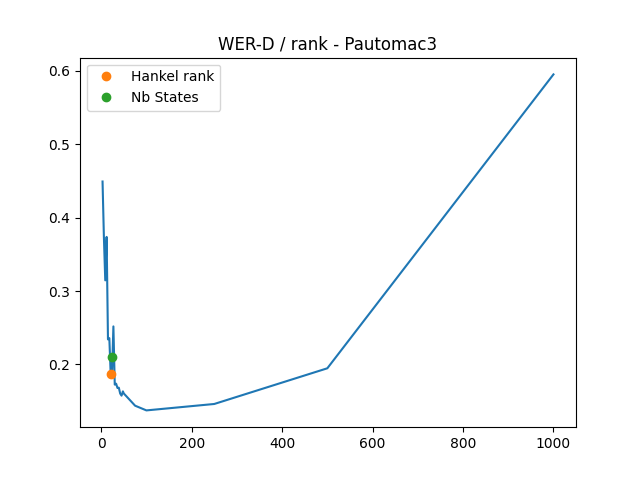}
\includegraphics[width=0.23\textwidth]{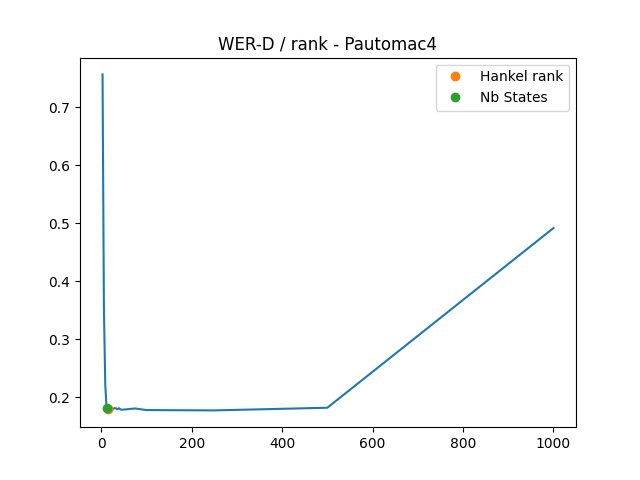}
\includegraphics[width=0.23\textwidth]{imgs/ranks/WER-D_rank-Pautomac5.png}
\includegraphics[width=0.23\textwidth]{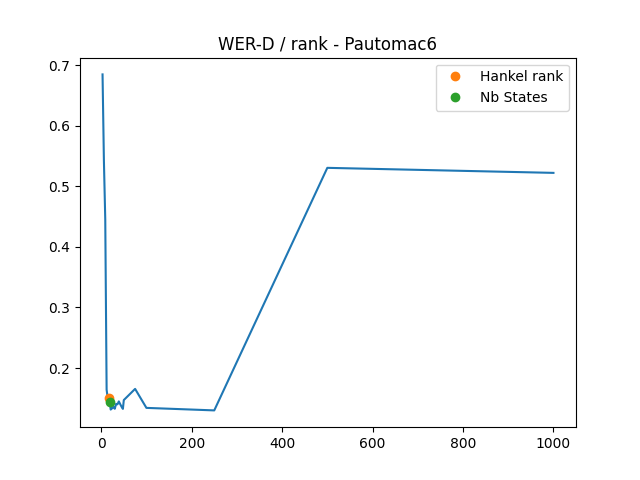}
\includegraphics[width=0.23\textwidth]{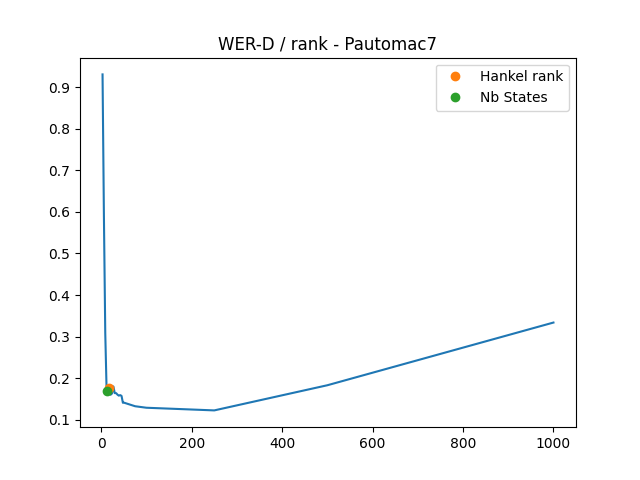}
\includegraphics[width=0.23\textwidth]{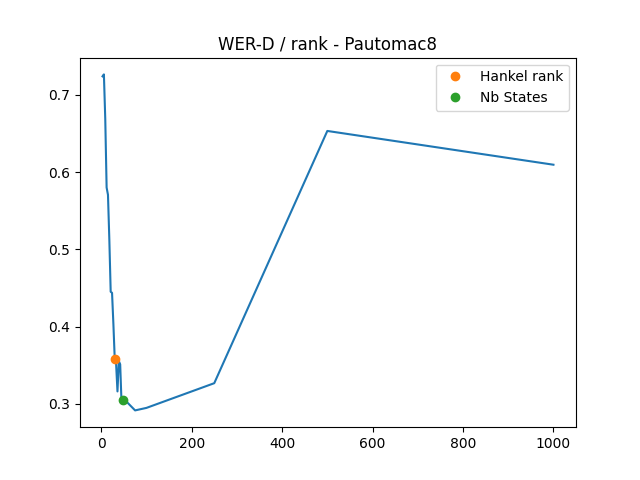}
\includegraphics[width=0.23\textwidth]{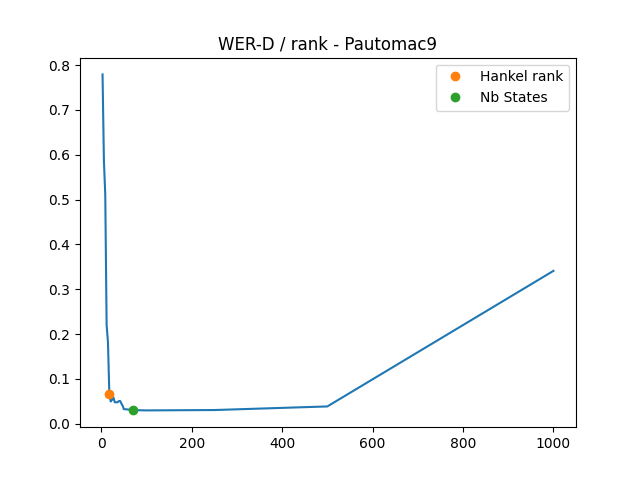}
\includegraphics[width=0.23\textwidth]{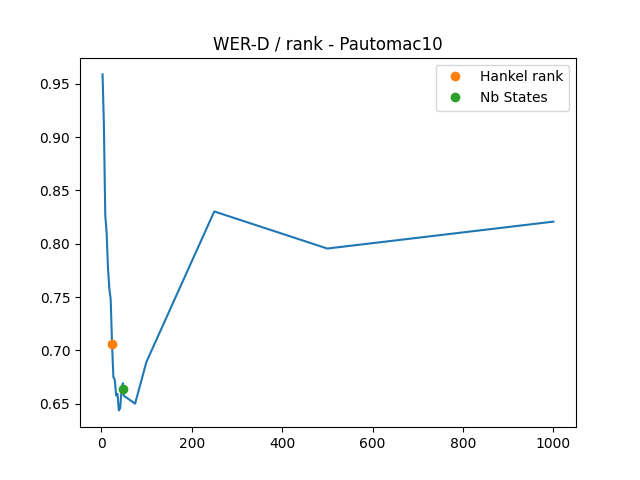}
\includegraphics[width=0.23\textwidth]{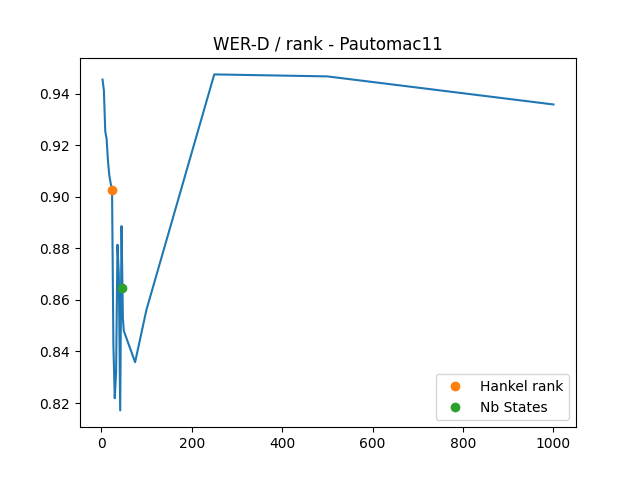}
\includegraphics[width=0.23\textwidth]{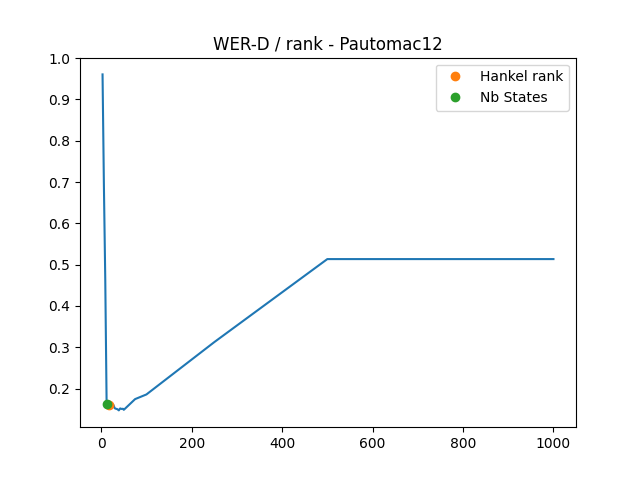}
\includegraphics[width=0.23\textwidth]{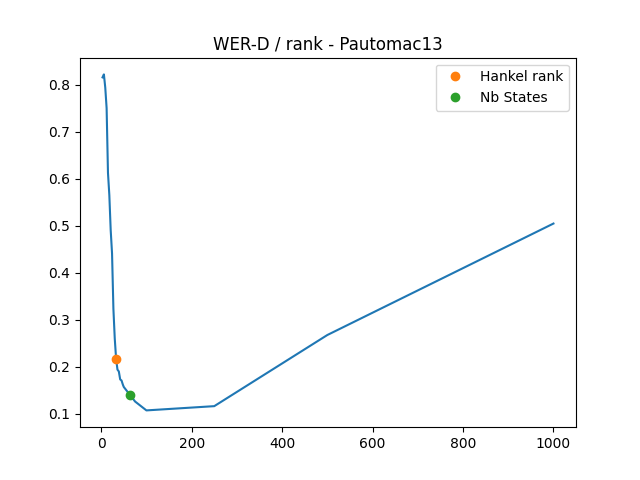}
\includegraphics[width=0.23\textwidth]{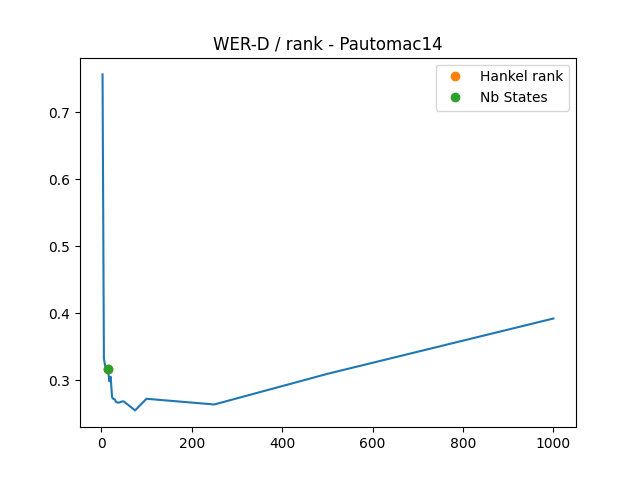}
\includegraphics[width=0.23\textwidth]{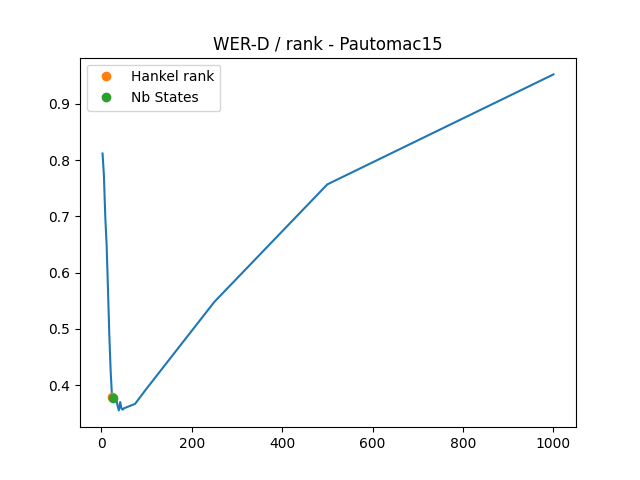}
\includegraphics[width=0.23\textwidth]{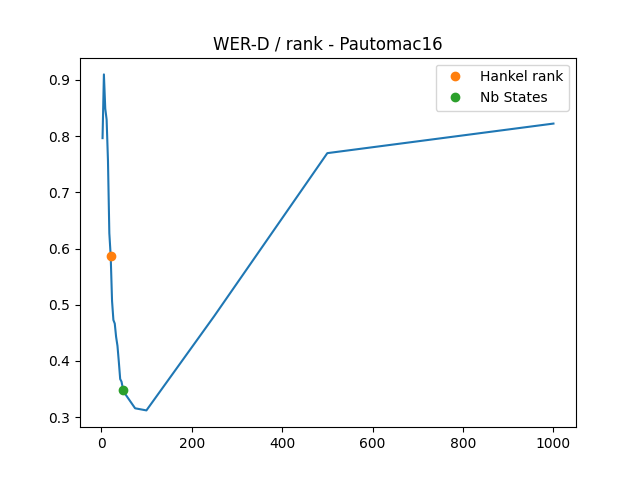}
\includegraphics[width=0.23\textwidth]{imgs/ranks/WER-D_rank-Pautomac17.png}
\includegraphics[width=0.23\textwidth]{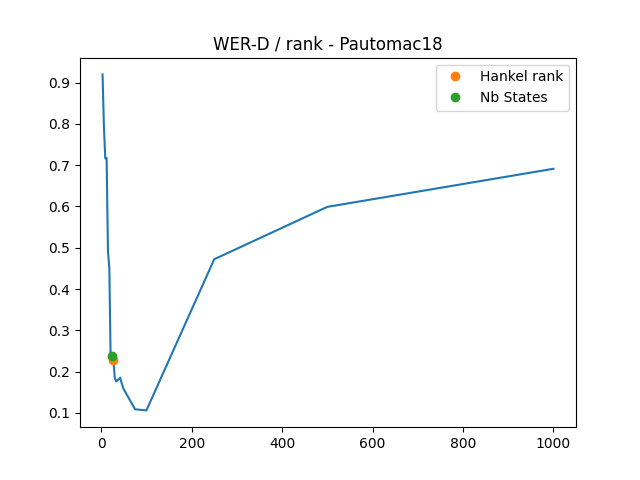}
\includegraphics[width=0.23\textwidth]{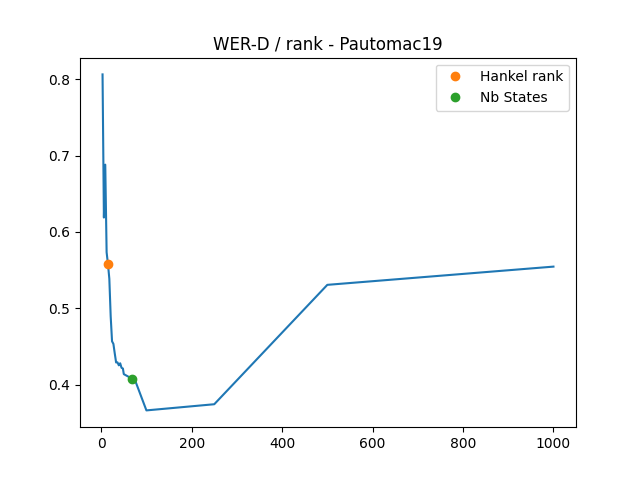}
\includegraphics[width=0.23\textwidth]{imgs/ranks/WER-D_rank-Pautomac20.png}
\includegraphics[width=0.23\textwidth]{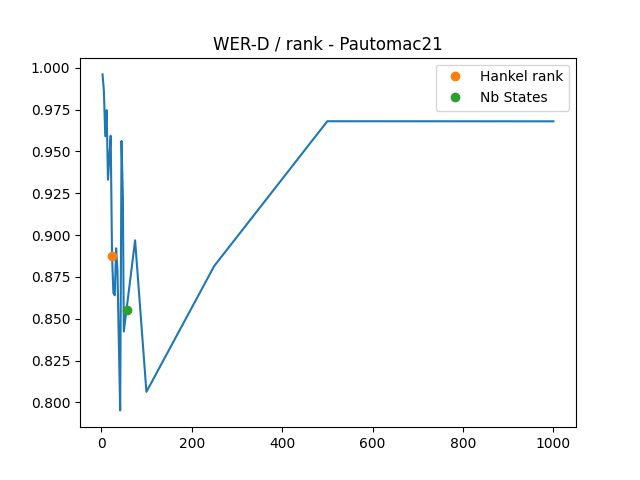}
\includegraphics[width=0.23\textwidth]{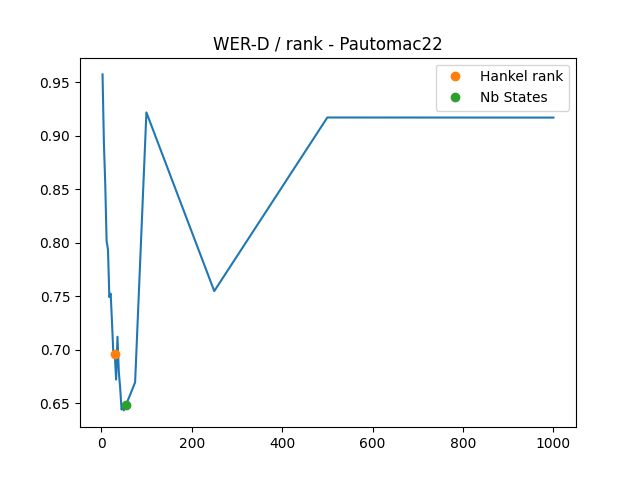}
\includegraphics[width=0.23\textwidth]{imgs/ranks/WER-D_rank-Pautomac23.png}
\includegraphics[width=0.23\textwidth]{imgs/ranks/WER-D_rank-Pautomac24.png}
\includegraphics[width=0.23\textwidth]{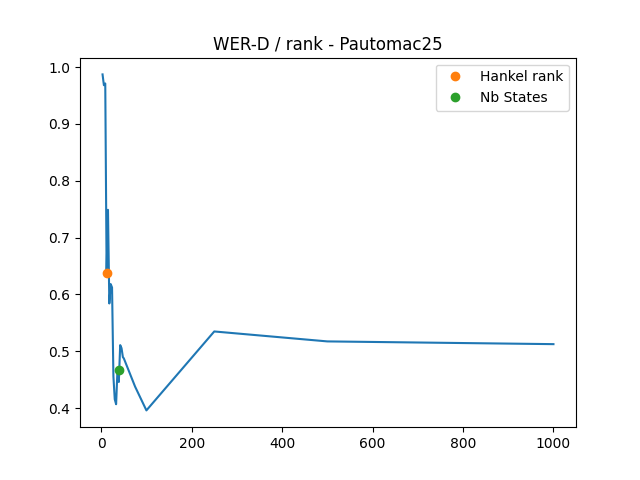}
\includegraphics[width=0.23\textwidth]{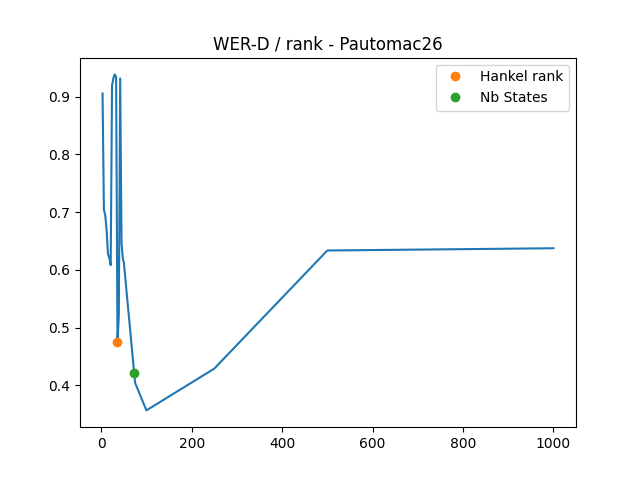}
\includegraphics[width=0.23\textwidth]{imgs/ranks/WER-D_rank-Pautomac27.png}
\includegraphics[width=0.23\textwidth]{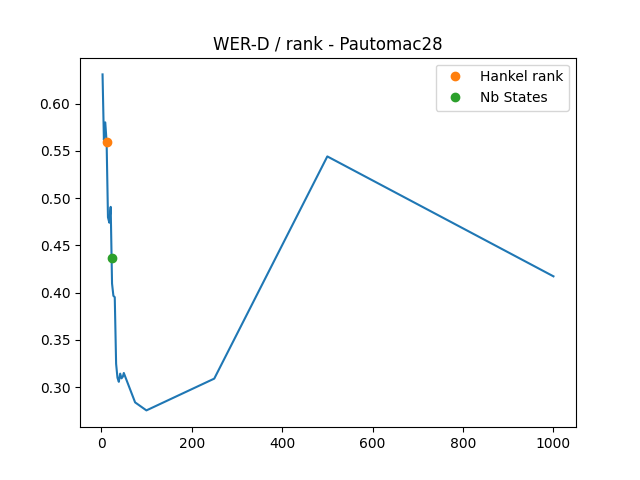}
\includegraphics[width=0.23\textwidth]{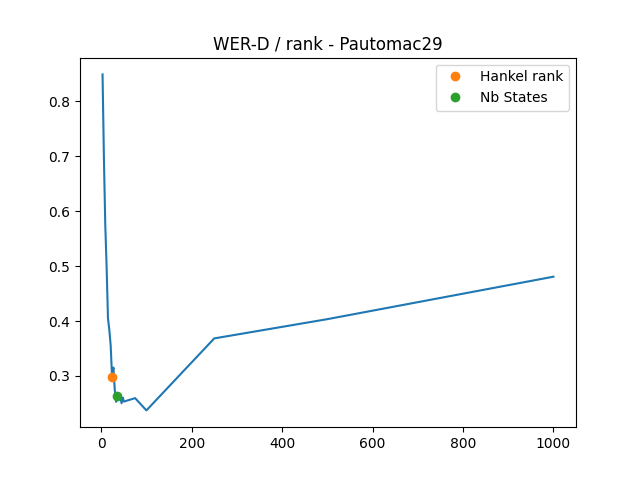}
\includegraphics[width=0.23\textwidth]{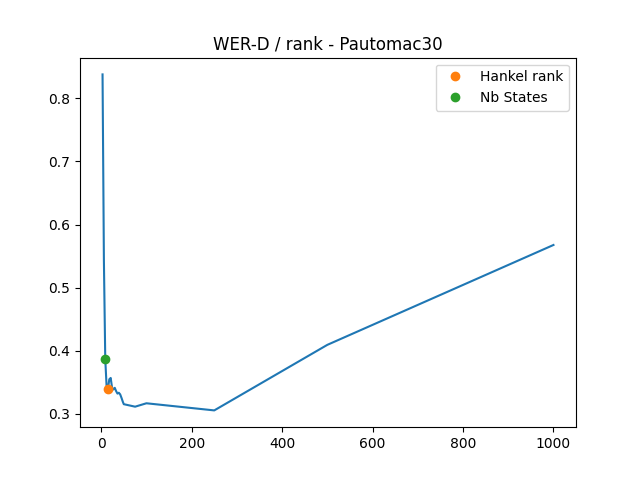}
\includegraphics[width=0.23\textwidth]{imgs/ranks/WER-D_rank-Pautomac31.png}
\includegraphics[width=0.23\textwidth]{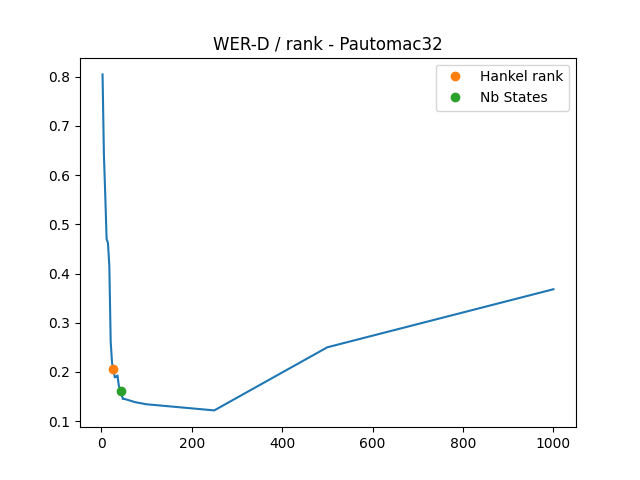}
\includegraphics[width=0.23\textwidth]{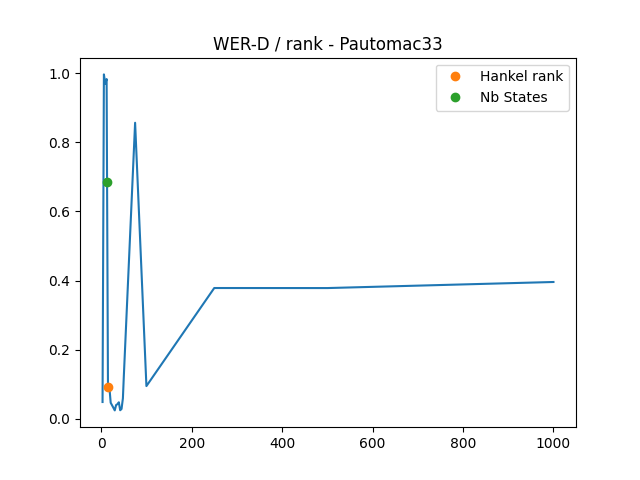}
\includegraphics[width=0.23\textwidth]{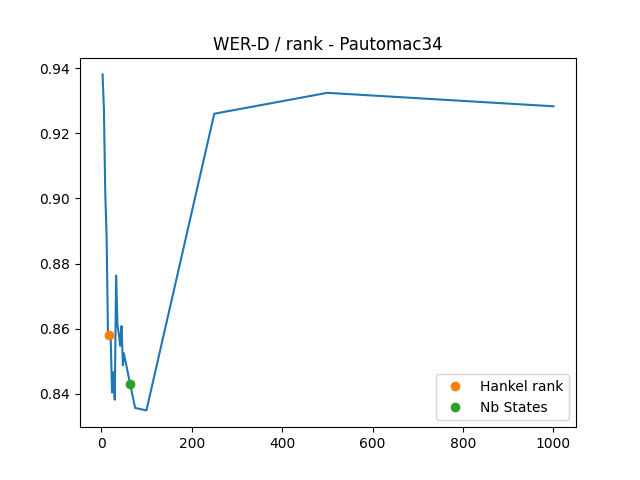}
\includegraphics[width=0.23\textwidth]{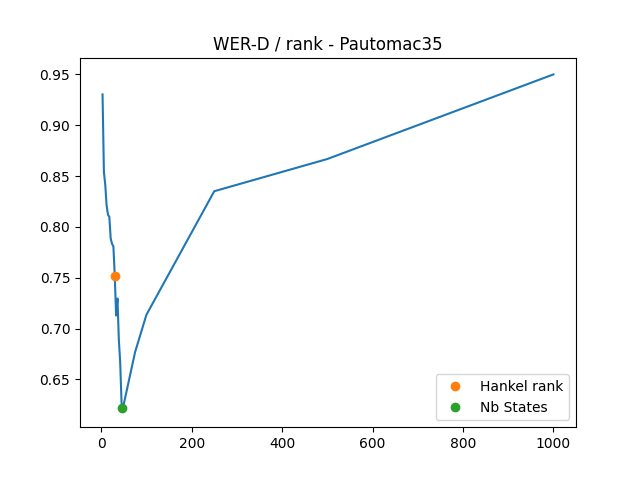}
\includegraphics[width=0.23\textwidth]{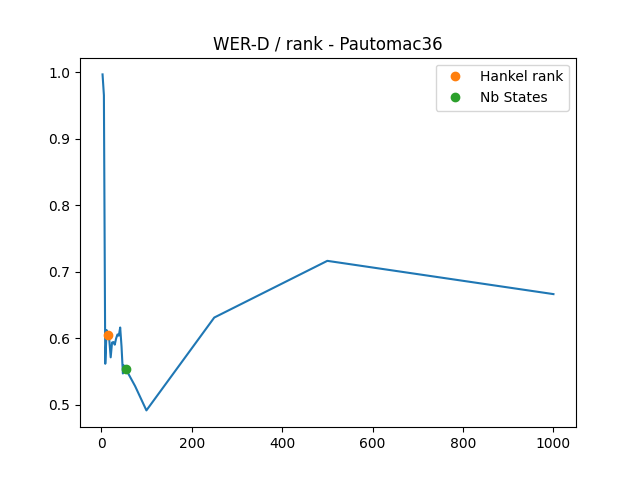}
\includegraphics[width=0.23\textwidth]{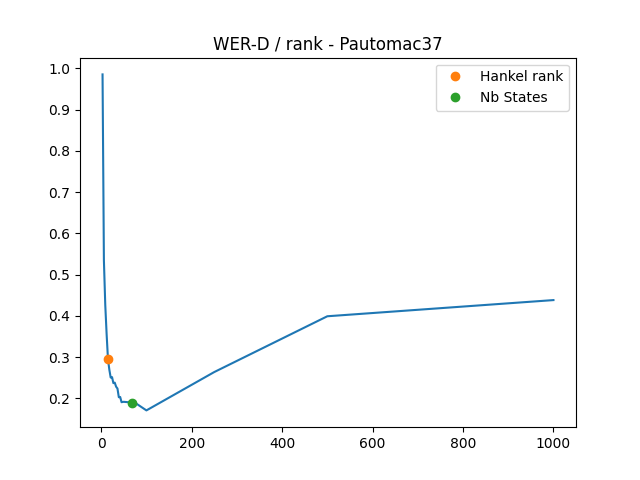}
\includegraphics[width=0.23\textwidth]{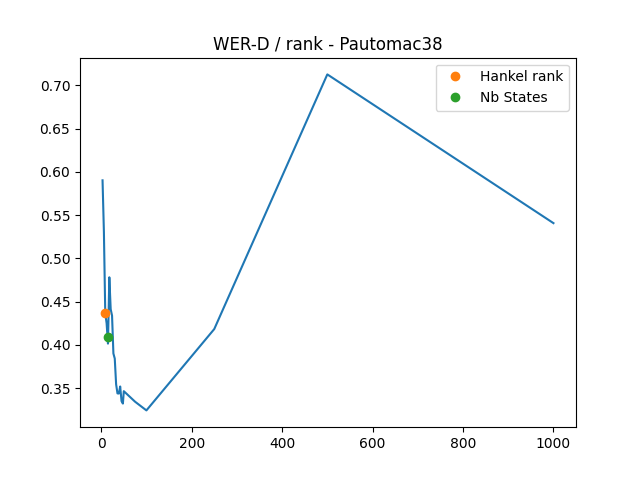}
\includegraphics[width=0.23\textwidth]{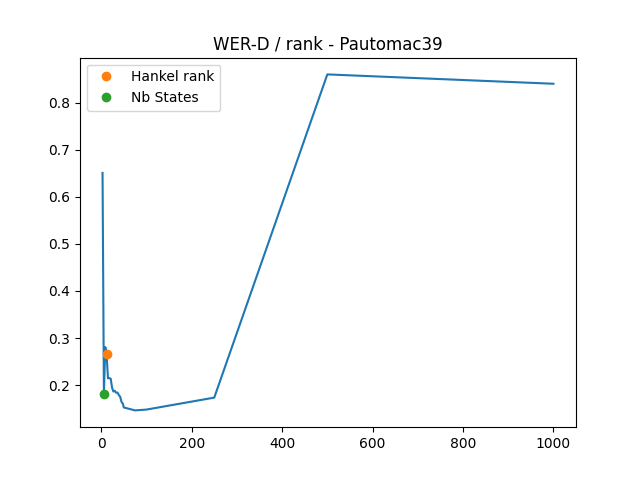}
\includegraphics[width=0.23\textwidth]{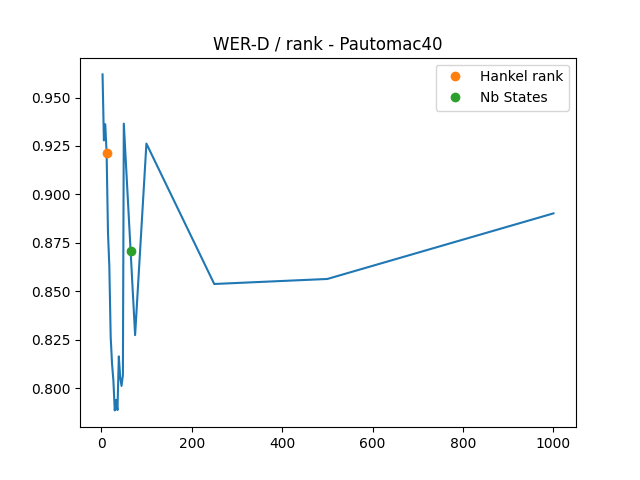}

\caption{Evolution of the quality in term of WER-D with the value of the rank hyper-parameter on the 40 first PAutomaC problems, evaluated on $S_{RNN}$ with basis size $1000\times 1000$.}
\label{fig:ranks-pautomac-1-40}
\end{center}
\end{figure}

\begin{figure}[htbp!]
 \begin{center}
\includegraphics[width=0.23\textwidth]{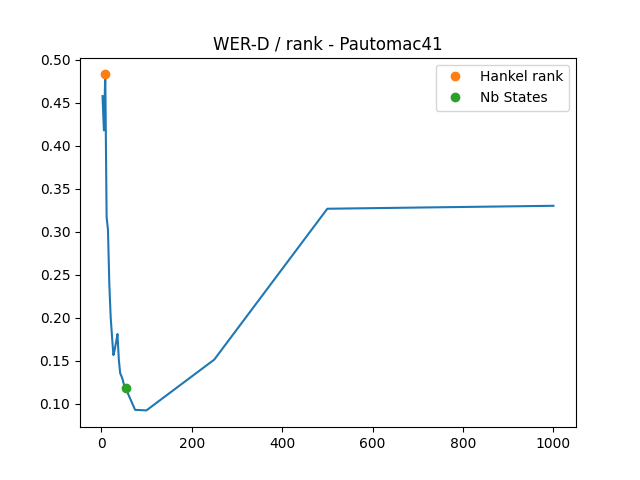}
\includegraphics[width=0.23\textwidth]{imgs/ranks/WER-D_rank-Pautomac42.png}
\includegraphics[width=0.23\textwidth]{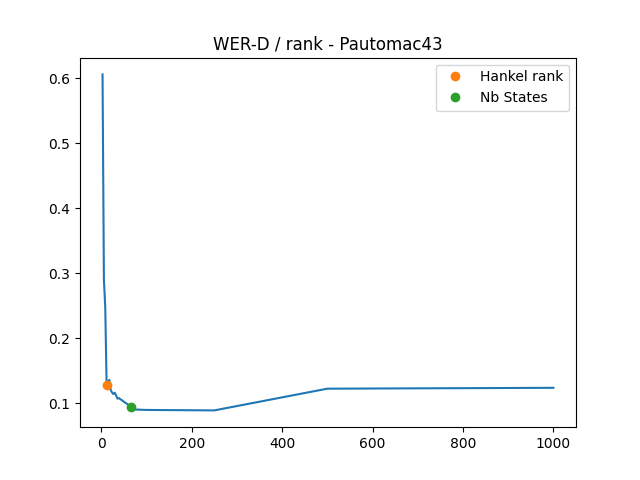}
\includegraphics[width=0.23\textwidth]{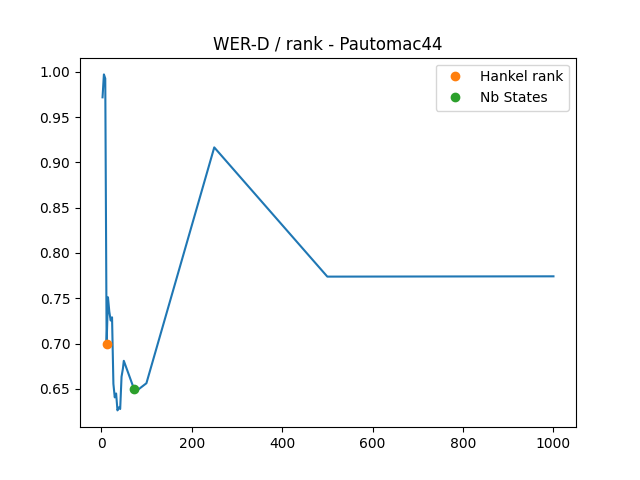}
\includegraphics[width=0.23\textwidth]{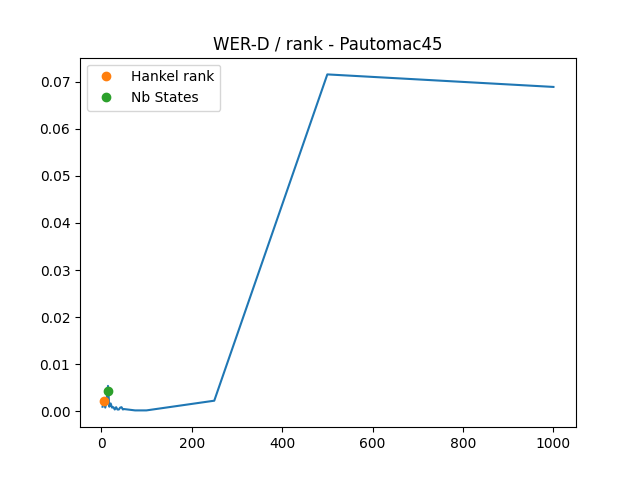}
\includegraphics[width=0.23\textwidth]{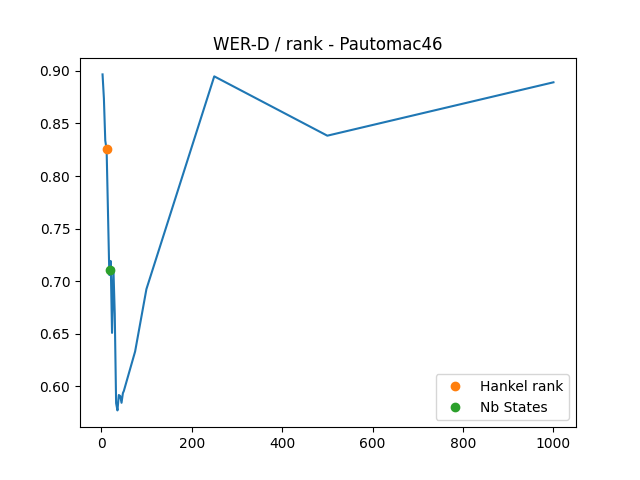}
\includegraphics[width=0.23\textwidth]{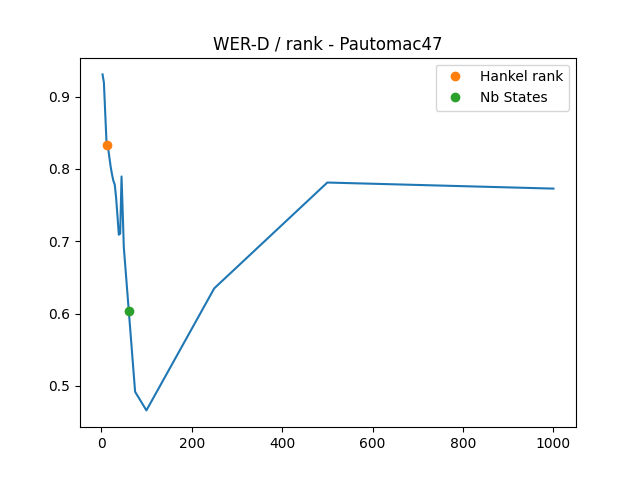}
\includegraphics[width=0.23\textwidth]{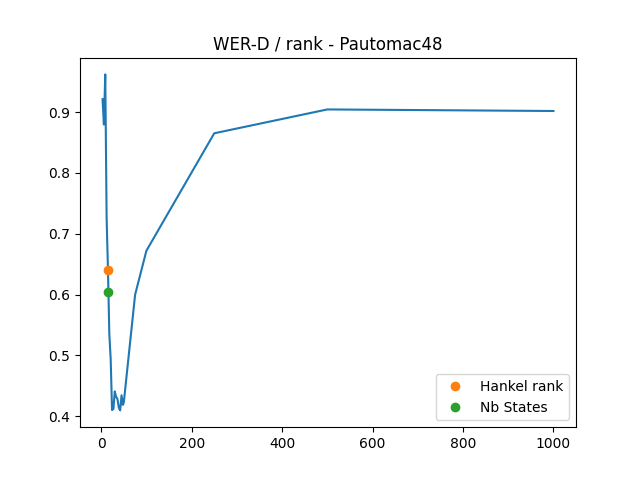}

\includegraphics[width=0.23\textwidth]{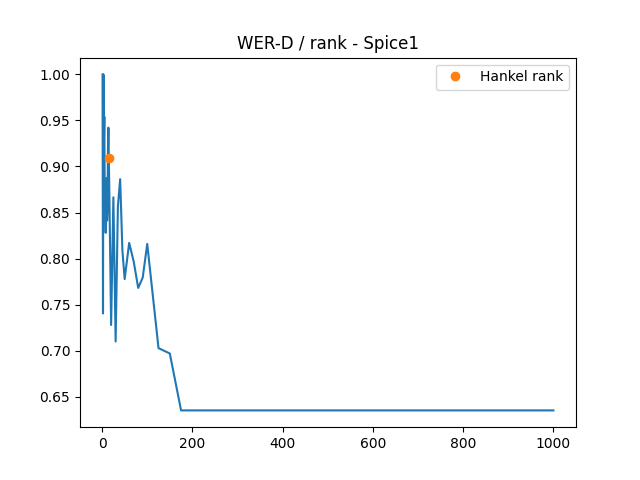}
\includegraphics[width=0.23\textwidth]{imgs/ranks/WER-D_rank-Spice2.png}
\includegraphics[width=0.23\textwidth]{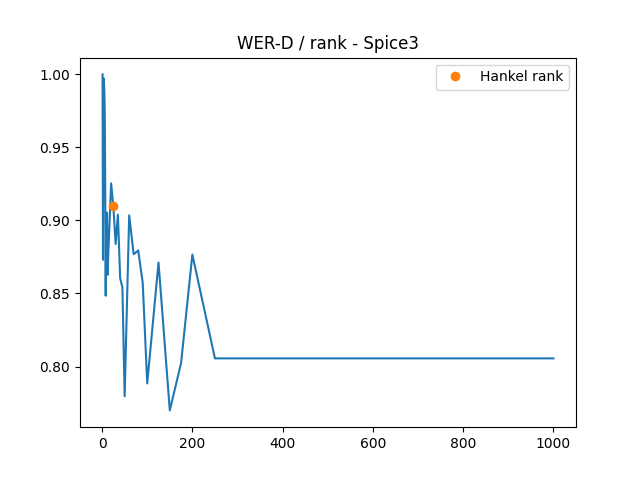}
\includegraphics[width=0.23\textwidth]{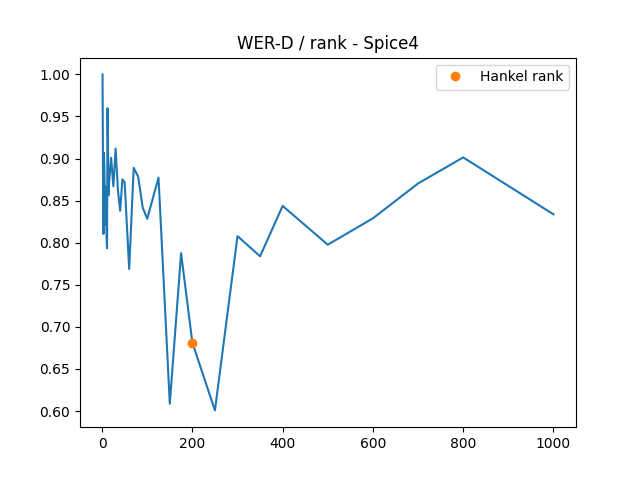}
\includegraphics[width=0.23\textwidth]{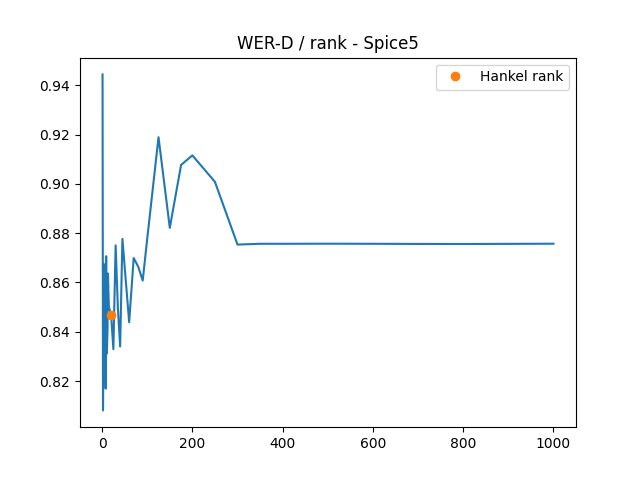}
\includegraphics[width=0.23\textwidth]{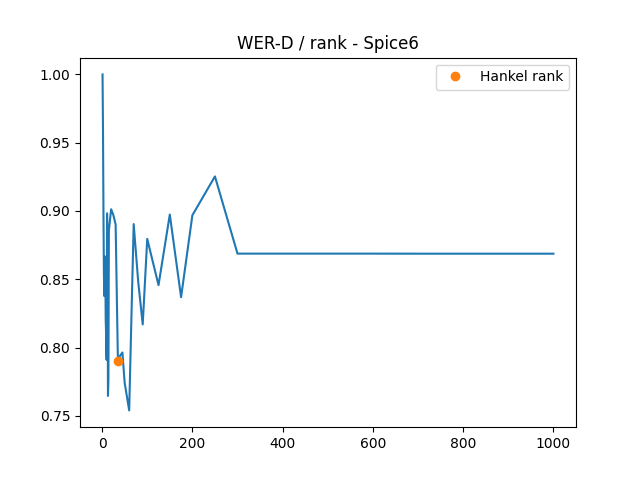}
\includegraphics[width=0.23\textwidth]{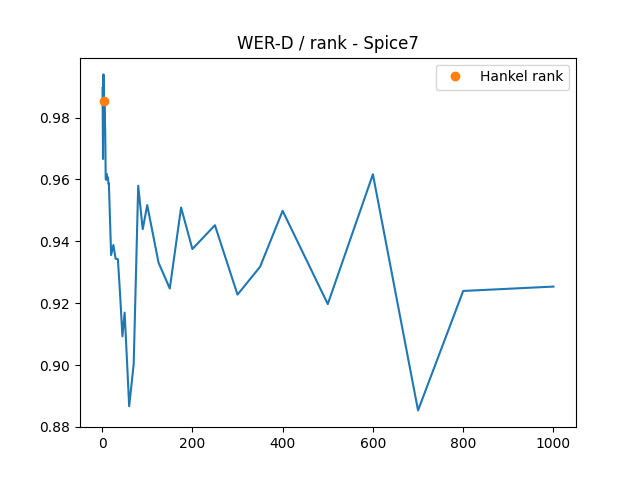}
\includegraphics[width=0.23\textwidth]{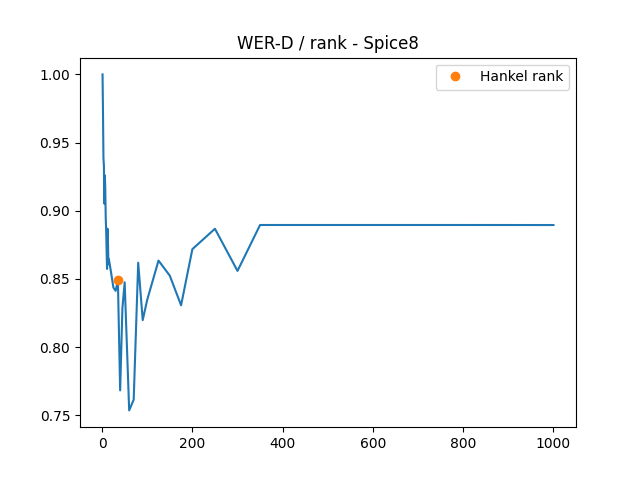}
\includegraphics[width=0.23\textwidth]{imgs/ranks/WER-D_rank-Spice9.png}
\includegraphics[width=0.23\textwidth]{imgs/ranks/WER-D_rank-Spice10.png}
\includegraphics[width=0.23\textwidth]{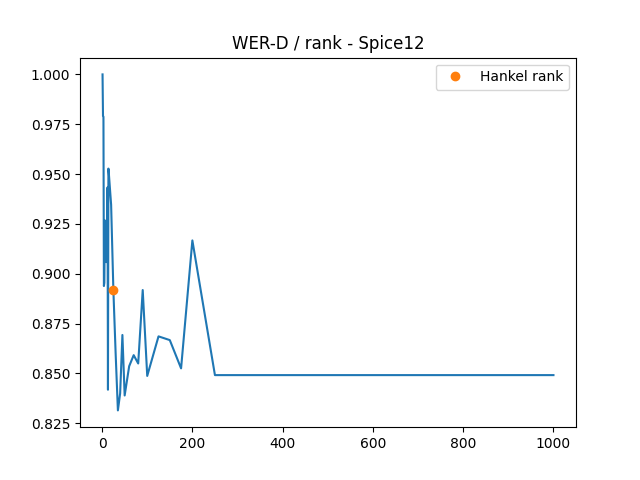}
\includegraphics[width=0.23\textwidth]{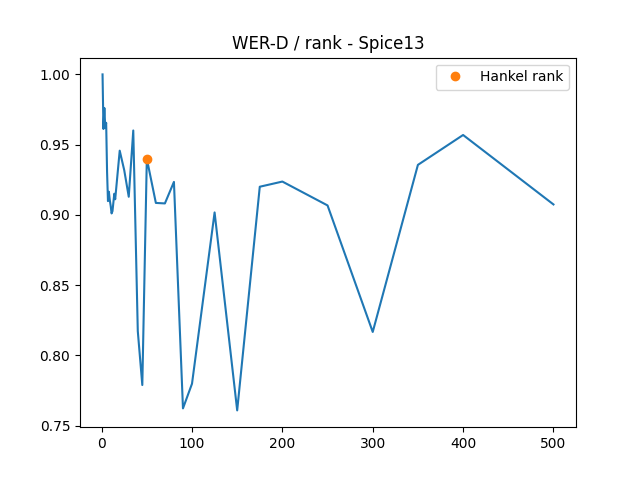}
\includegraphics[width=0.23\textwidth]{imgs/ranks/WER-D_rank-Spice14.png}
\includegraphics[width=0.23\textwidth]{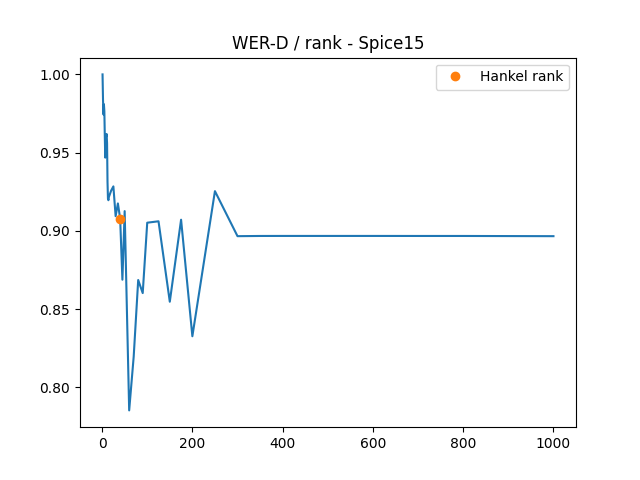}

\caption{Evolution of the quality in term of WER-D with the value of the rank hyper-parameter on the 8 last PAutomaC problems and the SPiCe ones, evaluated on $S_{RNN}$ with basis size $1000\times 1000$.}
\label{fig:ranks-pautomac-spice}
\end{center}
\end{figure}

\newpage
\pagenumbering{gobble}

\begin{figure}[htbp!]
\vspace{-1.5cm}
\begin{center}
\includegraphics[width=0.23\textwidth]{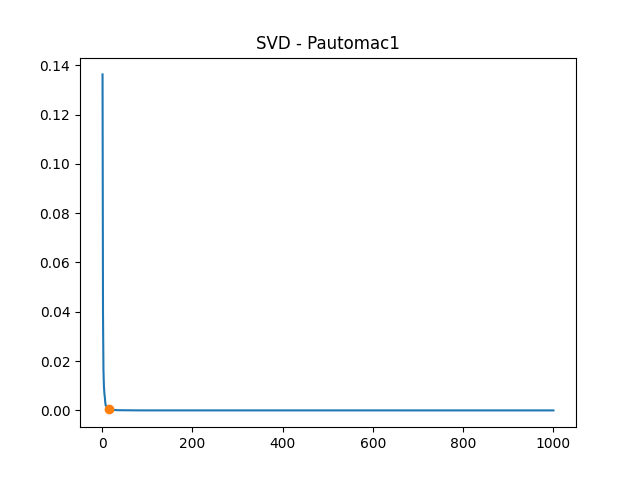}
\includegraphics[width=0.23\textwidth]{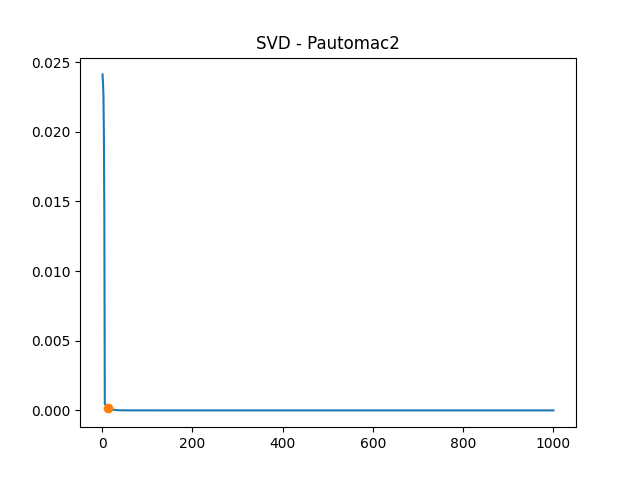}
\includegraphics[width=0.23\textwidth]{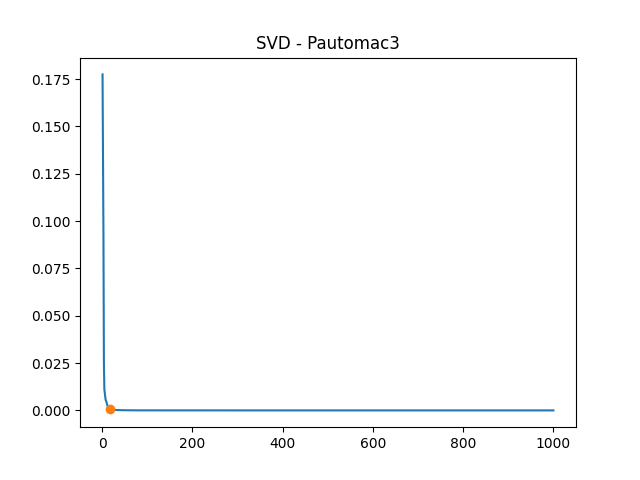}
\includegraphics[width=0.23\textwidth]{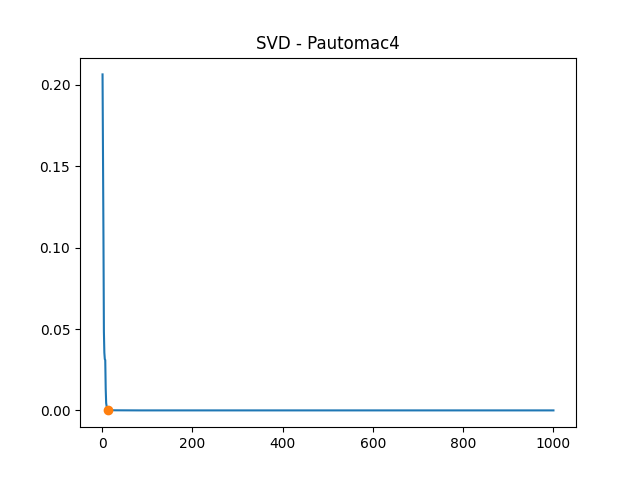}
\includegraphics[width=0.23\textwidth]{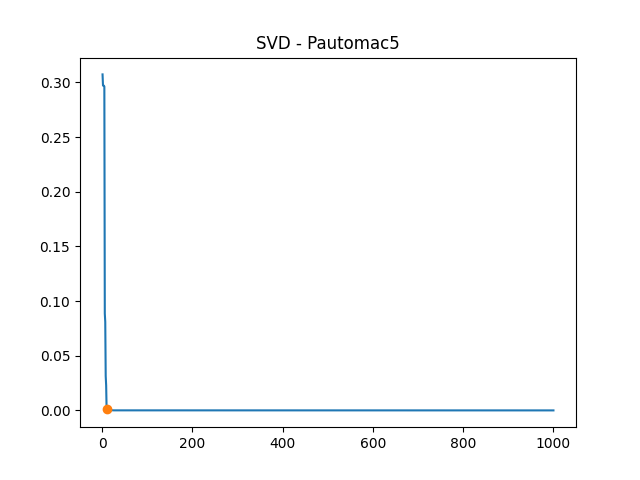}
\includegraphics[width=0.23\textwidth]{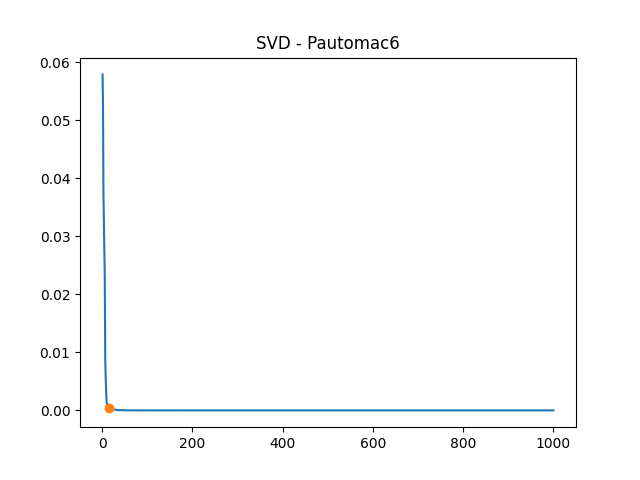}
\includegraphics[width=0.23\textwidth]{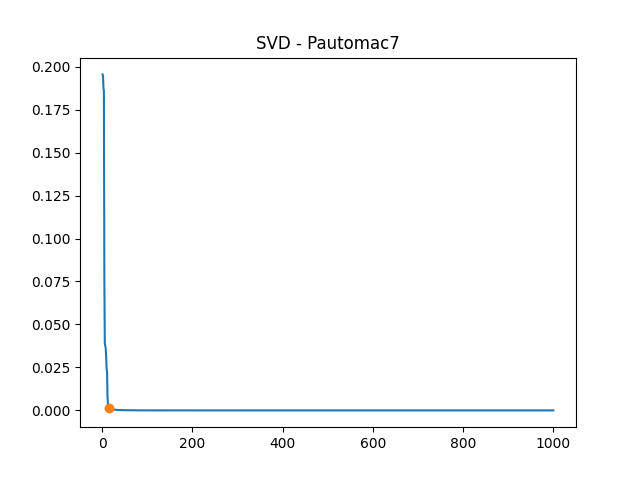}
\includegraphics[width=0.23\textwidth]{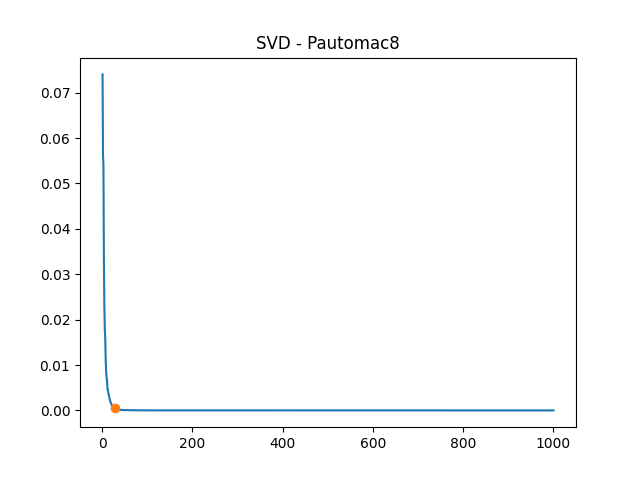}
\includegraphics[width=0.23\textwidth]{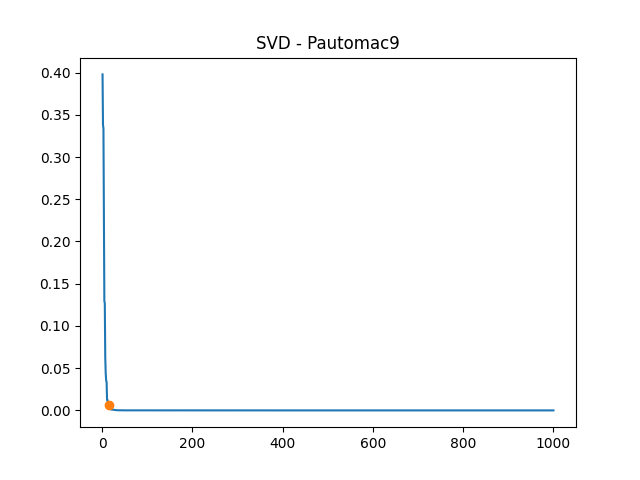}
\includegraphics[width=0.23\textwidth]{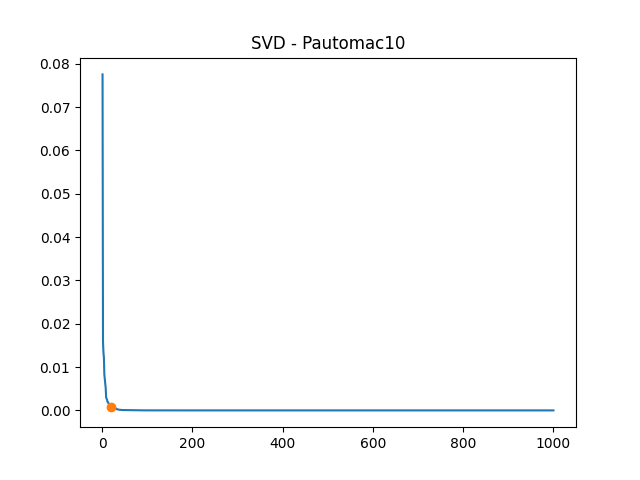}
\includegraphics[width=0.23\textwidth]{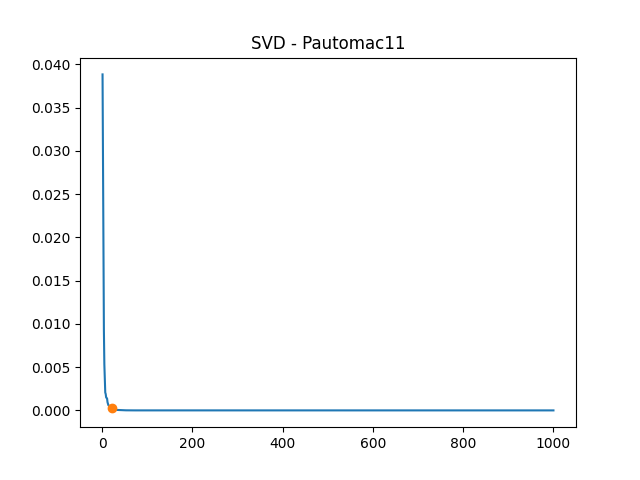}
\includegraphics[width=0.23\textwidth]{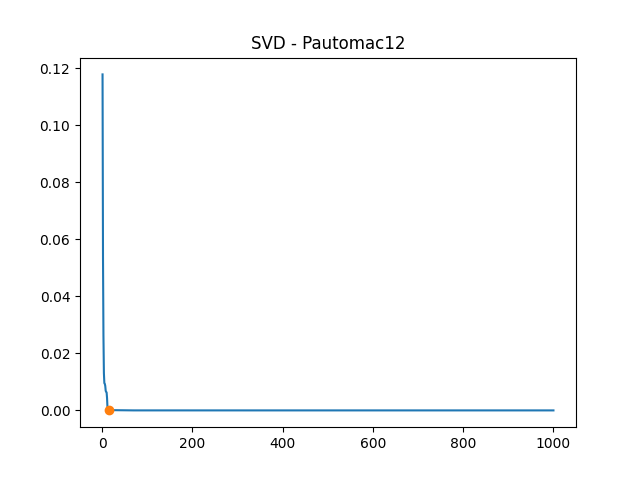}
\includegraphics[width=0.23\textwidth]{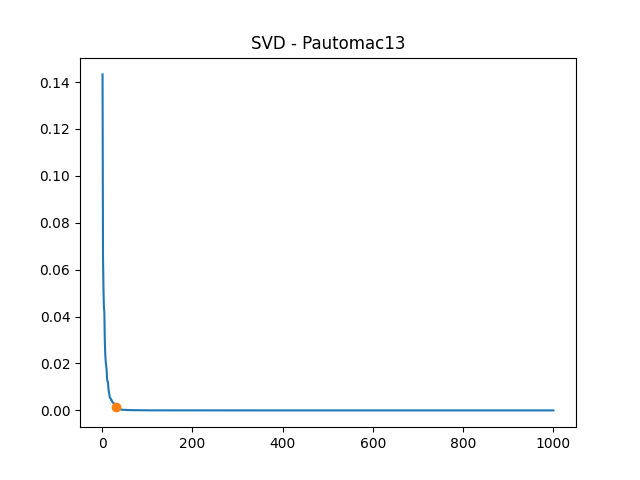}
\includegraphics[width=0.23\textwidth]{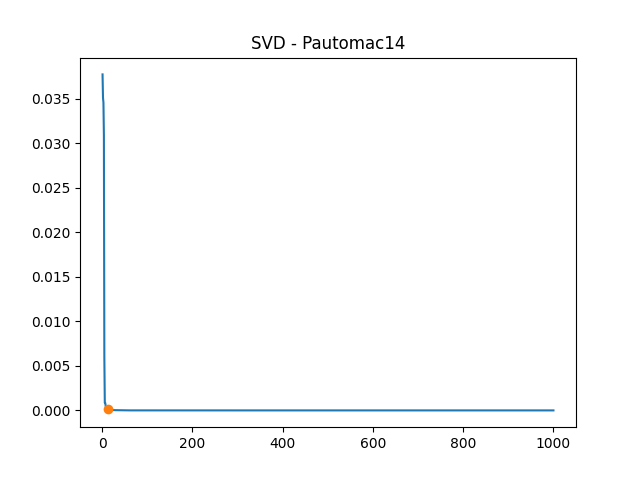}
\includegraphics[width=0.23\textwidth]{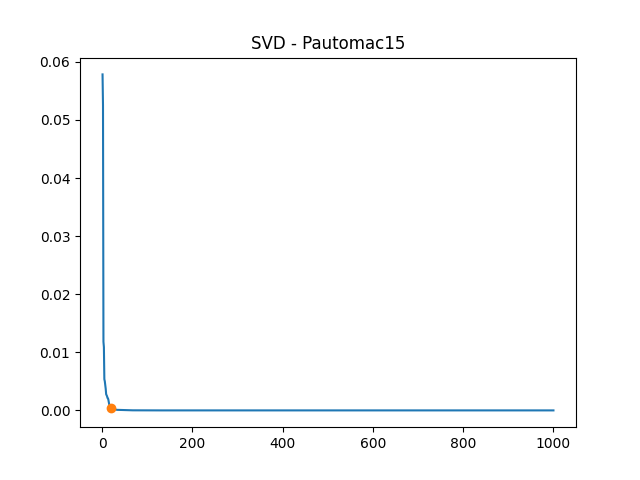}
\includegraphics[width=0.23\textwidth]{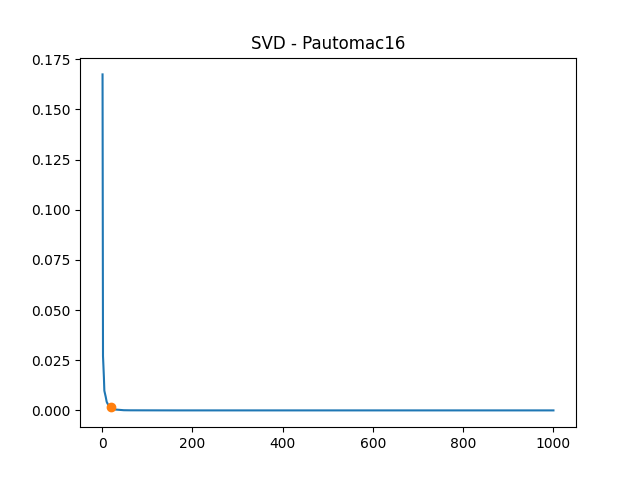}
\includegraphics[width=0.23\textwidth]{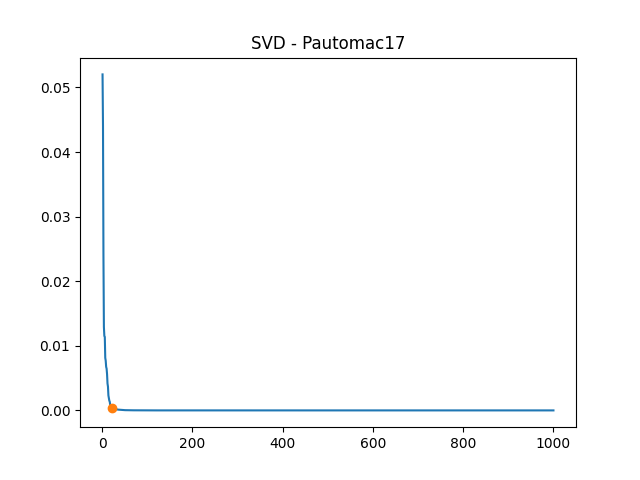}
\includegraphics[width=0.23\textwidth]{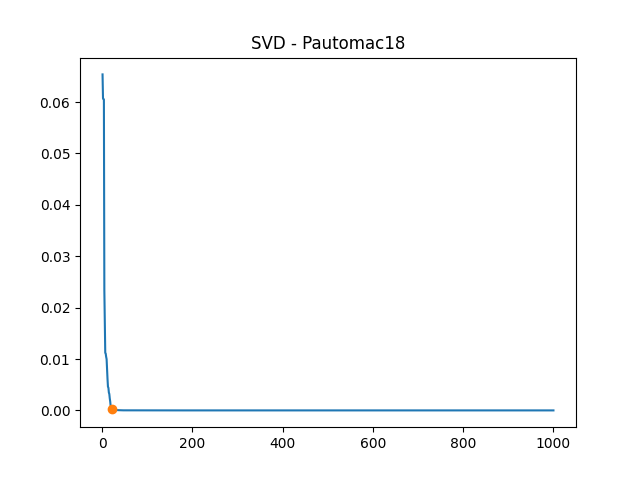}
\includegraphics[width=0.23\textwidth]{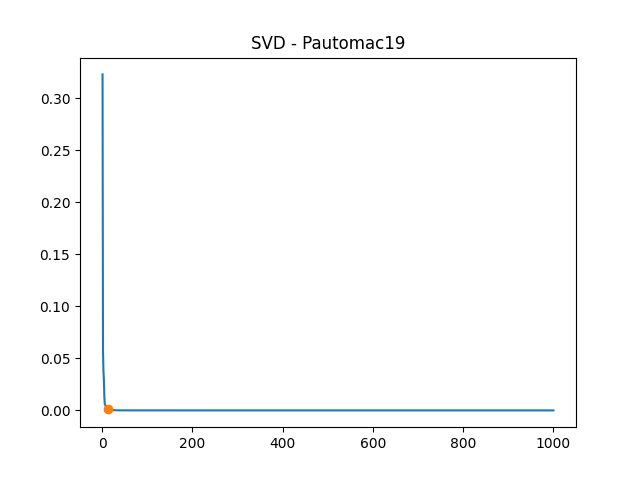}
\includegraphics[width=0.23\textwidth]{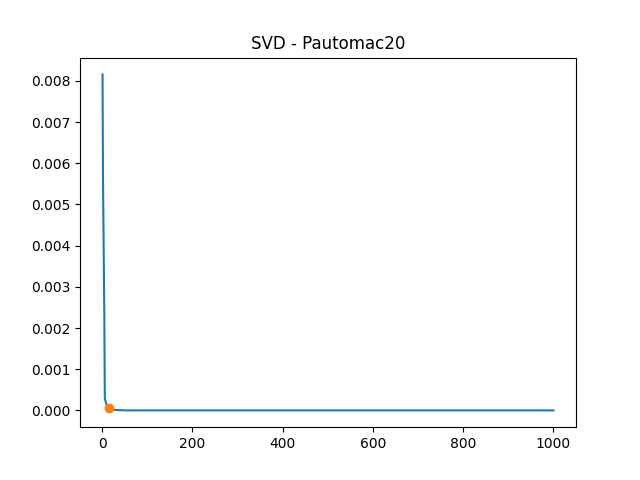}
\includegraphics[width=0.23\textwidth]{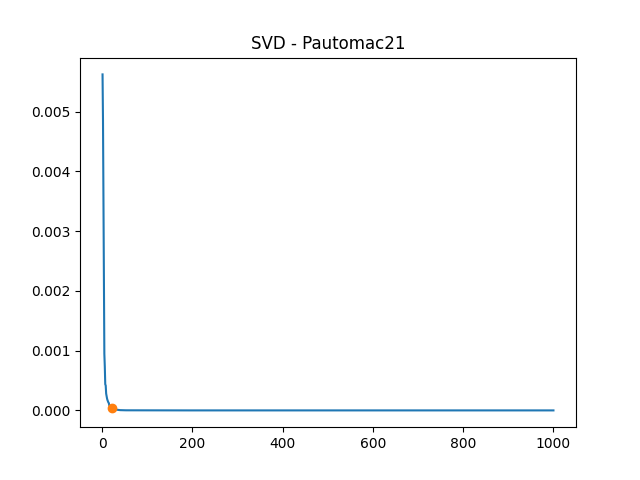}
\includegraphics[width=0.23\textwidth]{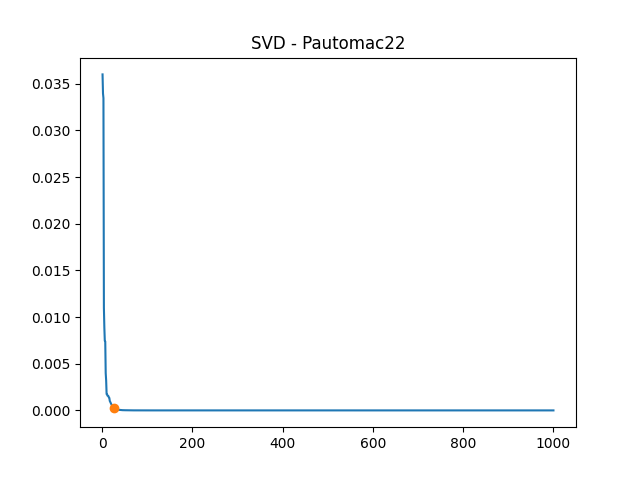}
\includegraphics[width=0.23\textwidth]{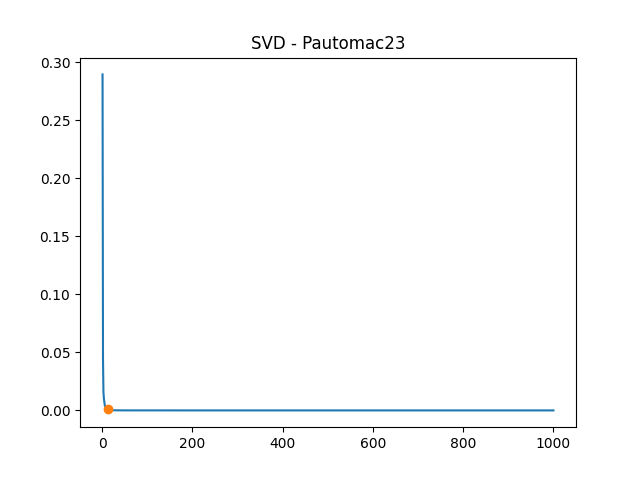}
\includegraphics[width=0.23\textwidth]{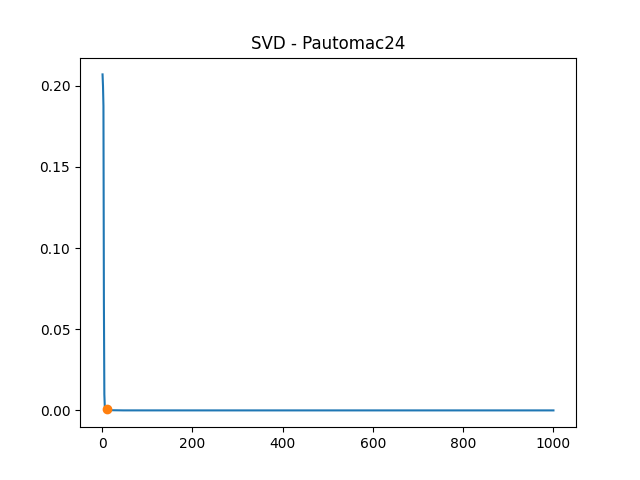}
\includegraphics[width=0.23\textwidth]{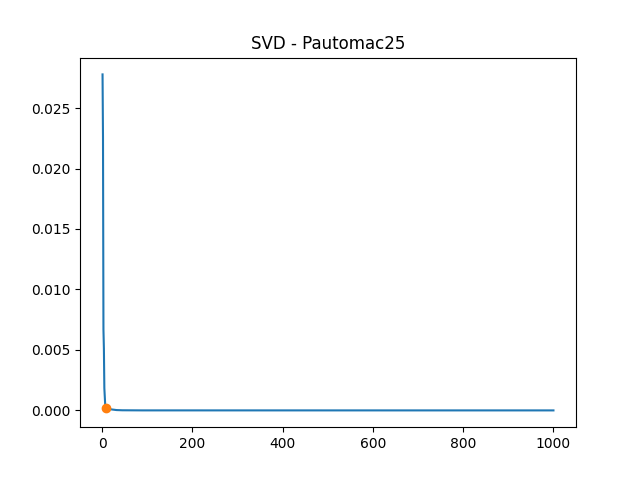}
\includegraphics[width=0.23\textwidth]{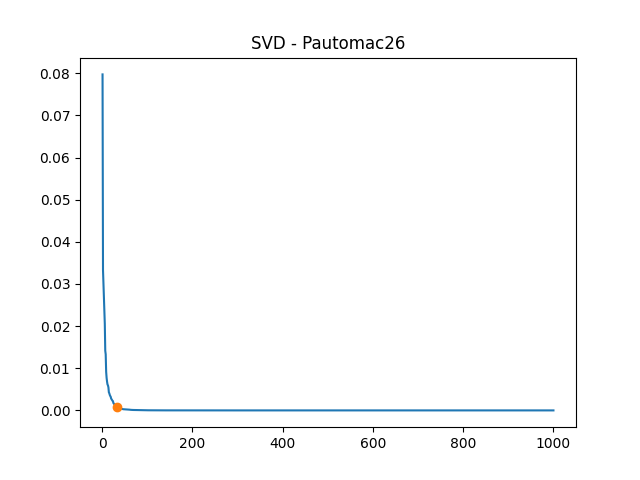}
\includegraphics[width=0.23\textwidth]{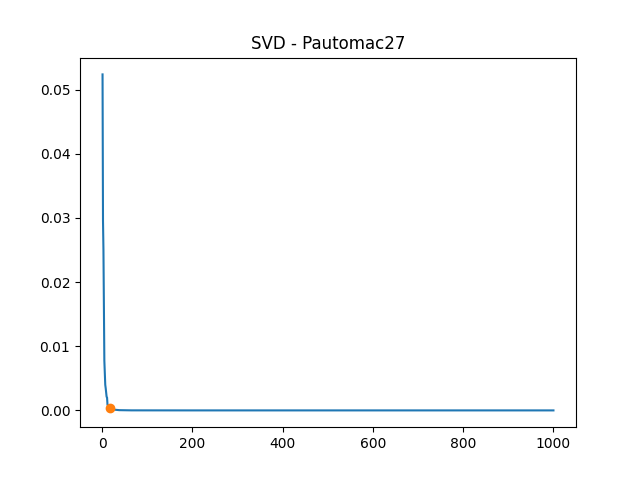}
\includegraphics[width=0.23\textwidth]{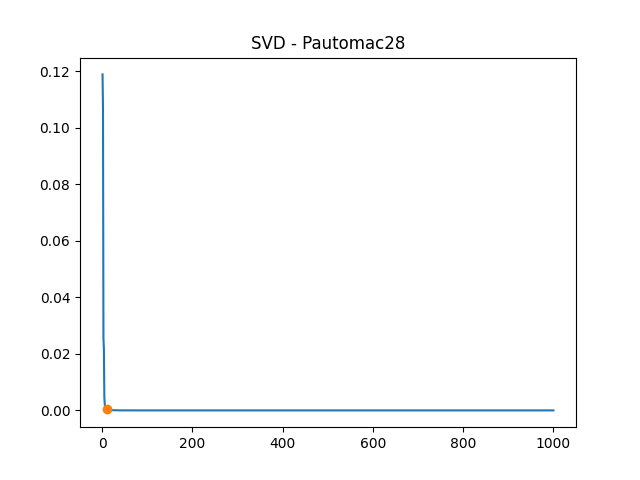}
\includegraphics[width=0.23\textwidth]{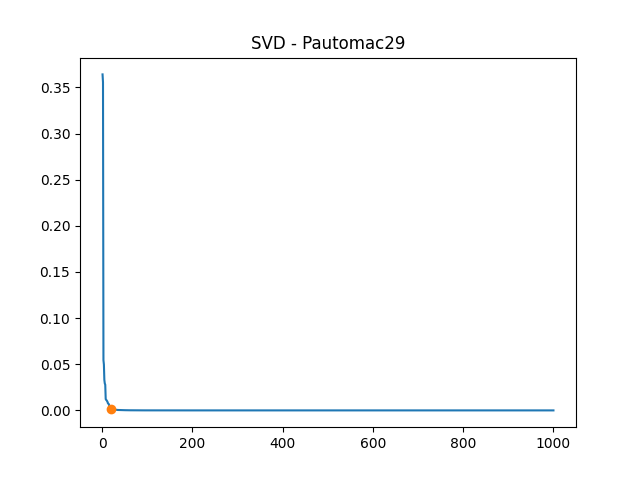}
\includegraphics[width=0.23\textwidth]{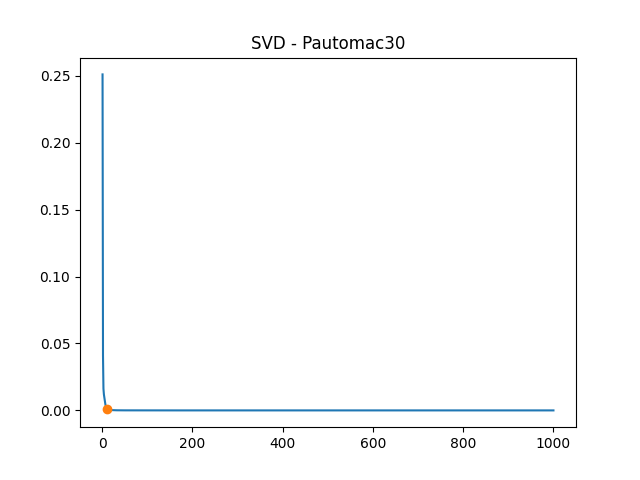}
\includegraphics[width=0.23\textwidth]{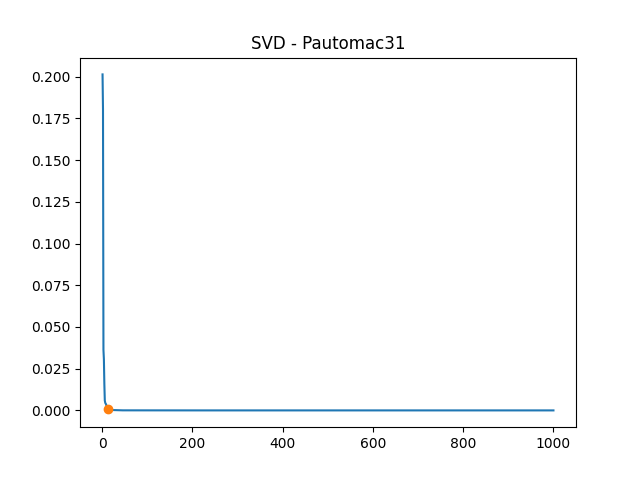}
\includegraphics[width=0.23\textwidth]{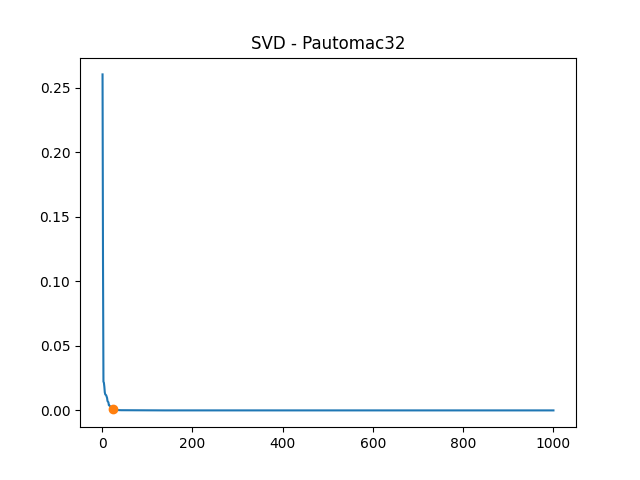}
\includegraphics[width=0.23\textwidth]{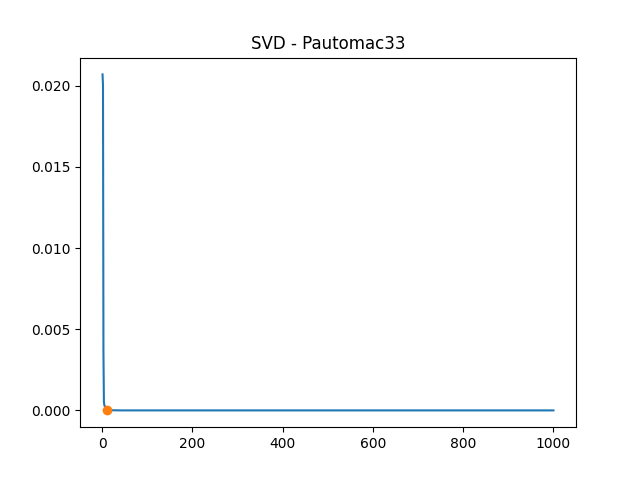}
\includegraphics[width=0.23\textwidth]{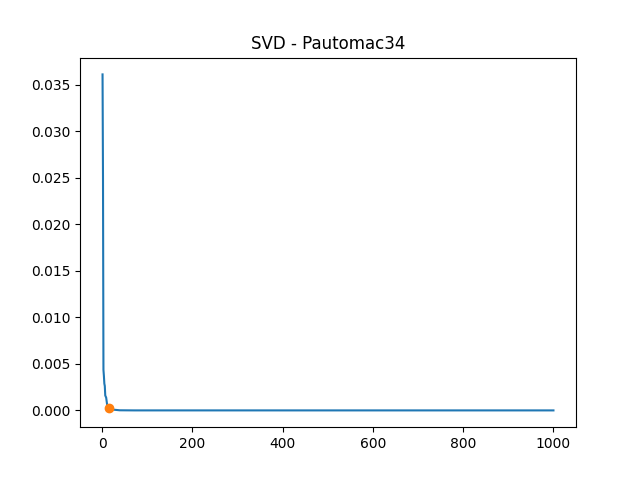}
\includegraphics[width=0.23\textwidth]{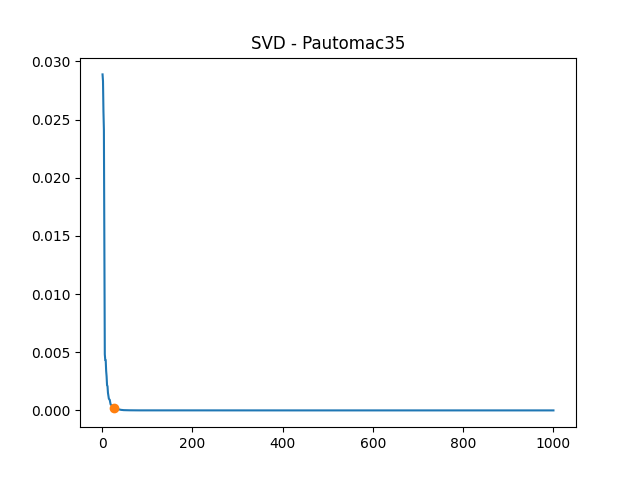}
\includegraphics[width=0.23\textwidth]{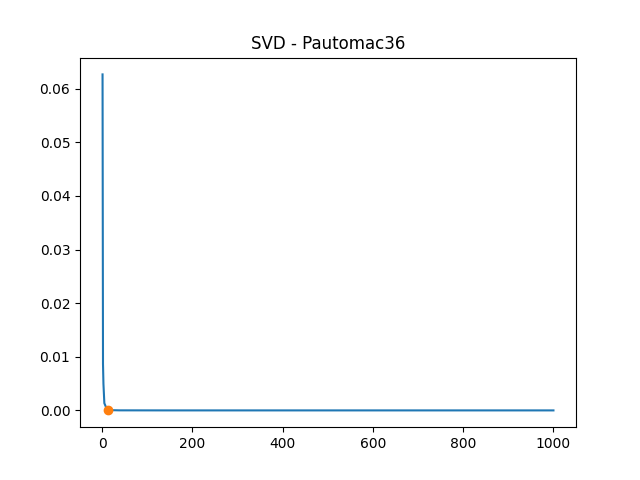}
\includegraphics[width=0.23\textwidth]{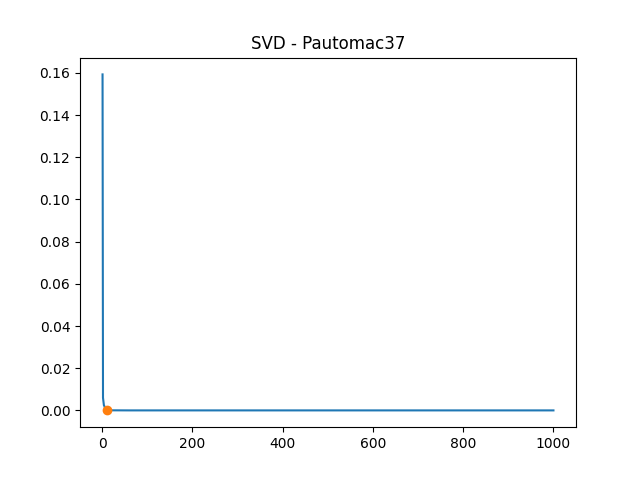}
\includegraphics[width=0.23\textwidth]{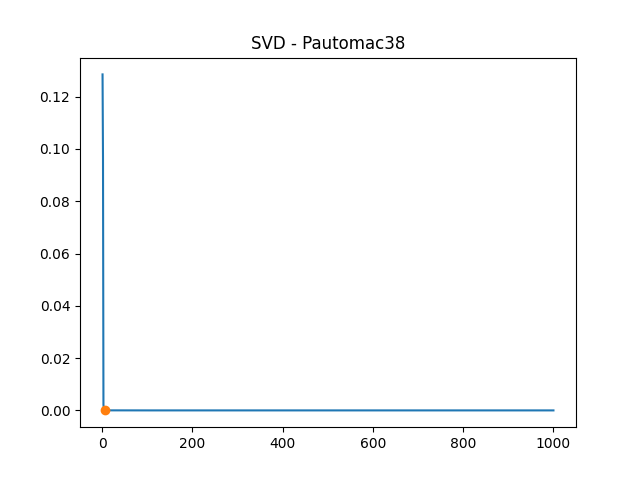}
\includegraphics[width=0.23\textwidth]{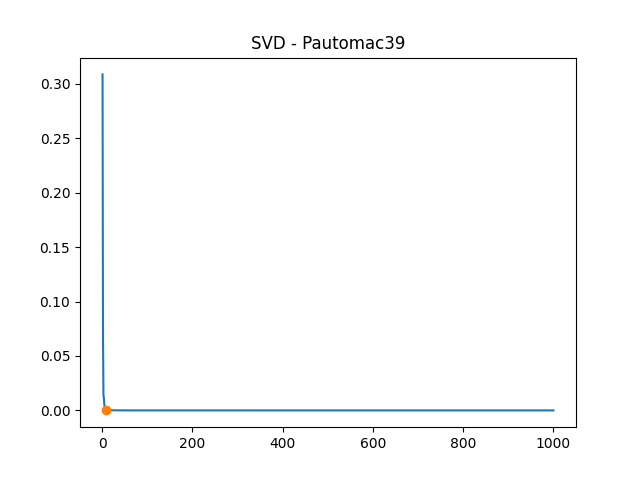}
\includegraphics[width=0.23\textwidth]{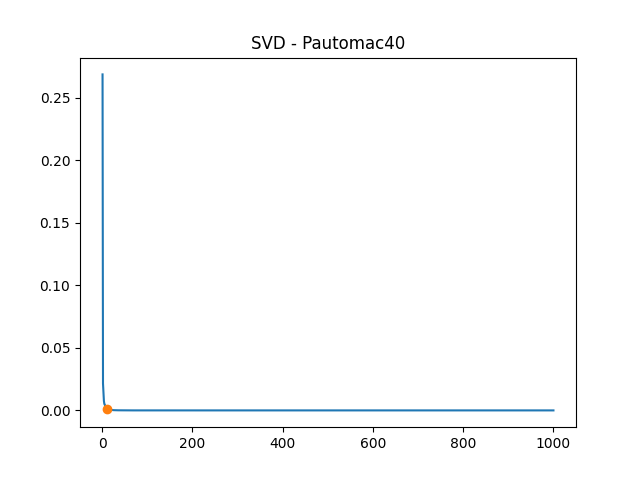}
\caption{Singular values of the Hankel matrix of the RNN learned on the 40 first PAutomaC datasets, with a $1000\times 1000$ basis. So called \textit{Hankel rank} is shown by a orange dot.} 
\label{fig:svds}
\end{center}
\end{figure}
\thispagestyle{empty}

\newpage
\begin{figure}[htbp!]
\begin{center}
\includegraphics[width=0.23\textwidth]{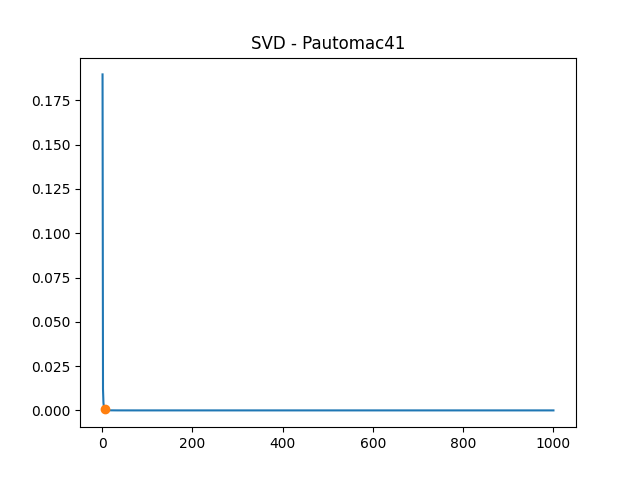}
\includegraphics[width=0.23\textwidth]{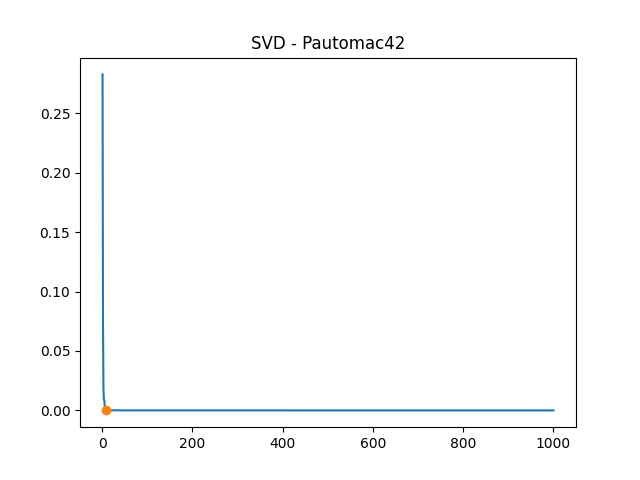}
\includegraphics[width=0.23\textwidth]{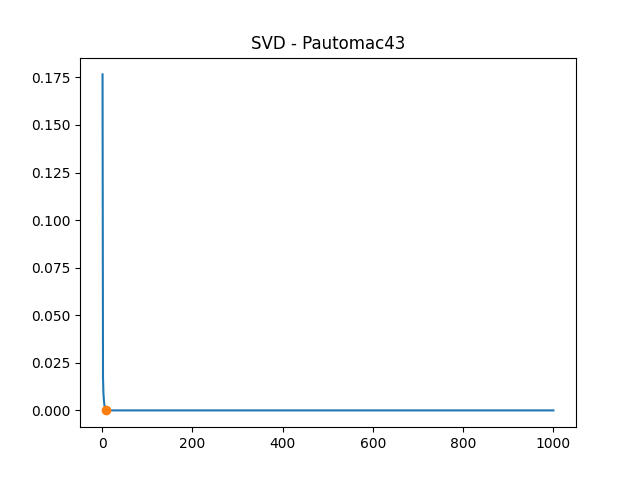}
\includegraphics[width=0.23\textwidth]{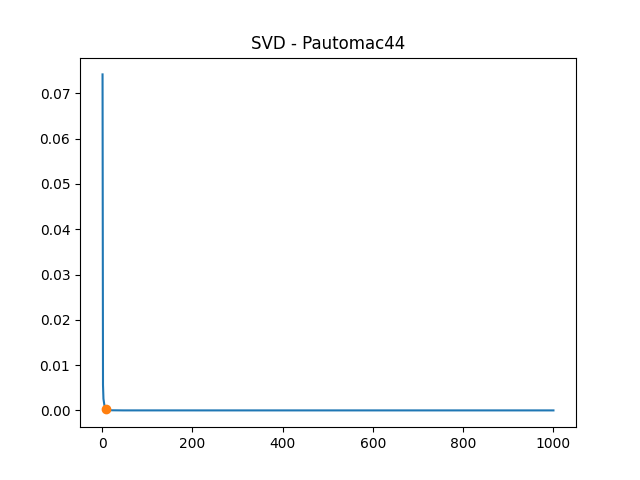}
\includegraphics[width=0.23\textwidth]{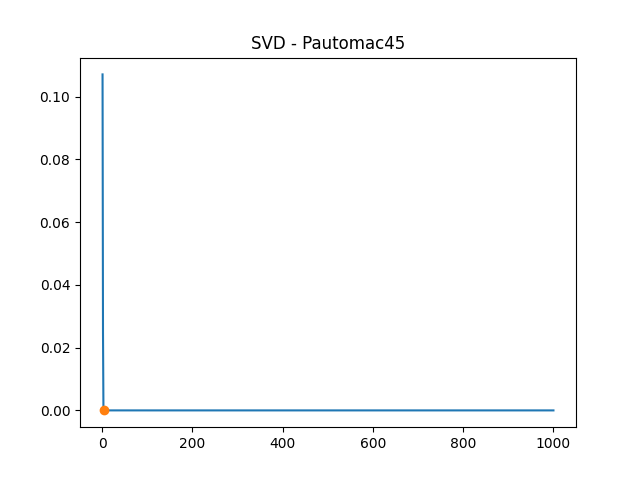}
\includegraphics[width=0.23\textwidth]{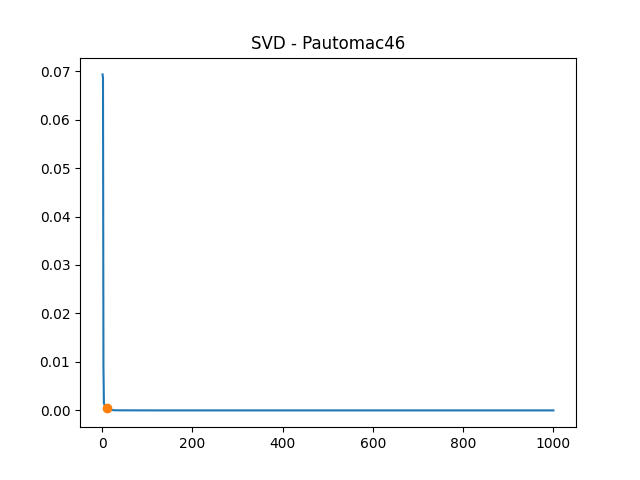}
\includegraphics[width=0.23\textwidth]{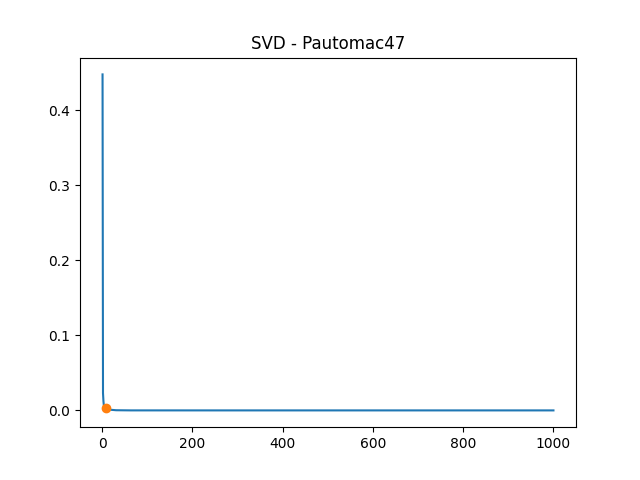}
\includegraphics[width=0.23\textwidth]{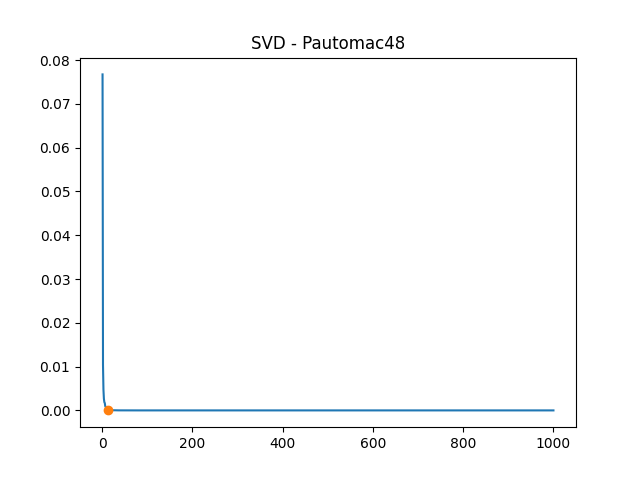}
\includegraphics[width=0.23\textwidth]{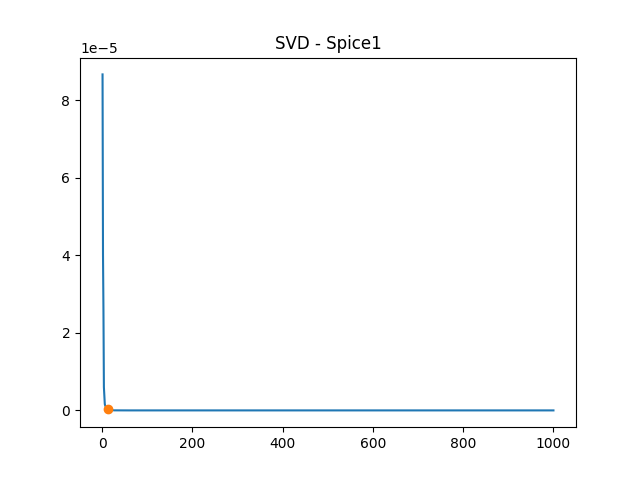}
\includegraphics[width=0.23\textwidth]{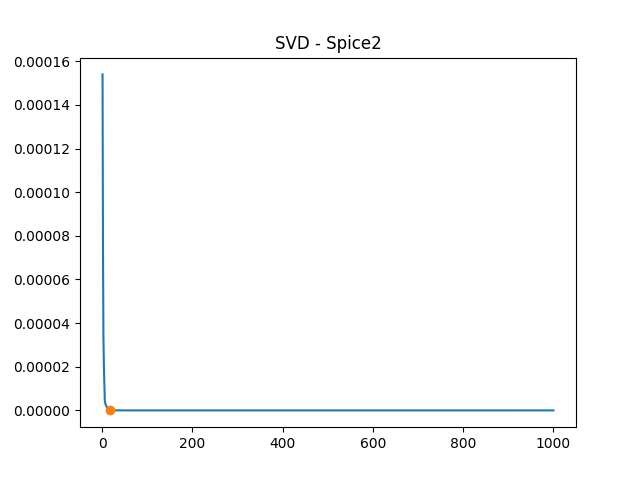}
\includegraphics[width=0.23\textwidth]{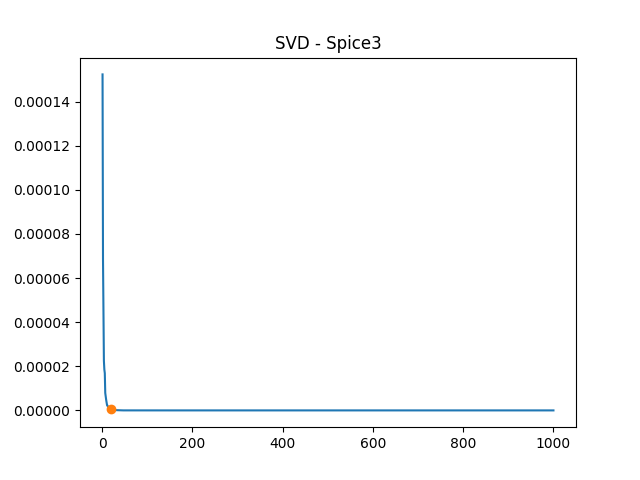}
\includegraphics[width=0.23\textwidth]{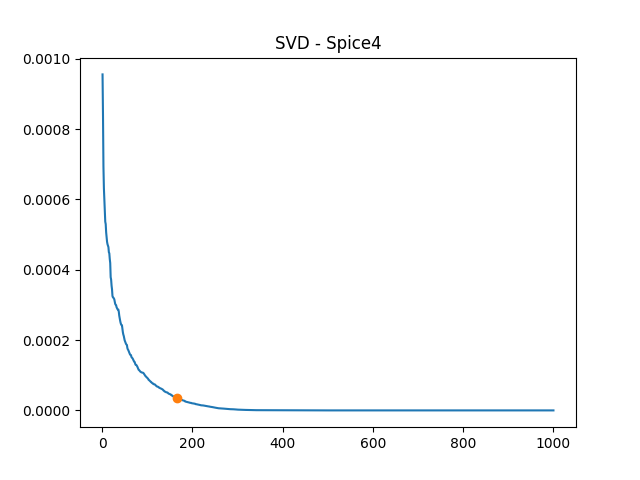}
\includegraphics[width=0.23\textwidth]{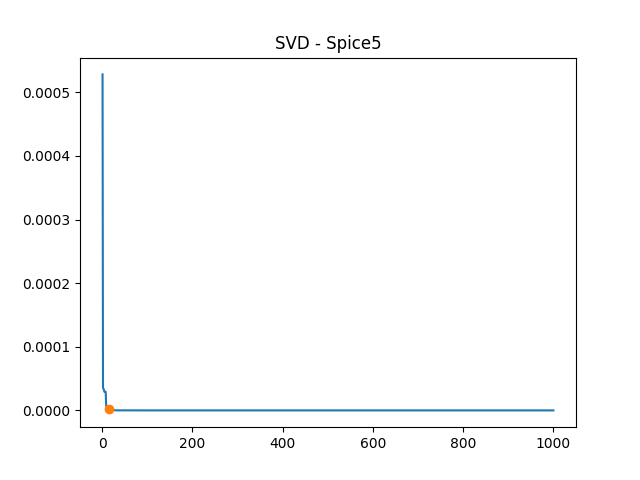}
\includegraphics[width=0.23\textwidth]{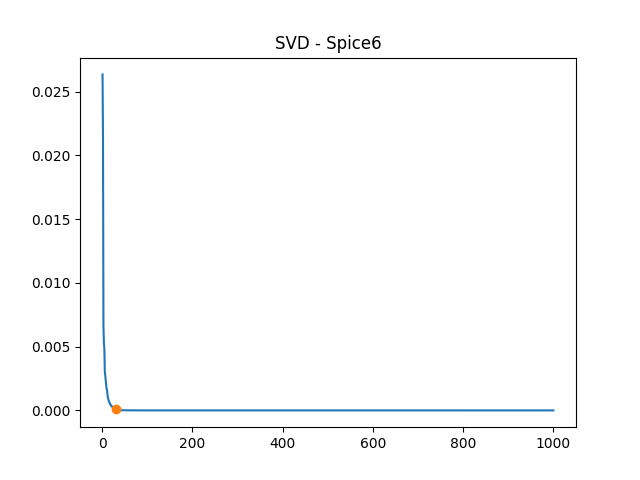}
\includegraphics[width=0.23\textwidth]{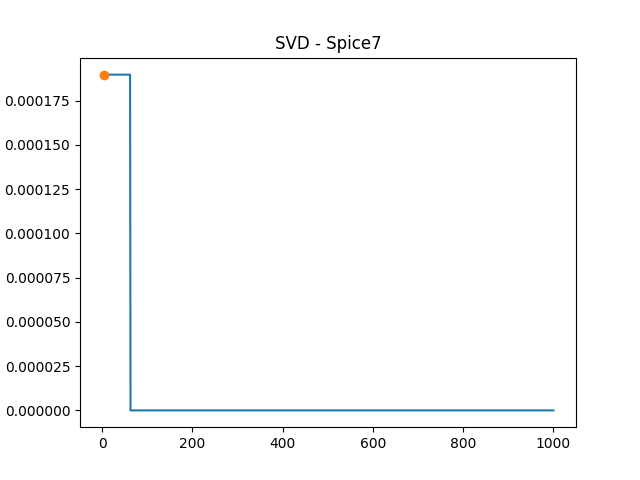}
\includegraphics[width=0.23\textwidth]{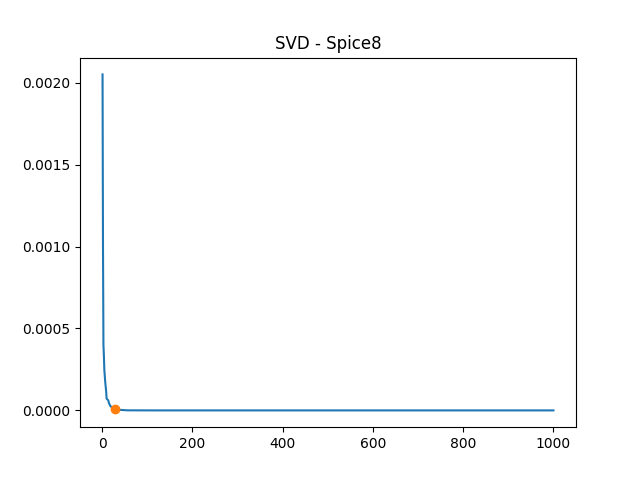}
\includegraphics[width=0.23\textwidth]{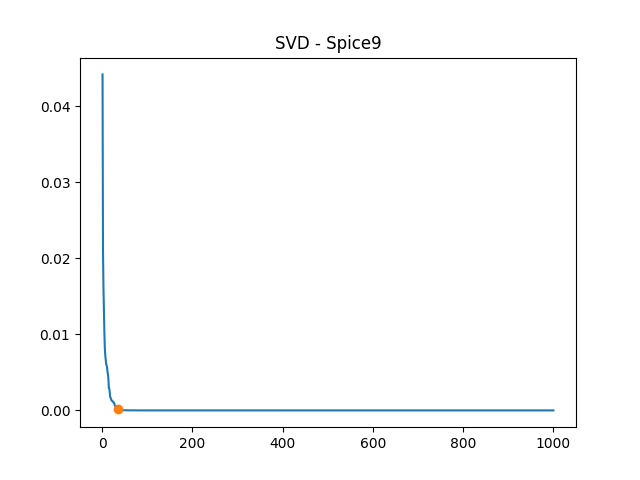}
\includegraphics[width=0.23\textwidth]{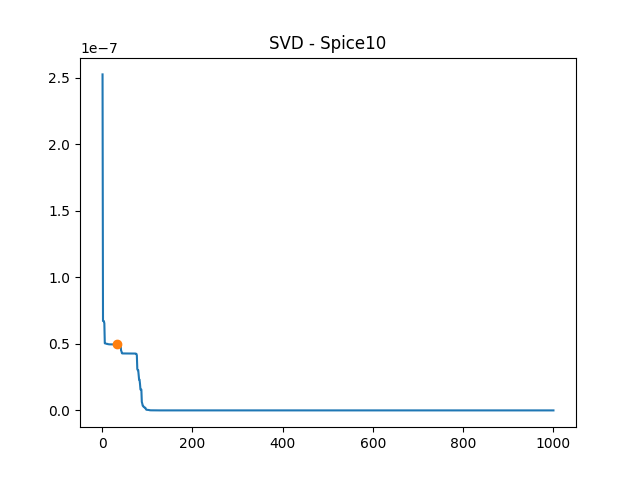}
\includegraphics[width=0.23\textwidth]{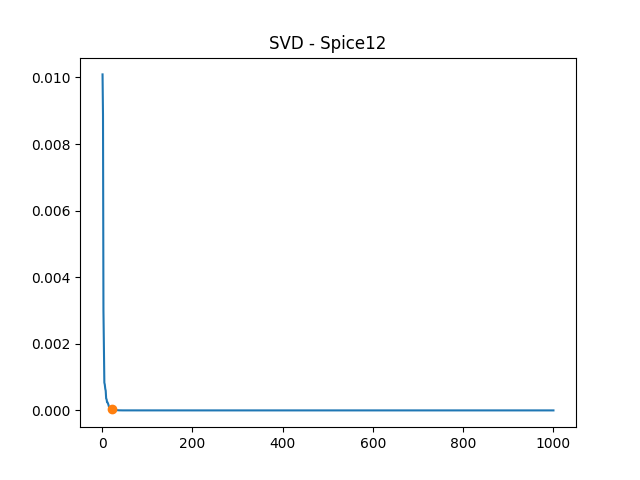}
\includegraphics[width=0.23\textwidth]{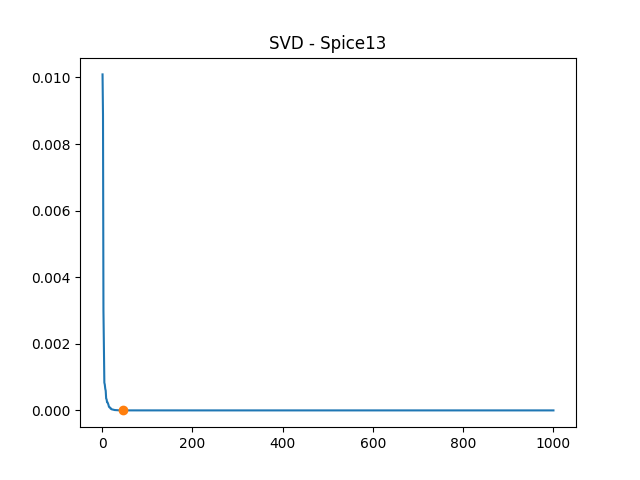}
\includegraphics[width=0.23\textwidth]{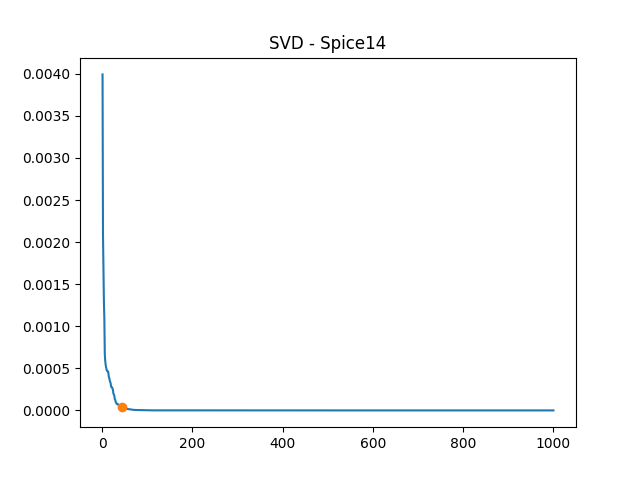}
\includegraphics[width=0.23\textwidth]{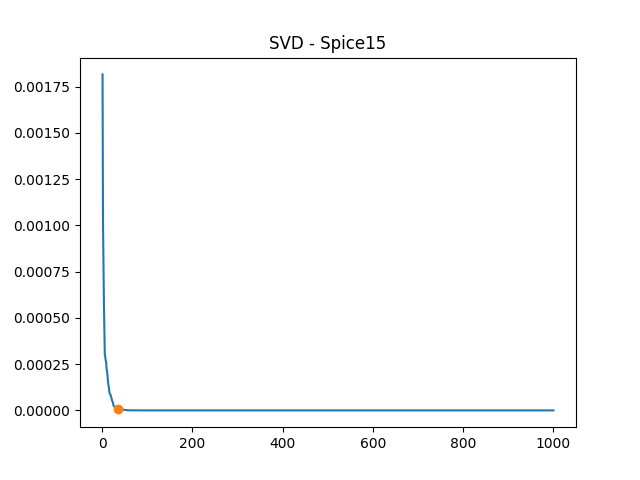}
\caption{Singular values of the Hankel matrix of the RNN learned on the last 8 PAutomaC datasets and the SPiCe ones, with $1000\times 1000$ bases.} 
\label{fig:svds2}
\end{center}
\end{figure}

\end{document}